\documentclass[10pt,twocolumn,letterpaper]{article}

\usepackage{wacv}              

\usepackage[pagebackref,breaklinks,colorlinks,allcolors=wacvblue]{hyperref}
\usepackage[capitalize]{cleveref} 
\usepackage{tabularx}
\usepackage{graphicx} 
\usepackage{pifont}
\newcommand{\cmark}{\ding{51}} 
\newcommand{\xmark}{\ding{55}} 
\usepackage{diagbox}
\usepackage{textcomp}
\DeclareUnicodeCharacter{25AA}{\textbullet}
\usepackage{float}
\usepackage{placeins}
\usepackage{adjustbox}
\usepackage{multirow, booktabs}
\usepackage{subcaption}

\definecolor{wacvblue}{rgb}{0.21,0.49,0.74}

\def\wacvPaperID{1113} 
\def\confName{WACV}
\def\confYear{2026}

\usepackage{multirow}
\usepackage{dsfont}
\usepackage{bbm}
\usepackage{enumitem}
\usepackage{makecell}
\usepackage[accsupp]{axessibility}

\title{Patch-wise Retrieval: A Bag of Practical Techniques for Instance-level Matching}

\author{
Wonseok Choi$^{1}$ \quad
Sohwi Lim$^{2}$ \quad
Nam Hyeon-Woo$^{1}$ \quad
Moon Ye-Bin$^{1}$ \quad \\
Dong-Ju Jeong$^{3}$ \quad
Jinyoung Hwang$^{3}$ \quad
Tae-Hyun Oh$^{2}$\\[0.4em]
$^{1}$POSTECH \quad
$^{2}$KAIST \quad
$^{3}$Samsung Research\\[0.2em]
}

\author{
Wonseok Choi$^{1}$ \quad
Sohwi Lim$^{2}$ \quad
Nam Hyeon-Woo$^{1}$ \quad
Moon Ye-Bin$^{1}$ \quad \\
Dong-Ju Jeong$^{3}$ \quad
Jinyoung Hwang$^{3}$ \quad
Tae-Hyun Oh$^{2}$\\[0.4em]
$^{1}$POSTECH \quad
$^{2}$KAIST \quad
$^{3}$Samsung Research\\[0.6em]
}

\begin{document}
\maketitle
\begin{abstract}
Instance-level image retrieval aims to find images containing the same object as a given query, despite variations in size, position, or appearance. To address this challenging task, we propose Patchify, a simple yet effective patch-wise retrieval framework that offers high performance, scalability, and interpretability without requiring fine-tuning. Patchify divides each database image into a small number of structured patches and performs retrieval by comparing these local features with a global query descriptor, enabling accurate and spatially grounded matching. To assess not just retrieval accuracy but also spatial correctness, we introduce LocScore, a localization-aware metric that quantifies whether the retrieved region aligns with the target object. This makes LocScore a valuable diagnostic tool for understanding and improving retrieval behavior. We conduct extensive experiments across multiple benchmarks, backbones, and region selection strategies, showing that Patchify outperforms global methods and complements state-of-the-art reranking pipelines. Furthermore, we apply Product Quantization for efficient large-scale retrieval and highlight the importance of using informative features during compression, which significantly boosts performance. Project website: https://wons20k.github.io/PatchwiseRetrieval/
\end{abstract}
\section{Introduction}

Instance retrieval is the task of finding images in a database that contain the same visual instance as the one presented in a query image. The target instance may be a specific object, product, person, or landmark, and retrieved images are expected to contain the same object despite variations in perspective, scale, lighting, or background~\cite{7298790, jenicek2019feardarkimageretrieval}. The task is both practical and broadly applicable, supporting a variety of real-world scenarios~\cite{lu2023ovir, sakaguchi2024object, yuan2021eproductmillionscalevisualsearch, zhan2021product1, drsplat25, bang2026bth, meol2026}, such as retrieving relevant personal photos from a user's device and our daily lives.
 
Since the appearance of the same instance in a target image, including variations in position, scale, or orientation, instance retrieval requires more discriminative and robust feature representations.
To improve instance retrieval under challenging conditions, recent methods~\cite{suma2024ames, tolias2025ilias} have adopted two main strategies: fine-tuning large encoders on domain-specific datasets or employing reranking pipelines with hundreds of dense local descriptors per image. While these approaches offer high accuracy, they introduce practical limitations. Acquiring large-scale instance-level annotations for fine-tuning is often infeasible, and maintaining dense local features per image significantly increases memory and computational overhead. Efficient alternatives that achieve strong performance without fine-tuning or dense local matching remain underexplored.

\begin{figure}[t]
    \centering
    \includegraphics[width=\linewidth]{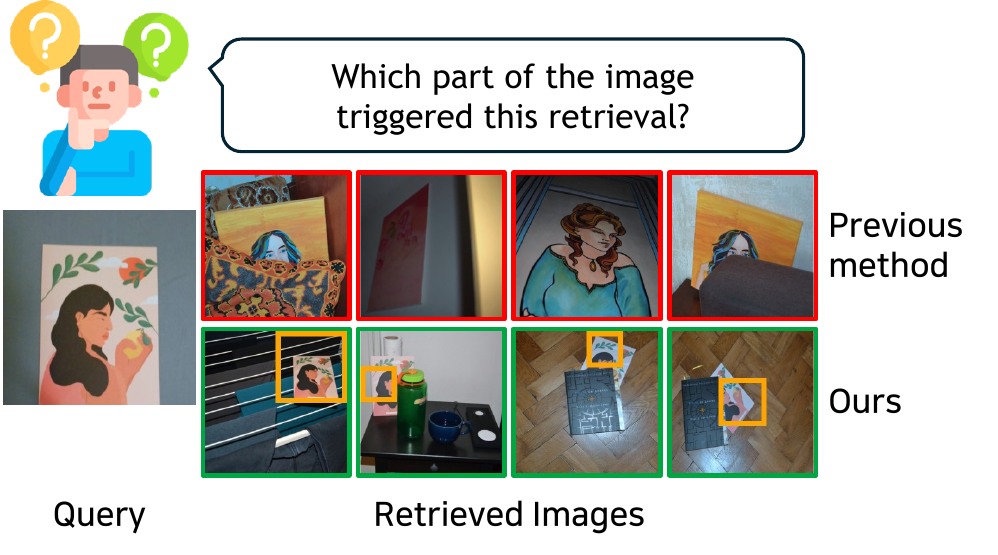} 
    \par\vspace{1em}
    {\small
    \begin{tabularx}{\linewidth}{l*{2}{>{\centering\arraybackslash}X}}
        \hline
        & \textbf{Previous method} & \textbf{Ours} \\
        \hline
        Interpretability & \textcolor{red}{\xmark} & \textcolor{OliveGreen}{\cmark} \\
        Performance & \textcolor{red}{\xmark} & \textcolor{OliveGreen}{\cmark} \\
        Scalability & \textcolor{red}{\xmark} & \textcolor{OliveGreen}{\cmark} \\
        \hline
    \end{tabularx}
    }
    \caption{
        \textbf{Overview of our patch-wise retrieval framework.}
        Given a query image, global methods~\cite{zhai2023sigmoid, oquab2023dinov2, tolias2025ilias} often retrieve visually similar images without indicating what triggered the match. Our approach retrieves correct instances with spatial interpretability by identifying the most relevant regions. 
        Despite using far fewer patches than conventional local feature methods, it achieves strong performance and high scalability.
        }
    \label{fig:teaser}
\end{figure}

In this work, we propose \textbf{Patchify}, a simple yet effective patch-wise retrieval pipeline that requires only a handful of local features per image. By leveraging multi-scale grid patches with pretrained visual encoders, Patchify enables efficient, interpretable, and accurate retrieval.

To validate our approach, we conduct a series of experiments evaluating the impact of local features across various settings. We show that patch-wise representations consistently outperform global descriptors, especially with models pretrained on large-scale datasets. We also compare different region sampling strategies, SOTA local feature methods, and find that our simple grid-based Patchify method achieves competitive performance with more complex alternatives. Finally, we apply product quantization for scalable retrieval and observe that training PQ with informative patches, such as those aligned with ground-truth regions, further improves accuracy.

\noindent We summarize our main contributions as follows:
\begin{itemize}
    \item We present Patchify retrieval pipeline, a simple yet effective patch-wise retrieval framework based on multi-scale patch matching that operates without fine-tuning or region proposals.
    \item We introduce LocScore, a localization-aware metric that provides a fine-grained evaluation of both retrieval accuracy and spatial alignment.
    \item We conduct extensive experiments across diverse backbones and region sampling strategies, and show that our method offers competitive performance with greater simplicity and interoperability.
    \item We demonstrate that combining local features with product quantization enables scalable retrieval, and find that using informative patches for PQ training further boosts accuracy.
\end{itemize}

\section{Related work}

\paragraph{Instance Level Image Retrieval Methods}
Early studies in instance level image retrieval relied on hand crafted descriptors such as SIFT~\citep{lowe2004distinctive} and SURF~\citep{bay2006surf}. These approaches extract sparse local keypoints and measure similarity with Bag of Words and geometric verification.

With deep learning, the field moved to dense global CNN descriptors, where pooling schemes such as GeM~\citep{gem8382272} became dominant and global aggregation methods strengthened robustness~\citep{gordo2016deep, babenko2016visual, tolias2016particular}. Large-scale pre-training further produced strong global descriptors~\citep{radford2021learning, ilharco2021openclip, caron2021emerging} that transfer across classification, segmentation and retrieval~\citep{gao2024clip, zhou2021ibot, liang2023openvocabularysemanticsegmentationmaskadapted, rao2022densecliplanguageguideddenseprediction, luo2021clip4clipempiricalstudyclip, patel2019self}. Nevertheless these global representations can overlook local evidence that discriminates the target instance when it is small, occluded, or away from the image center.

To recover local cues, two-stage pipelines typically perform retrieval with global descriptors and then rerank top candidates using local evidence. A common template compares sets of local descriptors among shortlisted images and promotes those with consistent correspondences, using attention-based matching, correlation agreement, or structural self-similarity~\citep{iccv21_reranking_transformers, cvpr22_correlation_verification, cvpr23_revisiting_self_similarity}. 

However, at large scale, storing local descriptors for every database image is prohibitive in memory, and extracting them on demand for each candidate introduces substantial latency. As a result, most approaches confine local processing to a limited first-stage shortlist. Recent work explores how to scale correspondence reasoning under such constraints. For example, long-context sequence modeling enables reasoning over many local descriptors within a single model invocation~\citep{cvpr24_locore}. 

In parallel, AMES proposes an efficient asymmetric similarity estimator that achieves practical memory usage and serves as a strong baseline within the ILIAS large-scale evaluation protocol~\citep{suma2024ames, tolias2025ilias}. Despite these advances, much of the existing literature is developed and evaluated primarily on landmark-centric datasets. This setting motivates approaches that generalize beyond landmarks while preserving the complementary strengths of global and local representations. Along this direction, UnED targets general-purpose instance retrieval by adapting modern backbones through simple linear transformations, and the ILIAS protocol demonstrates that such adaptation transfers effectively beyond landmark domains~\citep{ypsilantis2023towards, tolias2025ilias}.

Patchify densely samples small sliding windows across scales together with a full image patch, embeds them with a frozen backbone, and scores each database image by its best matching patch. This recovers local evidence that global-only descriptors miss on small or off-center targets and avoids the detector dependence of sparse keypoint or proposal methods. Unlike reranking, which needs hundreds of local descriptors per image and therefore computes or stores them only for a shortlist, Patchify indexes a small fixed set per image, so a handful of features provide local signal in the first pass with predictable memory and latency. It requires no training, works plug and play with newer backbones, and, with product quantization, keeps storage close to global baselines while preserving clear spatial attribution.

\paragraph{Explainability via Localization Cues}
Several retrieval works sought to recover localization cues while keeping a single global descriptor. Integral max pooling builds one global vector and then searches rectangular windows on the feature map of each candidate image using integral images and a generalized mean to approximate channel-wise max pooling \citep{tolias2016particular}. Deep image retrieval learns region-pooled features but aggregates them into one vector for scoring \citep{gordo2016deep}. Because multiple regions are fused into a single representation, the rank signal may not identify an exact region, and localization is obtained only after retrieval, which can dilute fine evidence relative to using explicit localized cues.

Patchify is explainable by design. Each database image is indexed by a small set of multi-scale grid patches, and the query is compared directly to these regional descriptors, so every match is attributed to a concrete patch. We also introduce LocScore, which combines retrieval rank with intersection over union between the winning patch and the ground truth region to measure alignment and retrieval quality together. We report continuous, mean, and thresholded forms to serve different robustness needs. In contrast to global aggregation, Patchify provides an explicit link from score to region while preserving strong retrieval accuracy.


\begin{figure*}[!ht]
    \centering
    \includegraphics[width=\linewidth]{Figures/main/main.pdf}
    \caption{\textbf{Overview of our Patchify retrieval pipeline.}
    We illustrate the L2 configuration, where multi-scale grid patches (1×1, 2x2, and 3×3) are extracted and individually passed through the image encoder for feature extraction and indexing. On the right, we show how retrieval is performed for a query: patch-level similarities are computed across the database, and each image is ranked based on the highest-scoring patch it contains.
    }
    \label{fig:patchify}
\end{figure*}

\section{Method}

Our goal is to design an instance-level image retrieval framework that is both accurate and interpretable, while maintaining efficiency in terms of computation and memory. To this end, we propose \textbf{Patchify}, a single-stage retrieval pipeline that matches structured, non-overlapping patches between query and database images, enabling spatial alignment and interpretability. We also introduce \textbf{LocScore}, a localization-aware metric that evaluates whether retrievals are based on the correct image regions. In the following sections, we describe the Patchify pipeline (\cref{sec:patchify}), the LocScore metric for evaluating spatial alignment (\cref{sec:locscore}).

\subsection{Patchify: Patch-wise Retrieval pipeline}
\label{sec:patchify}
Our method, Patchify, introduces a simple and efficient way to integrate local spatial cues into instance-level image retrieval without relying on complex region proposals. As illustrated in Figure~\ref{fig:patchify}, we divide each database image into a set of structured, non-overlapping grid patches at multiple scales. For example, the L2 configuration includes all patches from the 1×1, 2×2, and 3×3 grids, forming a cumulative multi-scale representation. Similarly, L0 uses only a single global 1×1 patch, while L3 further adds 4×4 grid patches on top of L2, providing finer spatial coverage. Each patch is independently encoded using a frozen image encoder (e.g., CLIP, DINOv2), resulting in a compact set of local descriptors per image.

During retrieval, the query image is encoded as a single global descriptor, following standard practice for scalability. We then compute the similarity between this query descriptor and all patch-level descriptors from each database image. For each image, the patch with the highest similarity score is selected to represent its relevance to the query. The final ranking is determined by sorting all database images based on these max similarity scores.

Compared to conventional two-stage retrieval pipelines that rely on hundreds of local features per image and expensive cross-feature comparisons, Patchify uses only a small number of structured patches, leading to significant reductions in both memory and computation. To further enhance efficiency, we apply Product Quantization~\citep{PQ} to compress the patch descriptors and enable fast approximate nearest neighbor search. Although the method is simple, it achieves strong performance and additionally provides spatial interpretability, as it clearly indicates which specific patch contributed to the retrieval result. This interpretable design not only enhances transparency in retrieval but also enables diagnostic evaluation through LocScore, which quantifies how well the retrieved region aligns with the target object.

\subsection {LocScore: Localization-aware Metrics}
\label{sec:locscore}
To evaluate the effectiveness of local features in instance retrieval, we introduce a localization-aware metric, LocScore, which quantifies not only whether the correct image is retrieved, but also how accurately the retrieved region aligns with the target object. Given a query image, the system returns a ranked list of retrieved images along with their most similar local patch.

To reflect not just the presence of correct retrievals but also their order in the ranked list, we weight each prediction by its retrieval precision. This allows us to evaluate localization performance in a retrieval-aware manner, rewarding predictions that are both accurate and highly ranked.

Let $B^{n,i}_{\text{gt}}$ and $B^{n,i}_{\text{pred}}$ denote the ground-truth and predicted bounding boxes, respectively, for the $i$-th ground-truth image of the $n$-th query. Suppose this ground-truth image is retrieved at rank $r^{n,i}$, and let $h^{n,i}$ denote the number of ground-truth positives retrieved within the top-$r^{n,i}$ positions.

We first define the localization score for a single query $n$ as:
\begin{equation}
    \text{LocScore}^{(n)} = \frac{1}{I_n} \sum_{i=1}^{I_n} \frac{h^{n,i}}{r^{n,i}} \cdot \text{IoU}(B^{n,i}_{\text{gt}}, B^{n,i}_{\text{pred}}) ,
\end{equation}
\noindent where $I_n$ is the number of ground-truth positives for query $n$. If a ground-truth image is not retrieved for a given query, its IoU is treated as zero.

The overall LocScore across all queries is then computed by averaging the per-query scores:
\begin{equation}
    \text{LocScore} = \frac{1}{N} \sum_{n=1}^{N} \text{LocScore}^{(n)},
\end{equation}
\noindent where $N$ is the total number of queries.

\vspace{0.5em}
We further introduce a thresholded variant of the metric to provide a binary notion of successful localization. While the original formulation captures the degree of spatial alignment through continuous IoU values, this version evaluates whether the predicted region satisfies a minimum localization quality. Specifically, only matches whose IoU exceeds a specified threshold $\delta$ are considered correct:
\begin{equation}
    \text{LocScore}^{(n)}(\delta) = \frac{1}{I_n} \sum_{i=1}^{I_n} \frac{h^{n,i}}{r^{n,i}} \cdot \mathbb{I}\left[\text{IoU}(B^{n,i}_{\text{gt}}, B^{n,i}_{\text{pred}}) \ge \delta\right] ,
\end{equation}
\begin{equation}
    \text{LocScore}(\delta) = \frac{1}{N} \sum_{n=1}^{N} \text{LocScore}^{(n)}(\delta),
\end{equation}\label{eq:one}
\noindent where $\mathbb{I}[\cdot]$ is the indicator function.

\vspace{0.5em}
To reduce sensitivity to the choice of threshold $\delta$, we average over a set of thresholds $\mathcal{T}$:
\begin{equation}
    \text{mLocScore} = \frac{1}{|\mathcal{T}|} \sum_{\delta \in \mathcal{T}} \text{LocScore}(\delta),
\end{equation}
\noindent where $\mathcal{T} = \{0.3, 0.4, 0.5\}$ in our experiments.

\vspace{0.5em}
This formulation enables both fine-grained and thresholded interpretations of localization performance. Our proposed metrics evaluate two critical aspects: whether the correct image is retrieved, and whether the retrieval is based on the correct local region. Even when the correct image is retrieved, a low LocScore indicates that the retrieval relied on spatially irrelevant content. This makes LocScore a powerful and interpretable diagnostic tool, akin to class activation maps (CAM) in explainable AI.

\section{Experiment}

We evaluate our Patchify method (\Cref{fig:patchify}) on two standard instance retrieval benchmarks, INSTRE~\cite{wang2015instre} and ILIAS core set~\cite{tolias2025ilias} for most of the comparison, reporting both mean Average Precision (mAP) and our localization-aware metric, LocScore, which reflects retrieval accuracy and spatial alignment. In the main paper, we present results using the continuous LocScore formulation, and provide analyses with thresholded variants in the Supplementary Material (\Cref{sec:sup_locscore}).

\subsection{Effectiveness of Patch-wise Representation}
We compare Patchify-based local representations with conventional global descriptors to understand their relative effectiveness in instance-level retrieval. 
For this experiment, We adopt two widely used pretrained encoders, DINOv2~\cite{oquab2023dinov2} and CLIP~\cite{ilharco2021openclip}, to compare global and local features.

\begin{figure*}[!t]
  \centering
  \includegraphics[width=0.99\textwidth]{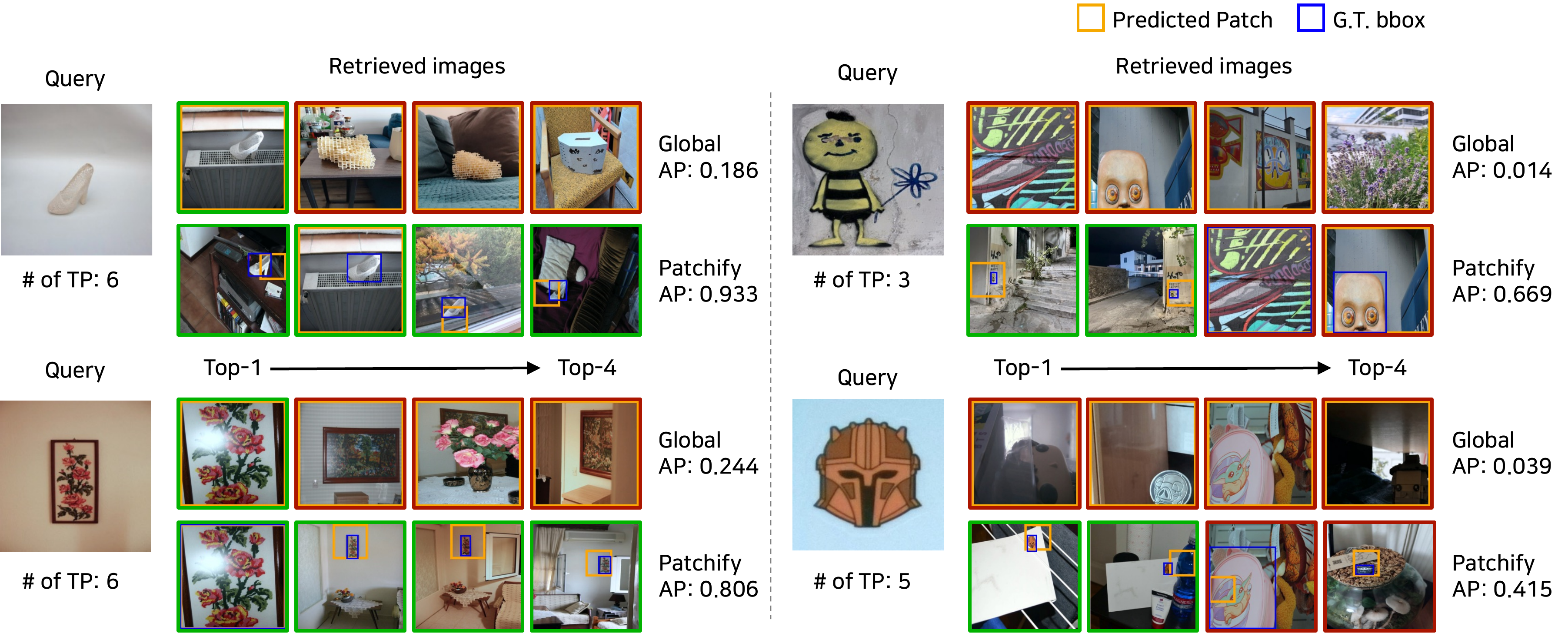}
  \caption{\textbf{Qualitative comparison between Global and Patchify on ILIAS.} While global features often fail to retrieve small or non-centered instances, Patchify successfully localizes and retrieves correct instances with better spatial grounding.}
  \label{fig:qualitative}
\end{figure*}

\paragraph{Global vs. Local: A Comparative Analysis}
As summarized in \Cref{tab:global_local_map_dual}, local features consistently outperform global ones across different datasets, encoders, and metrics. This highlights the importance of spatial granularity in improving both retrieval performance and localization quality. We observe local features provide a substantial boost, particularly in challenging scenarios such as occluded or off-center objects. 
To further contextualize these findings, we conduct a complementary study that compares patch-wise and global representations under challenging conditions such as variation in instance size, spatial position, and image lighting. This study identifies which feature type performs better under each condition and clarifies when local features provide meaningful advantages. The full setup and results are provided in the Supplementary Material. 

\begin{table}[!ht]
\centering
\caption{mAP (\%) and LocScore (\%) of global and local features on INSTRE and ILIAS. We adopt DINOv2 and CLIP as visual encoders. Using local features are consistently helpful in improving the performance regardless of the kind of visual encoder, dataset, and metrics.}
\resizebox{\linewidth}{!}{ 
\begin{tabular}{llcccc}
\toprule
\multirow{2}[2]{*}{\textbf{Encoder}} & \multirow{2}[2]{*}{\textbf{Type}} 
& \multicolumn{2}{c}{\textbf{INSTRE}} 
& \multicolumn{2}{c}{\textbf{ILIAS}} \\
\cmidrule(lr){3-4} \cmidrule(lr){5-6}
 &  & \textbf{mAP} & \textbf{LocScore} & \textbf{mAP} & \textbf{LocScore} \\
\midrule
\multirow{2}{*}{DINOv2} 
& global & 57.70 & 15.07 & 40.56 & 12.18 \\
& local  & 72.54 & 22.22 & 57.52 & 18.49 \\
\cmidrule{1-6}
\multirow{2}{*}{CLIP} 
& global & 73.84 & 18.10 & 31.60 & 9.18 \\
& local  & 87.57 & 30.37 & 53.35 & 17.95 \\
\bottomrule
\end{tabular}

}
\label{tab:global_local_map_dual}
\end{table}

\paragraph{Qualitative result} 
As shown in \Cref{fig:qualitative}, we compare the retrieved images of the global and local methods. We observe that Patchify can identify small and non-centered objects. 
These observations indicate that local features play a crucial role in instance retrieval, particularly under diverse visual conditions affecting the target objects.

\paragraph{Investigation of visual backbone}
To better understand the impact of model architecture and pretraining data scale on retrieval performance, we conduct a comparative analysis across various visual encoders. Our results indicate that Transformer-based models such as DINOv2, CLIP, and SigLIP consistently outperform CNN-based counterparts, demonstrating the effectiveness of modern architectural designs in both retrieval accuracy and localization quality. Notably, ConvNeXt pretrained on LAION-2B, despite being a CNN, achieves performance on par with Transformer-based models. This observation underscores the importance of large-scale pretraining, which significantly enhances the generalization ability of learned features across architectures. Detailed results can be found in Section~\ref{sec:sup_backbone}.

\subsection{Interpreting LocScore}
\label {subsec:locscore}

\begin{figure}[!t]
  \centering
  \includegraphics[width=\linewidth]{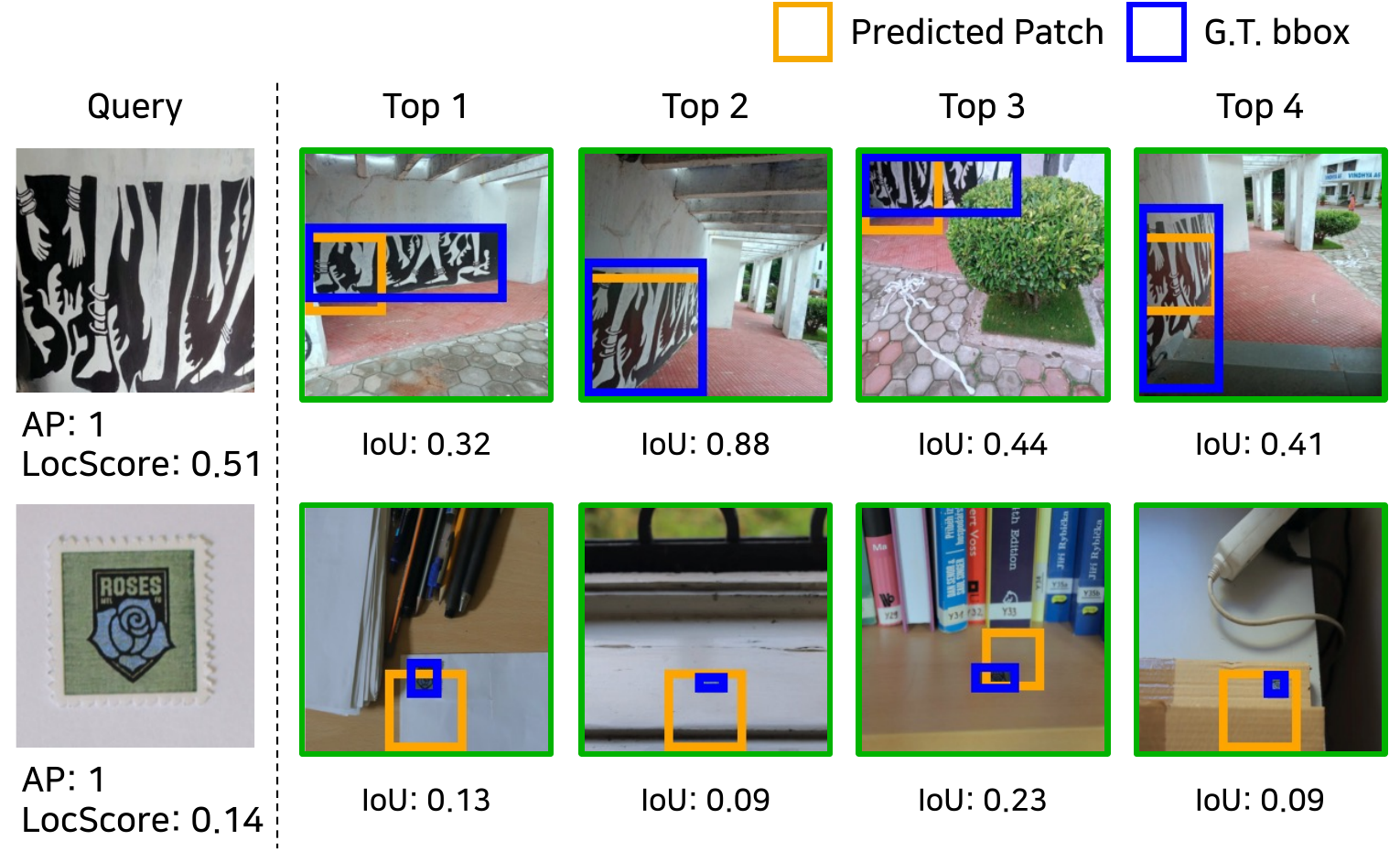}
  \caption{\textbf{Comparison of AP and LocScore for evaluating retrieval quality.}
    Both query examples achieve perfect AP (\ie, all ground-truth images are retrieved), yet their LocScores differ significantly due to localization accuracy. This highlights how LocScore complements AP by incorporating spatial alignment quality.
    }
  \label{fig:comparison_ap_locscore}
\end{figure}

In this section, we first compare AP and LocScore and explain how they differ. Next, we present a case study that shows which factors most strongly change LocScore and how rank position and overlap shape the score. Finally, we describe the variants of LocScore and explain when each variant is most appropriate in practice. Due to space limits the detailed discussion of limitations is provided in the Supplementary Material.

\paragraph{AP and LocScore in comparison.}
In practice, AP can be high while LocScore is low, and this divergence is common when localization is imprecise. Figure~\ref{fig:comparison_ap_locscore} shows two queries that both reach AP equal to one, yet their LocScores differ widely. The reason is that AP evaluates ranking only, whereas LocScore couples ranking with spatial alignment by multiplying the usual rank contribution with the intersection over union between zero and one. Therefore LocScore is bounded above by AP and it is usually smaller. Moreover LocScore increases only when relevant images appear early and the predicted region on those images has high overlap. If early results align poorly the score remains low even when AP is perfect. When objects are extremely small the grid can cap the attainable overlap, and in that regime, a thresholded variant becomes useful.

\begin{figure}[!t]
    \centering
      \includegraphics[width=\linewidth]{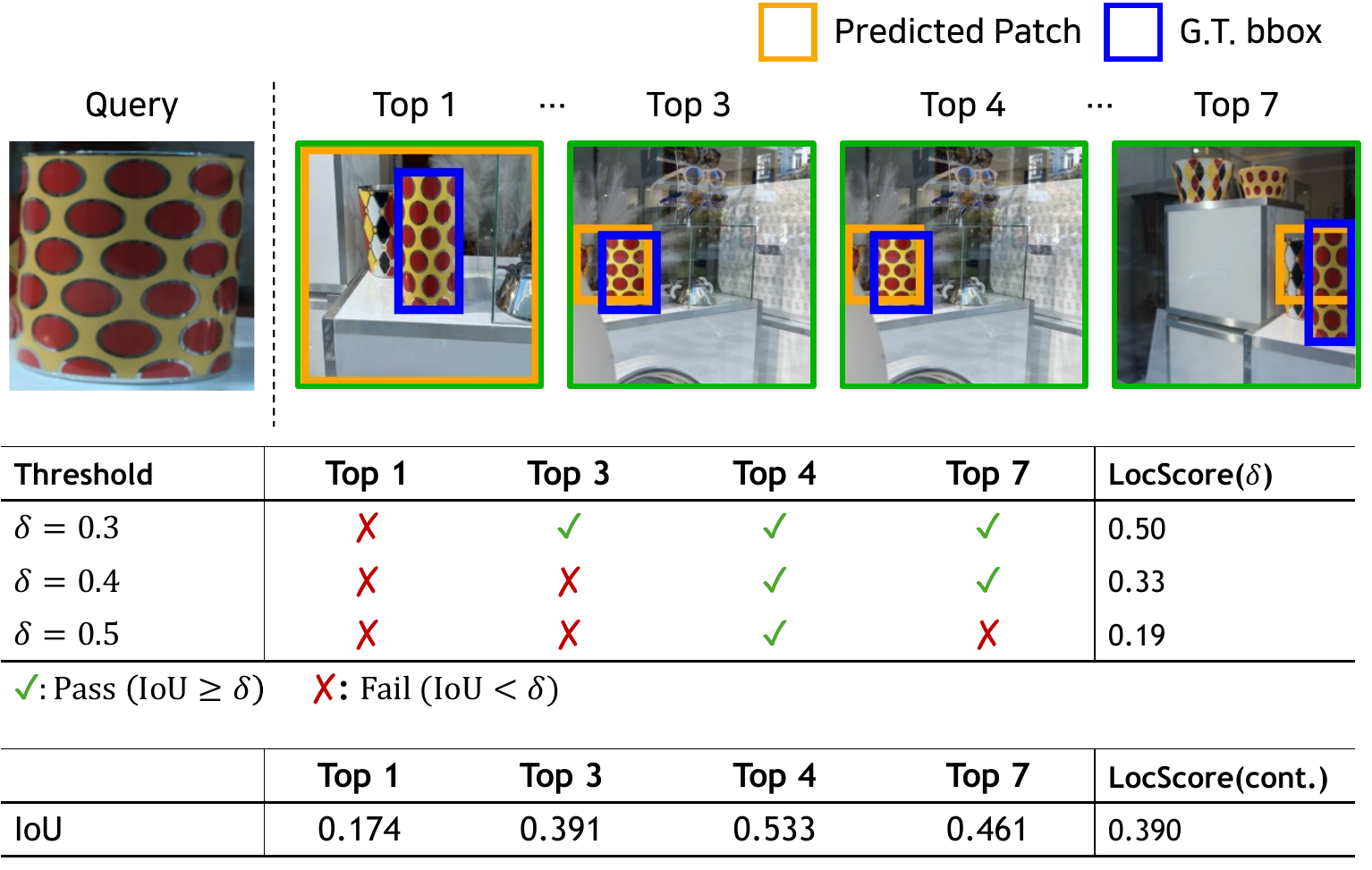}
    \par\vspace{1em}
    \caption{\textbf{LocScore variants comparison.}
    We visualize how the proposed LocScore metric changes as the IoU threshold $\delta$ varies and how it is different from the continuous variant. For each threshold, retrieved results are marked as correct if the predicted patch sufficiently overlaps with the ground-truth box (IoU $\geq \delta$). However, Continuous LocScore strictly evaluates the spatial precision based on IoU.
    }
    \label{fig:comparison_locscore_thr}
\end{figure}

\begin{figure}[!t]
  \centering
  \includegraphics[width=\linewidth]{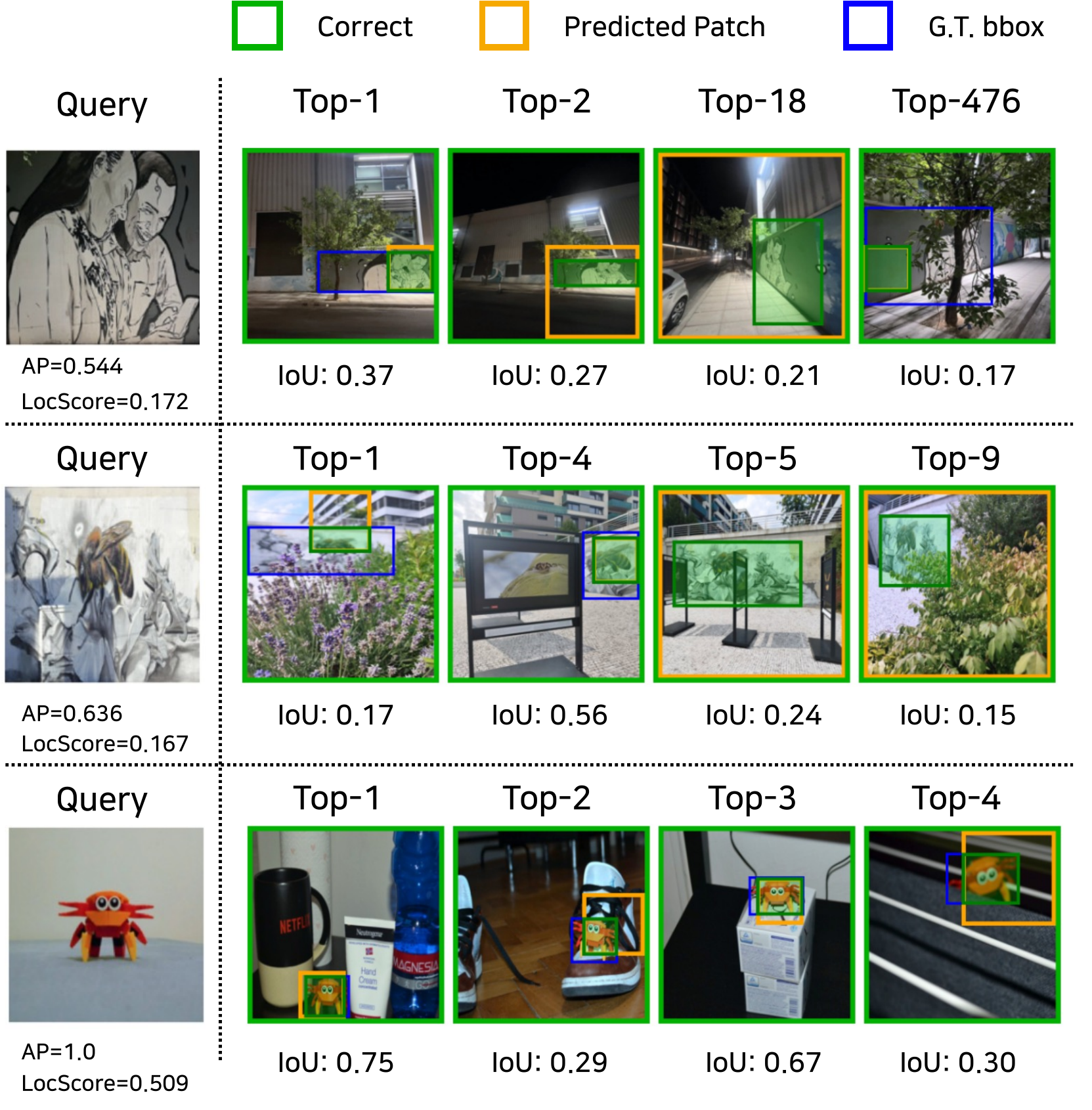}
  \caption{\textbf{Case study of LocScore.} Early high-overlap matches raise the score, while late matches or patch granularity limit it.}
  \label{fig:locscore_in_practice}
\end{figure}

\paragraph{Variants of LocScore.}
Some applications prefer a graded notion of localization, whereas others require a clear pass or fail decision. Figure~\ref{fig:comparison_locscore_thr} illustrates the thresholded form. The continuous form weights each positive by its measured overlap and rewards how well the triggering region aligns. In contrast the thresholded form at a chosen delta counts a positive only when the overlap is at least that delta and ignores partial matches. As delta increases, more results fail the test, and the score decreases. When a dataset contains many tiny targets, a small but positive delta, such as 0.09, can better reflect practical acceptability and it can make the two rows in Figure~\ref{fig:comparison_ap_locscore} score similarly.

\paragraph{What drives LocScore.}
Early matches with high overlap lift LocScore more than late, accurate boxes, and patch size and position often limit the attainable overlap. Figure~\ref{fig:locscore_in_practice} visualizes these effects. The top row gains a higher score because early positives already align well. The middle row ranks competitively, yet the target is large and horizontally elongated, so fixed grid patches cover it poorly, and the score drops. The bottom row shows both strong ranking and reasonable alignment, however the score remains around 0.5 because patch granularity and placement cap the overlap. Therefore, using a finer grid, adding mild overlap between patches, or selecting more suitable regions tends to raise LocScore more effectively than it raises AP. We empirically verify these properties in the following section ~\ref{subsec:diff_region_selection}.

\subsection{Method Comparison}
In this section, we investigate how different region selection strategies affect instance retrieval performance, and compare our Patchify method against recent state-of-the-art approaches, both as a standalone feature extractor and within reranking pipelines.

\begin{table}[!ht]
\centering
\caption{Comparison of different localization strategies in terms of mAP (\%) and LocScore (\%) on INSTRE and ILIAS.}
\resizebox{1.0\linewidth}{!}{ 
\begin{tabular}{lcccc}
\toprule
\multirow{2}{*}{\textbf{Method}} 
& \multicolumn{2}{c}{\textbf{INSTRE}} 
& \multicolumn{2}{c}{\textbf{ILIAS}} \\
\cmidrule(lr){2-3} \cmidrule(lr){4-5}
& \textbf{mAP} & \textbf{LocScore} & \textbf{mAP} & \textbf{LocScore} \\
\midrule
Patchify & 87.01 & 24.29 & 65.16 & 19.75 \\
\cmidrule{1-5}
Sliding Window (0.5) & 89.26 & 27.50 & 68.24 & 22.36 \\
Sliding Window (0.25) & 90.62 & 29.87 & 70.02 & 24.57 \\
\cmidrule{1-5}
Region Proposal & 92.96 & 71.44 & 84.19 & 68.23 \\
\bottomrule
\end{tabular}
}
\label{tab:localization}
\end{table}

\begin{figure*}[!t]
    \centering
    \includegraphics[width=0.85\textwidth, height=0.4\textheight, keepaspectratio]{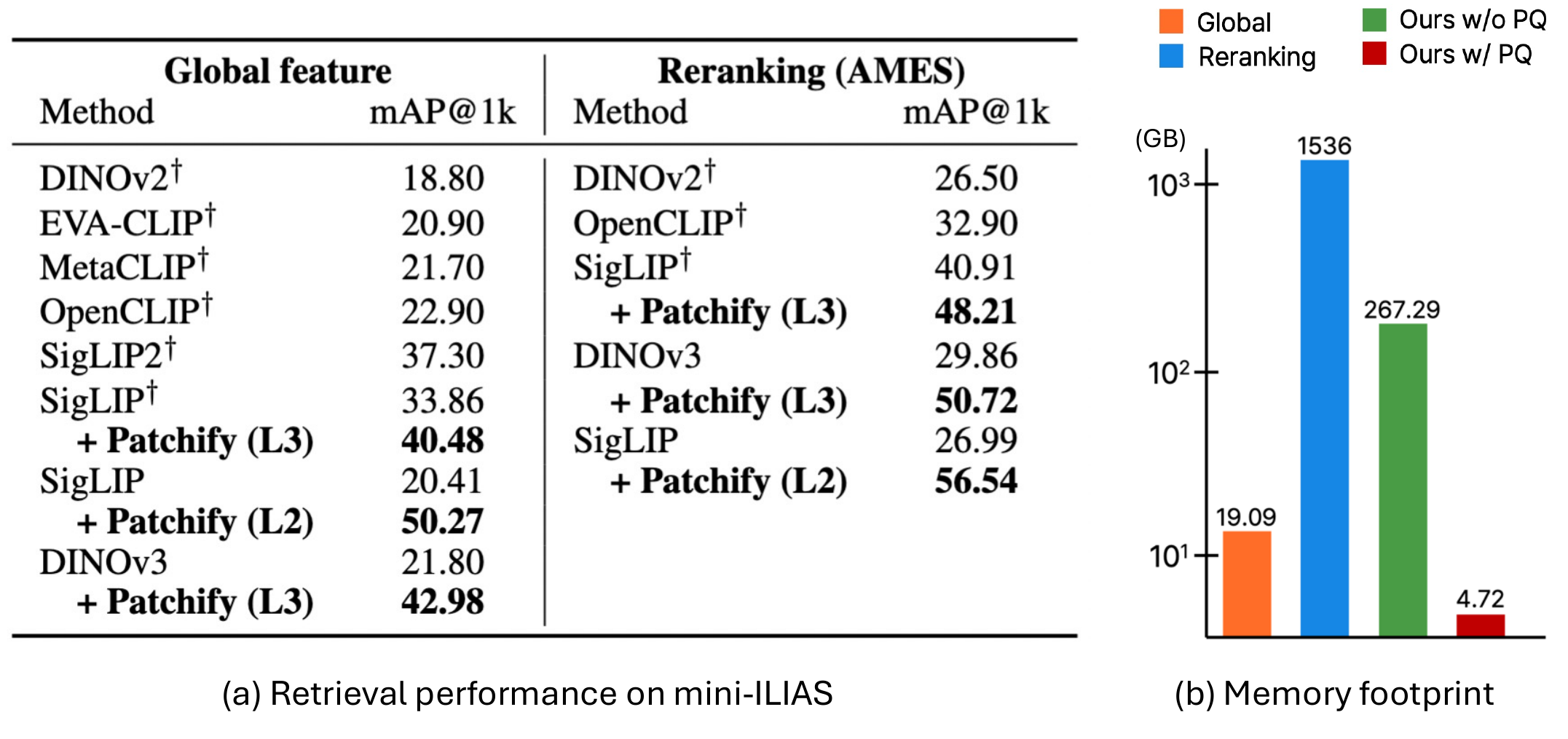}
    \caption{\textbf{Performance on mini-ILIAS in terms of mAP@1k and DB memory.} A dagger ($^\dagger$) denotes the linear adaptation baseline from the ILIAS~\cite{tolias2025ilias}, where the SigLIP encoder is fine-tuned on UnED~\cite{Ypsilantis_2023_ICCV} with a linear probing strategy. Patchify markedly improves over the global baselines, reaching or even surpassing the reranking performance of AMES.}
    \label{fig:mini_ilias_joint}
\end{figure*}

\subsubsection{Different region selection strategies}
\label{subsec:diff_region_selection}

We investigate how different region selection strategies influence instance retrieval, comparing our grid-based Patchify approach with sliding windows and region proposal methods using strong encoders such as SigLIP. 

This analysis provides a comprehensive perspective on how various region-level sampling strategies influence instance retrieval performance. By examining the trade-offs between spatial structure, semantic precision, and coverage, we aim to reveal the strengths and limitations of simple grid-based methods in comparison to more fine-grained or learned approaches. Detailed descriptions of each approach are provided in Supplementary Material (Section~\ref{sec:region_strategies}).  

As shown in Table~\ref{tab:localization}, we observe that sliding window methods outperform Patchify in both mAP and LocScore, with the 0.25 stride variant achieving the best results. Overall, higher mAP tends to coincide with higher LocScore, indicating a positive correlation between retrieval accuracy and spatial alignment. These trends hold consistently across architectures, where increasing patch granularity enhances performance for both CNNs and Transformers. These consistent patterns suggest that better retrieval models not only retrieve the correct images but also localize the target objects more precisely. However, these benefits come at the cost of additional computation and memory, whereas Patchify maintains competitive performance with much greater efficiency. 


\subsubsection{Comparison with SOTA Methods}
\label{sec:sota_comparison}
To contextualize the effectiveness of Patchify within the broader instance retrieval landscape, we evaluate it against current state-of-the-art methods on the mini-ILIAS~\cite{tolias2025ilias}, which augments the ILIAS core with hard negatives from YFCC100M~\cite{thomee2016yfcc100m} and is empractically harder than random distractors~\cite{tolias2025ilias}. We divide our analysis into two retrieval settings: single-stage retrieval based on global representations, and two-stage reranking.

\paragraph{Single-Stage Retrieval with Global Representations.} We report results for global feature baselines, SigLIP linear adaptation from the ILIAS paper, and the powerful backbone DINOv3~\cite{siméoni2025dinov3}. For numerical comparison, we cite values trained under the same setting from ILIAS~\cite{tolias2025ilias}, shown above the double rule, while the results below are from our own experiments. As shown in Figure~\ref{fig:mini_ilias_joint} (a), Patchify achieves stronger performance than all the baselines. These findings underscore the strength of our Patchify method, which delivers competitive performance in a fully zero-shot setting without requiring any fine-tuning. Furthermore, the use of a stronger backbone with Patchify leads to additional performance gains, suggesting that our patch-based framework benefits directly from improvements in underlying visual encoders.

\paragraph{Integration into Reranking Frameworks.}
We further evaluate how Patchify integrates into two-stage retrieval pipelines by combining it with AMES~\cite{suma2024ames}, a state-of-the-art reranking method based on local descriptor alignment. In this experiment, we don't consider any other recent reranking baselines since they all rely on first-stage retrieval, which our Patchify can serve as. This comparison allows us to assess whether Patchify can serve not only as a standalone retrieval strategy, but also as an effective feature representation within existing reranking architectures.
As shown in Figure~\ref{fig:mini_ilias_joint} (a), Patchify demonstrates strong compatibility with reranking architectures, achieving superior performance when integrated into AMES compared to prior state-of-the-art models. These results indicate that Patchify not only excels as a standalone retrieval mechanism but also functions effectively as the first-stage representation in reranking systems. Surprisingly, Patchify markedly improves over the global baselines, reaching or even surpassing the reranking performance.

\paragraph{Memory Footprint.}
We also compare the memory requirements of Patchify against other baselines. To directly contrast Patchify with reranking methods, we report the storage needed to encode and save the entire database at each stage. 
Note that for reranking, this does not include the additional storage required for global feature retrieval. As shown in Figure~\ref{fig:mini_ilias_joint} (b), Patchify is far more memory-efficient, using 5.75$\times$ less storage (267 vs. 1536 GB). This indicates that even while storing only a handful of patch descriptors per image, Patchify achieves strong performance, highlighting its effectiveness. 
\subsection{Applying Compression to Local Features}

A well-known limitation of local feature-based retrieval methods is their high memory usage, as they require storing a large number of descriptors per image. Unlike conventional approaches, Patchify uses only a small number of structured patch features per image, significantly reducing the memory and computational burden. To further enhance scalability for large-scale retrieval, we apply Inverted File with Product Quantization (IVFPQ)~\citep{PQ}. For more detailed experimental setup, please see Section~\ref{sec:pq}.

\begin{table}[t]
\centering
\small
\caption{Comparison of mAP (\%) performance under different multi-scale levels and PQ usage.}
\resizebox{\linewidth}{!}
{%
\begin{tabular}{ccccccc}
\toprule
\multirow{2}[2]{*}{\textbf{Level}} & \multirow{2}[2]{*}{\textbf{Patchify}} & \multirow{2}[2]{*}{\textbf{PQ}} & \multicolumn{2}{c}{\textbf{INSTRE}} & \multicolumn{2}{c}{\textbf{ILIAS}} \\
\cmidrule(rl){4-5}\cmidrule(rl){6-7}
&& & \textbf{mAP} & \textbf{LocScore} & \textbf{mAP} & \textbf{LocScore} \\
\midrule
\multirow{2}{*}{L0} 
& \textcolor{red}{\xmark} & \textcolor{red}{\xmark} & 78.48 & 18.92 & 55.03 & 14.31 \\ 
& \textcolor{red}{\xmark} & \textcolor{OliveGreen}{\cmark} & 81.99 & 19.48 & 54.97 & 14.30 \\ 
\cmidrule{1-7}
\multirow{2}{*}{L1} 
& \textcolor{OliveGreen}{\cmark} & \textcolor{red}{\xmark} & 83.06 & 21.72 & 59.45 & 16.63 \\ 
& \textcolor{OliveGreen}{\cmark} & \textcolor{OliveGreen}{\cmark} & 86.20 & 22.92 & 57.37 & 15.65 \\ 
\cmidrule{1-7}
\multirow{2}{*}{L2} 
& \textcolor{OliveGreen}{\cmark} & \textcolor{red}{\xmark} & 85.97 & 23.89 & 63.32 & 18.84 \\ 
& \textcolor{OliveGreen}{\cmark} & \textcolor{OliveGreen}{\cmark} & 88.20 & 25.00 & 59.28 & 17.00 \\ 
\cmidrule{1-7}
\multirow{2}{*}{L3} 
& \textcolor{OliveGreen}{\cmark} & \textcolor{red}{\xmark} & 87.01 & 24.29 & 65.16 & 19.75 \\ 
& \textcolor{OliveGreen}{\cmark} & \textcolor{OliveGreen}{\cmark} & 88.52 & 25.08 & 59.96 & 17.33 \\ 
\bottomrule
\end{tabular}%
}
\label{tab:multiscale_pq}
\end{table}

\paragraph{Performance comparison.}
\Cref{tab:multiscale_pq} presents the retrieval performance before and after applying PQ. As expected, compressing feature representations with PQ can lead to a slight performance drop due to quantization. This trend is observed on the ILIAS benchmark, where the data scale and difficulty are substantially higher. Interestingly, on the smaller and less challenging INSTRE benchmark, PQ even yields performance improvements, potentially due to noise reduction or enhanced centroid generalization. Despite compression, local features with PQ still outperform global features in both benchmarks. These results indicate that local characteristics and PQ offer complementary benefits: delivering strong retrieval accuracy with substantially reduced storage.

\begin{table}[t]
\centering
\caption{Comparison of mAP (\%) performance using different training features for PQ.}
\resizebox{\columnwidth}{!}{%
\begin{tabular}{lccccc}
\toprule
\textbf{Training Feature} & L0 (1) & L1 (5) & L2 (14) & L3 (30) & G.T. (1) \\
\midrule
mAP (\%) & 60.19 & 55.80 & 55.53 & 53.13 & \textbf{65.07} \\
\bottomrule
\end{tabular}
}
\label{tab:pq_training_level_comparison}
\end{table}

\paragraph{Training with Informative Features.}
Product Quantization (PQ) requires a training phase to learn cluster centroids, and the choice of training features can significantly impact retrieval performance. To explore this, we compare PQ trained with features at different patch levels (L0, L1, L2, L3) and with ground-truth-aligned features cropped from instance bounding boxes. As summarized in Table~\ref{tab:pq_training_level_comparison}, the highest performance is achieved when PQ is trained on ground-truth features, emphasizing the value of using semantically informative representations. Interestingly, features from intermediate patch levels (L1 and L2) perform worse than L0, likely due to the inclusion of background noise or irrelevant content. These results suggest that tighter spatial alignment during PQ training plays a crucial role, and highlight the importance of effective region selection strategies. This highlights a previously underexplored yet critical aspect of PQ training and may inform more robust design choices in future PQ-based retrieval pipelines. Additional qualitative comparisons and visualizations of PQ training features are provided in Section~\ref{sec:pq}.

\section{Conclusion}
In this work, we introduced Patchify, a simple yet effective patch-wise retrieval framework that achieves strong performance while maintaining both scalability and interpretability. By decomposing each database image into a small set of structured, multi-scale patches, Patchify enables spatially-aware matching without requiring fine-tuning or region proposals. Our experiments demonstrate that Patchify consistently outperforms global feature baselines, adapts well across a variety of backbones, and integrates seamlessly into existing reranking pipelines.
To evaluate the spatial quality of retrievals, we proposed LocScore, a localization-aware metric that captures not only retrieval accuracy but also the alignment between retrieved regions and target objects. This metric provides valuable diagnostic insight into how retrieval decisions are made and helps bridge performance evaluation with model interpretability.
Furthermore, we showed that Patchify is highly scalable through the use of PQ. Our analysis revealed that the choice of training features for PQ has a significant impact on performance. In particular, features that are well-aligned with target objects lead to better compression and retrieval accuracy, highlighting the importance of informative patch selection during index construction.
Overall, Patchify offers a practical and interpretable design for instance-level image retrieval, enabling efficient matching with minimal computation and memory overhead. Our findings point toward promising future directions for retrieval systems that combine spatial precision, scalability, and transparency.

\section*{Acknowledgment}
This work was supported by Electronics and Telecommunications Research Institute (ETRI) grant funded by the Korean government [25ZD1160, Development of ICT Convergence Technology for Daegu-Gyeongbuk Regional Industry], Institute of Information \& communications Technology Planning \& Evaluation (IITP) grant funded by the Korea government(MSIT) (No. 2022-0-00124, No.RS-2022-II220124, Development of Artificial Intelligence Technology for Self-Improving Competency-Aware Learning Capabilities; No.RS-2023-00225630, Development of Artificial Intelligence for Text-based 3D Movie Generation; No.RS-2019-II191906, Artificial Intelligence Graduate School Program(POSTECH))
{
    \small
    \bibliographystyle{ieeenat_fullname}
    \bibliography{main}
}

\clearpage
\setcounter{page}{1}
\setcounter{section}{0} 
\renewcommand{\thesection}{S\arabic{section}} 
\renewcommand{\thefigure}{S\arabic{figure}}
\setcounter{figure}{0}
\maketitlesupplementary

\section{Overview}

This supplementary material provides additional experiments and implementation details that could not be included in the main paper due to space constraints. Following the structure of the main text, we elaborate on the following aspects:

\begin{itemize}
    \item \textbf{Section~\ref{sec:img_char}} explains experimental details on the \textit{Impact of Image Characteristics}, such as instance size, position, and brightness, on retrieval performance.
    \item \textbf{Section~\ref{sec:sup_backbone}} expands on the \textit{Investigation of Visual Backbones}, comparing CNN and Transformer encoders and the effect of pretraining dataset size.
    \item \textbf{Section~\ref{sec:region_strategies}} details additional comparisons of \textit{Different Region Selection Strategies}, including Patchify, Sliding Window, and Region Proposal methods.
    \item \textbf{Section~\ref{sec:sota}} provides a deeper dive into our \textit{Comparison with SOTA Methods} in both single-stage and reranking settings.
    \item \textbf{Section~\ref{sec:pq}} includes extended analysis on \textit{Product Quantization}, covering memory efficiency, training feature selection, and compression-performance trade-offs.
    \item \textbf{Section~\ref{sec:sup_locscore}} further analyze and report thresholded version and average LocScore.
    \item \textbf{Section~\ref{sec:qual}} shows additional \textit{Qualitative Results} that highlight retrieval accuracy and localization behavior across diverse scenarios.
\end{itemize}

These materials are intended to support and complement the findings in the main paper, offering deeper insight into our proposed approach and its practical implications.


\begin{figure*}[!ht]
  \centering
  \begin{subfigure}[b]{1\linewidth}
    \centering
    \includegraphics[width=0.4\linewidth]{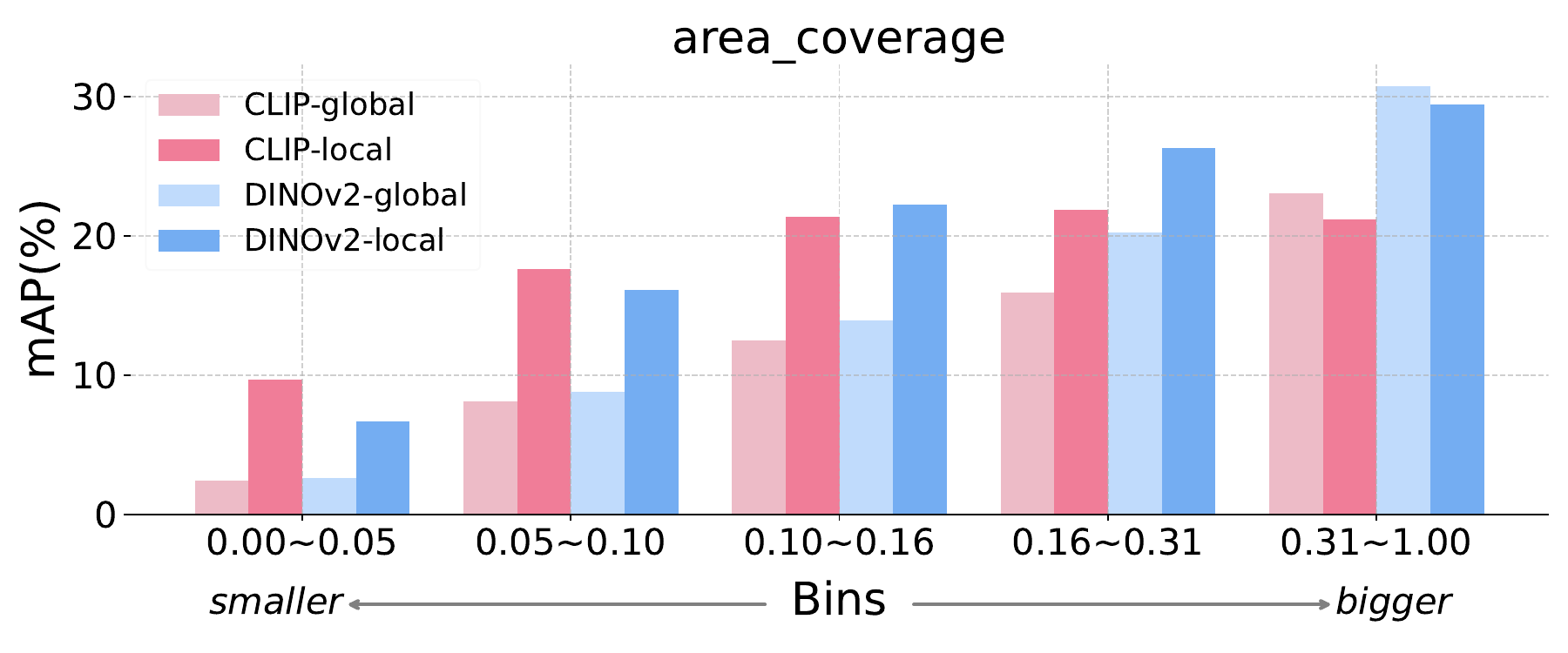}
    \includegraphics[width=0.4\linewidth]{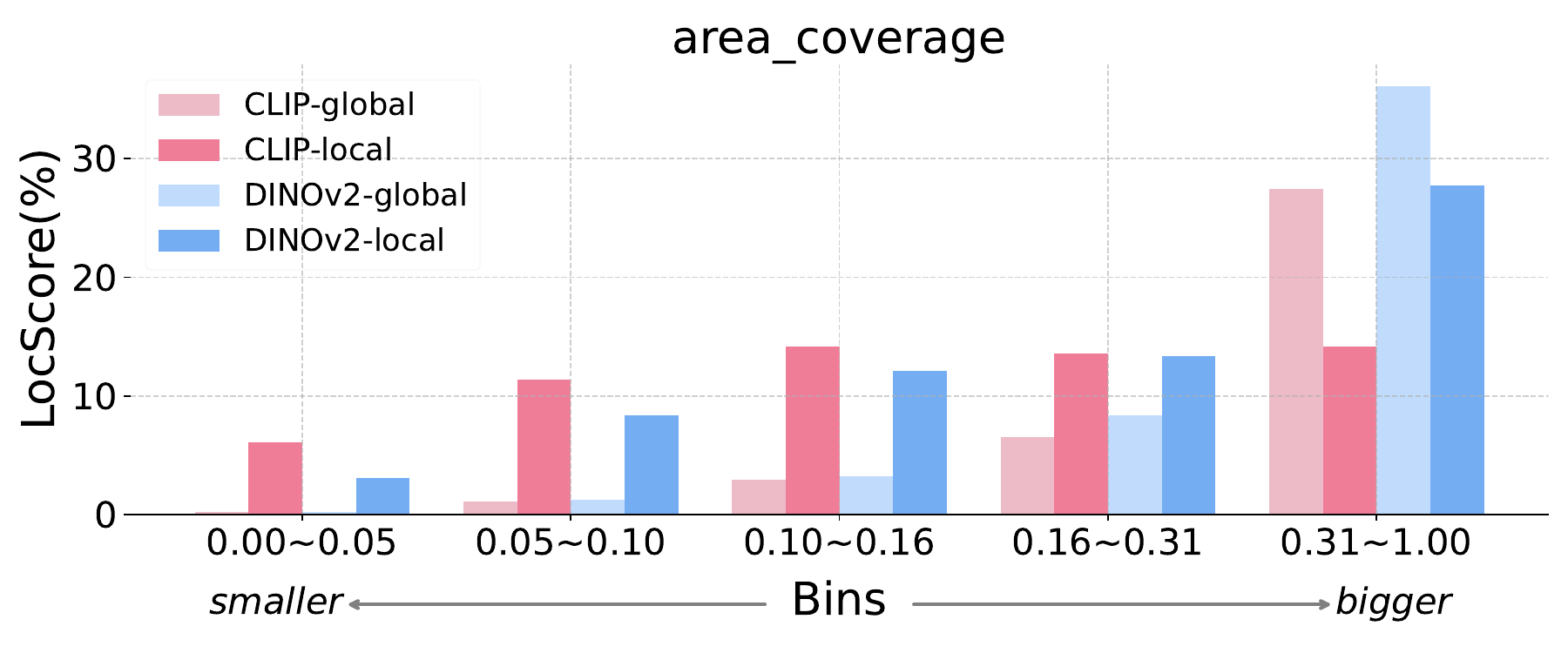}
    \caption{Object bounding box ratio}
    \label{fig:bounding_box_ratio}
  \end{subfigure}
  \hfill
  \begin{subfigure}[b]{1\linewidth}
    \centering
    \includegraphics[width=0.4\linewidth]{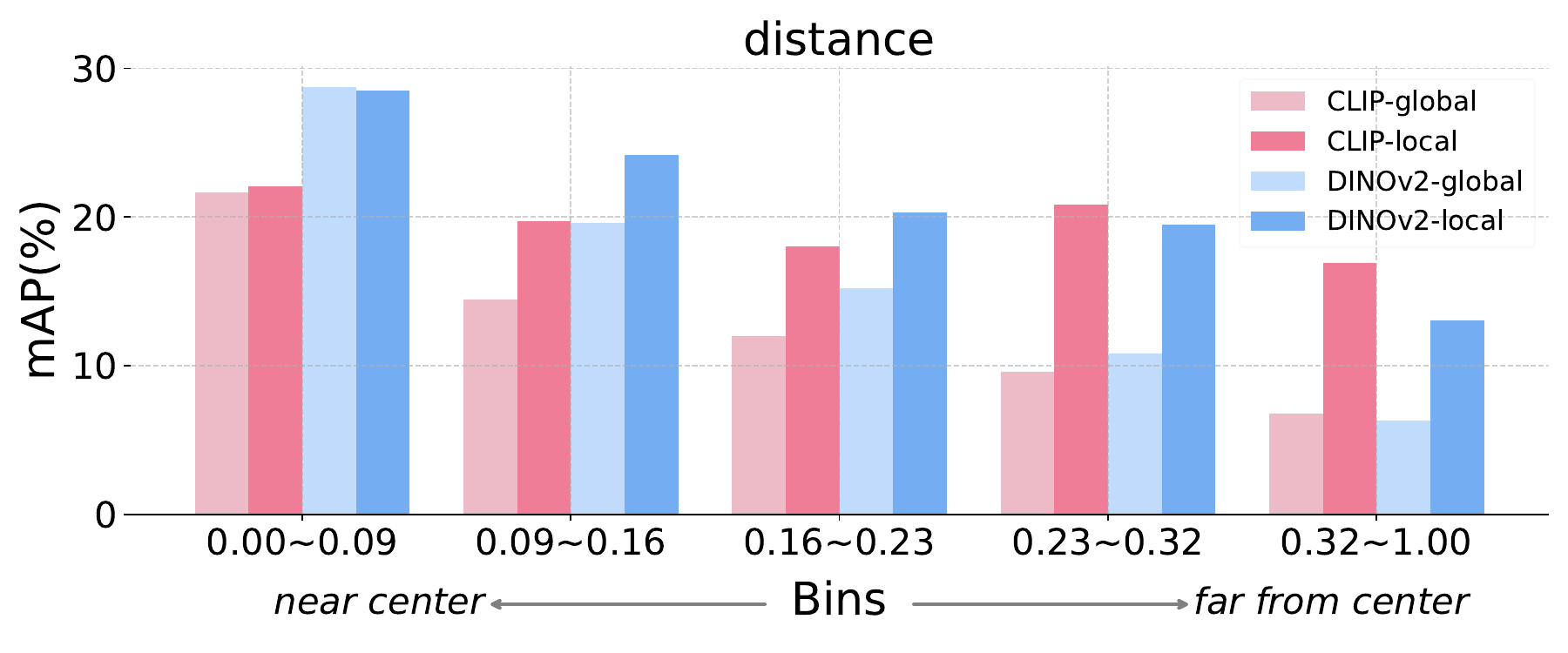}
    \includegraphics[width=0.4\linewidth]{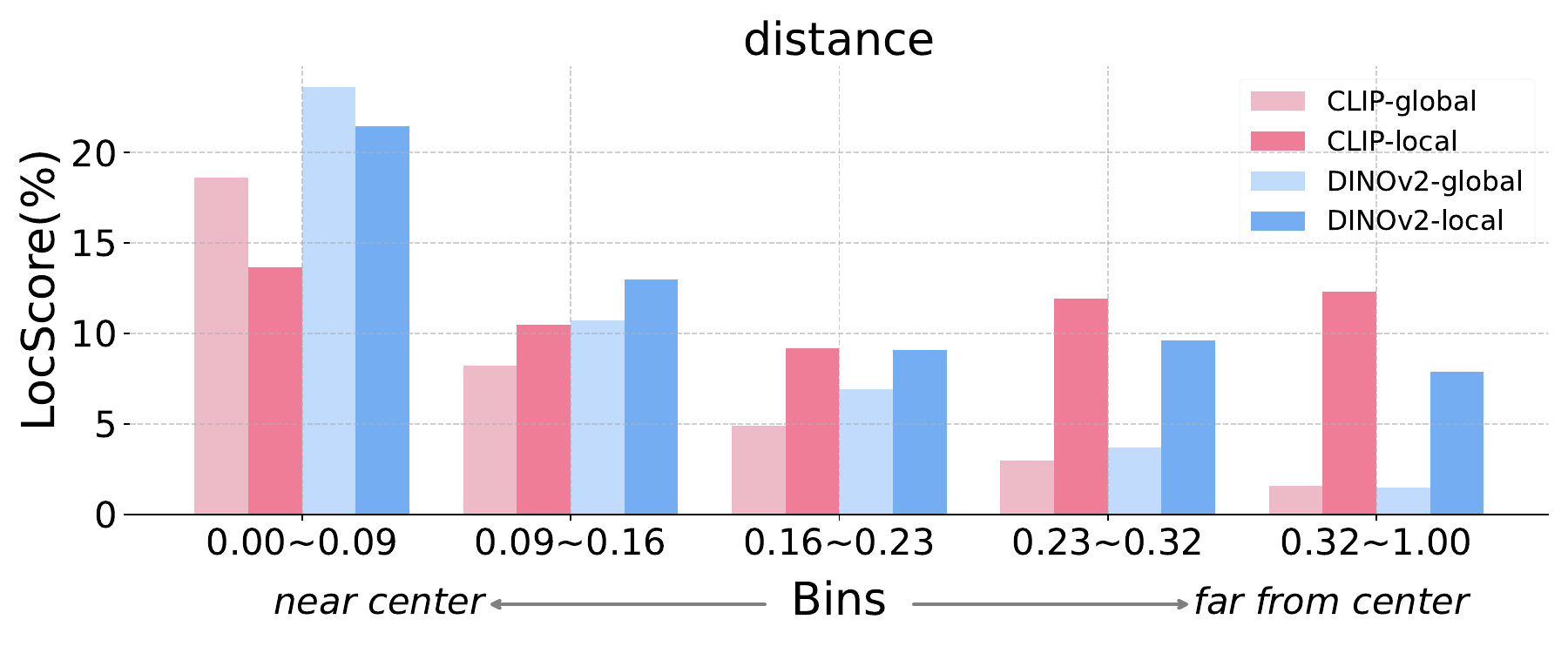}
    \caption{Object distance from image center}
    \label{fig:distance}
  \end{subfigure}
  \hfill
    \begin{subfigure}[b]{1\linewidth}
    \centering
    \includegraphics[width=0.4\linewidth]{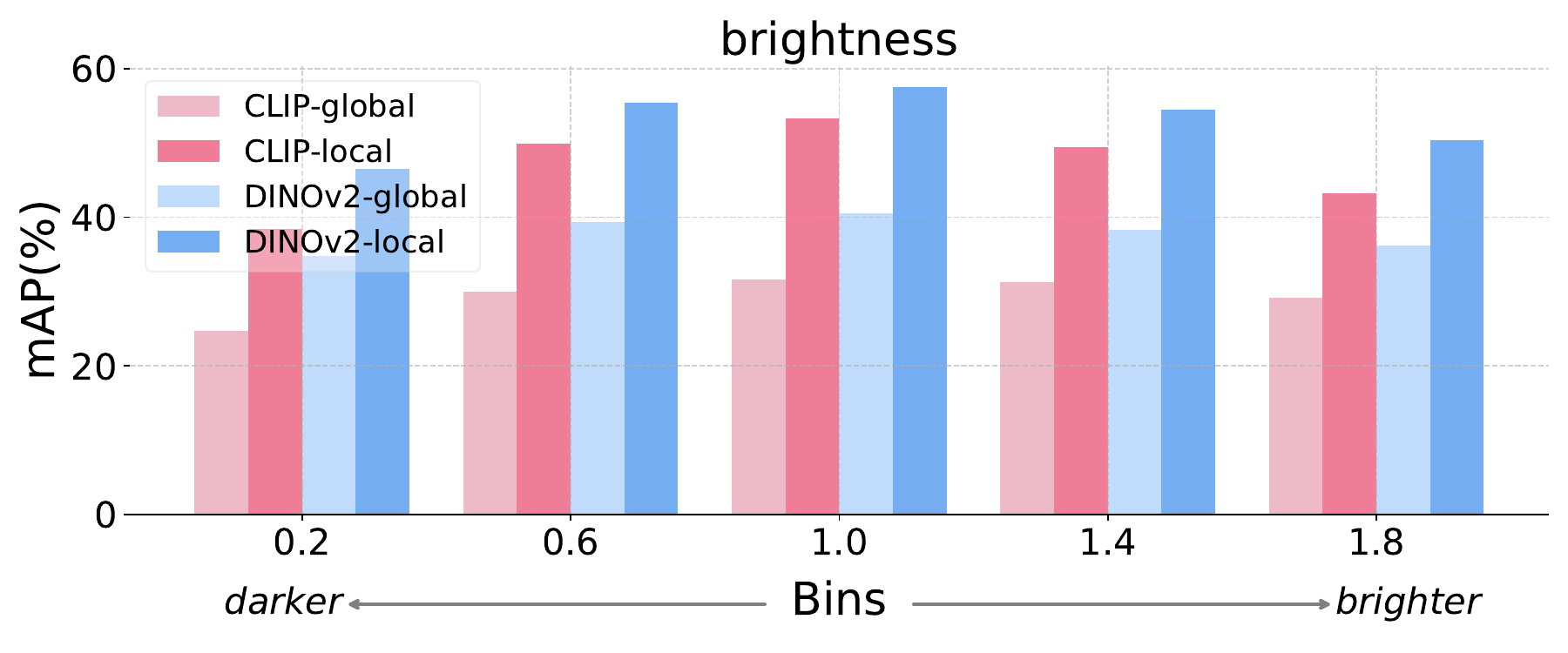}
    \includegraphics[width=0.4\linewidth]{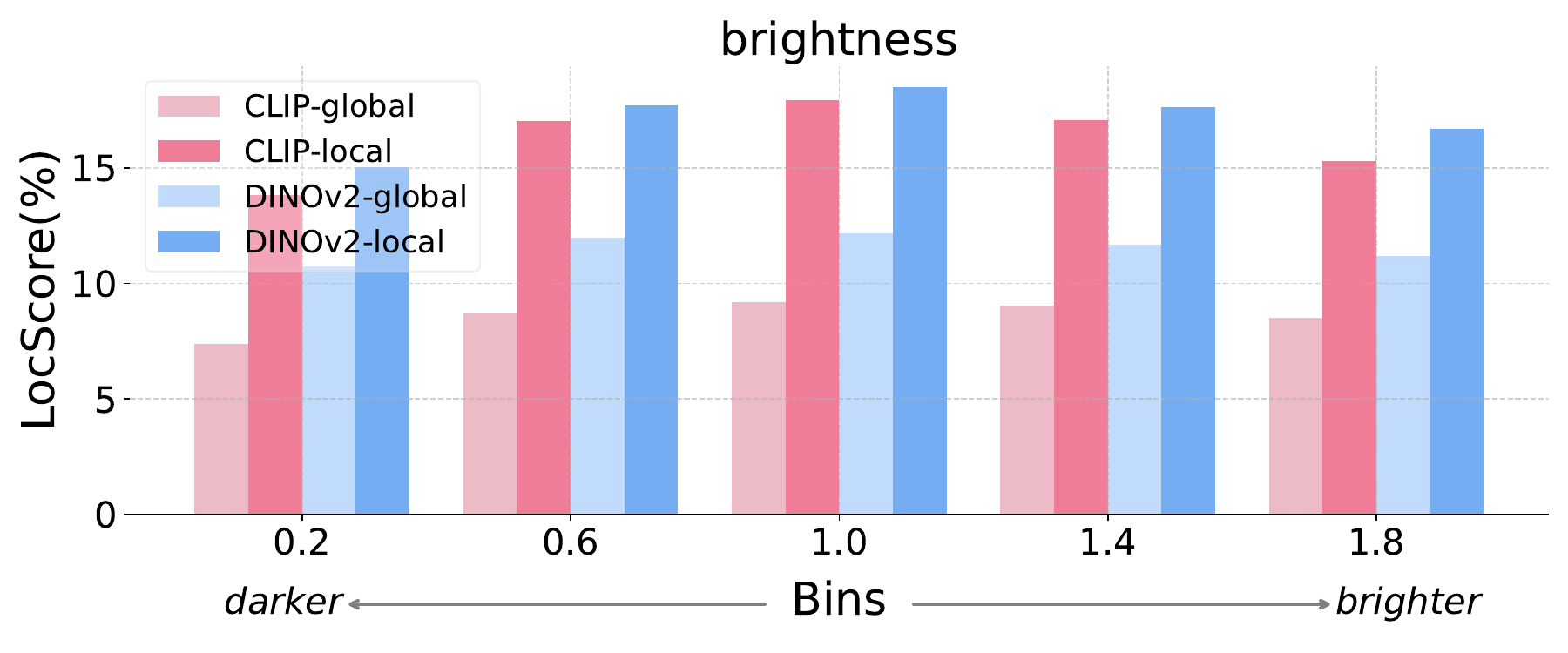}
    \caption{Brightness of image}
    \label{fig:brightness}
  \end{subfigure}
  \caption{mAP(\%) and LocScore(\%) of global and local features in terms of object size, distance from image center, and brightness of image. We measure the performance on ILIAS. Both global and local features show a similar trend. In most cases, local features consistently outperform global features across all settings.}
  \label{fig:data_config}
\end{figure*}

\section{Impact of Image characteristics}
\label{sec:img_char}
\subsection{Analysis}
We investigate the effectiveness of global and local features with respect to three factors: instance size, location, and brightness. The first two factors are directly related to our proposed metrics, LocScore; the last factor is a common challenge in instance retrieval.
Figure~\ref{fig:data_config} shows the comparison between local and global features. Overall, we observe that local features generally improve performance, demonstrating robustness under varying image conditions.

As shown in \Cref{fig:bounding_box_ratio}, global features outperform local features when the instance occupies a large portion of the image (rightmost group). However, as the instance size decreases, local features begin to outperform global ones, demonstrating their advantage in small object retrieval.
Next, we assess the effect of object location by measuring Euclidean distance between the center of the image and the center of the ground-truth bounding box.
As shown in Figure~\ref{fig:distance}, local features maintain more stable retrieval accuracy as the instance moves away from the center, demonstrating greater robustness to spatial displacement.

Lastly, we analyze the impact of image brightness, quantified as the average L-channel value in the Lab color space. 
In \Cref{fig:brightness}, both global and local features achieve higher performance on images with medium brightness, while performance drops on very dark or very bright images. 
Nevertheless, local features consistently outperform global features across all brightness levels, even though the overall trend is similar. For a detailed experimental setup, please refer to Section~\ref{sec:img_char}.

\subsection{Experimental Details}
We adopt CLIP and DINOv2 as the visual encoder for the evaluation. We compare global and local (L3) features with mAP and LocScore on ILIAS core set.
\paragraph{Object bounding box ratio}
We sort database images with object size by computing the object bounding box area. Then we divide the total object size values equally into 5 bins. For the evaluation, queries with no true positive images in a specific bin are excluded.
Global feature works well in big objects but not in small ones. 
\paragraph{Object distance from image center}
We compute L2 distance between the bounding box and the image center and sort them. 
Then we divide all values equally into 5 bins. Similarly, queries with no true positive images in a specific bin are excluded.
Global features work well when the object is centered and not so well when it is not centered. 

\paragraph{Brightness of image}
Unlike the previous two experiments, the evaluation benchmark datasets did not exhibit a meaningful distribution with respect to brightness. To address this, we applied tone mapping by scaling the L channel in the Lab color space to 0.2×, 0.6×, 1.0×, 1.4×, and 1.8× of its original value, creating five brightness-adjusted versions of each dataset. We then measured performance on each transformed set.

\section{Investigation of visual backbone}
\label{sec:sup_backbone}
Recently, many visual encoders have been proposed, and we have to choose which visual encoder we use. In this section, we further explore which visual encoders offer the most effective representations for instance retrieval. We analyze a diverse set of models, including both CNN-based (e.g., VGG~\citep{simonyan2015deepconvolutionalnetworkslargescale}, ResNet~\cite{he2015deepresiduallearningimage}, Inception~\cite{szegedy2014goingdeeperconvolutions}, ConvNext~\cite{liu2022convnet}) and Transformer-based (e.g., DINOv2, CLIP, SigLIP~\citep{zhai2023sigmoid}) encoders. 
This analysis aims to provide deeper insights into the architectural factors that contribute to strong instance-level retrieval performance. Note that we extract local features using Patchify. Throughout all experiments in this paper, unless otherwise specified, we use the Large model configuration and an input image resolution of $384 \times 384$.

\begin{figure*}[t!]
  \centering
  \begin{subfigure}[t]{0.9\linewidth}
    \centering
    \includegraphics[width=0.45\linewidth]{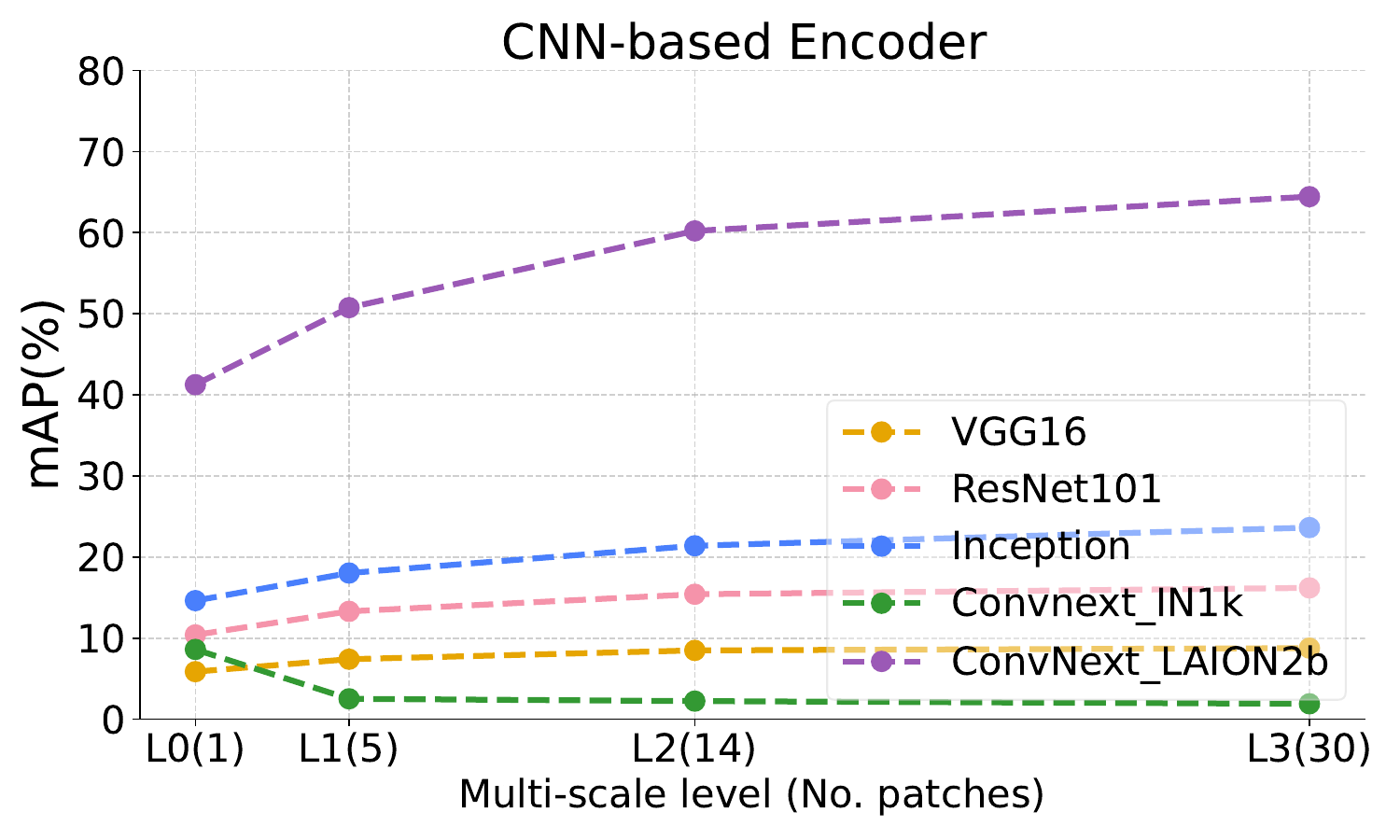}
    \includegraphics[width=0.45\linewidth]{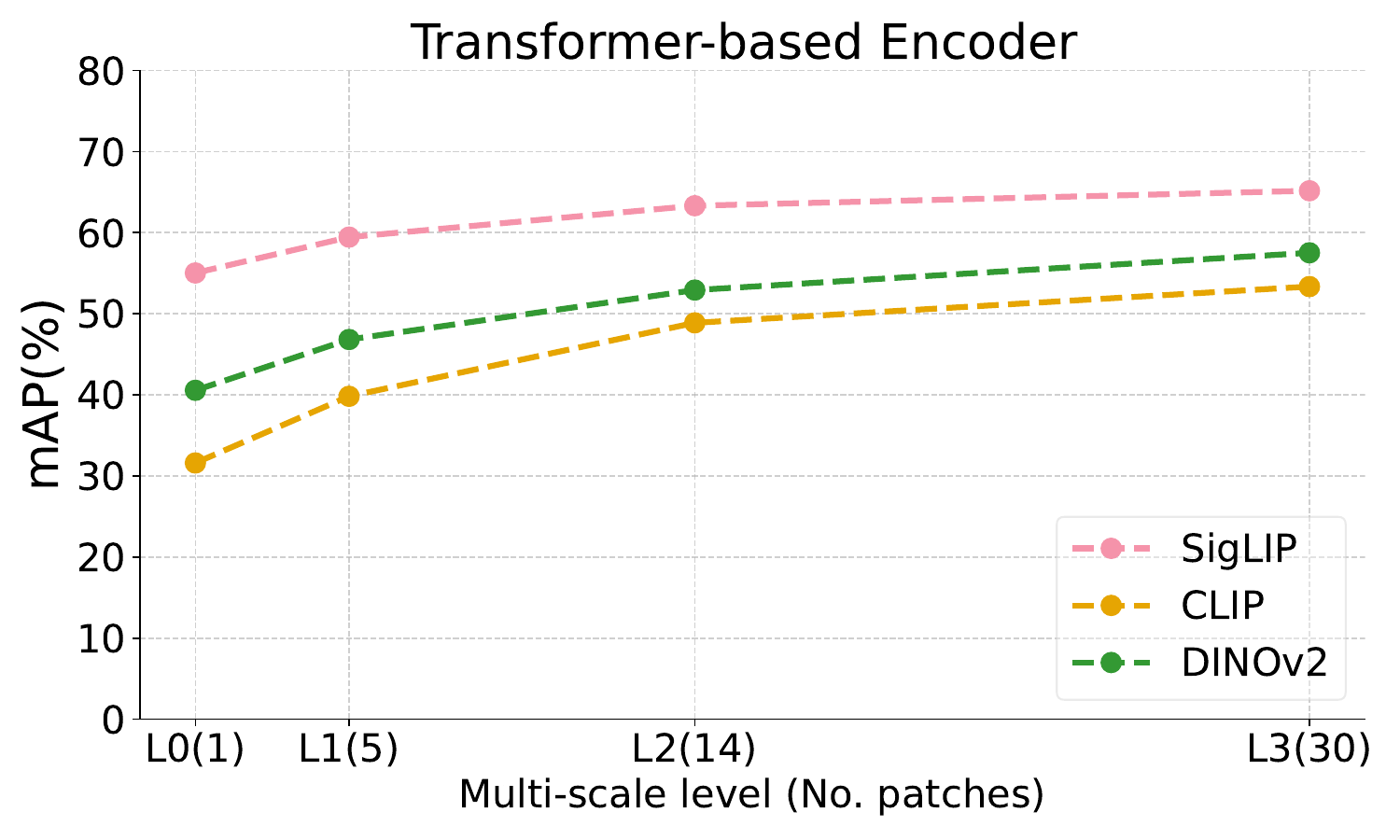}
    \caption{mAP}
    \label{fig:ilias_map_cnn_transformer}
  \end{subfigure}
  \vspace{0.5em} 
  \begin{subfigure}[t]{0.9\linewidth}
    \centering
    \includegraphics[width=0.45\linewidth]{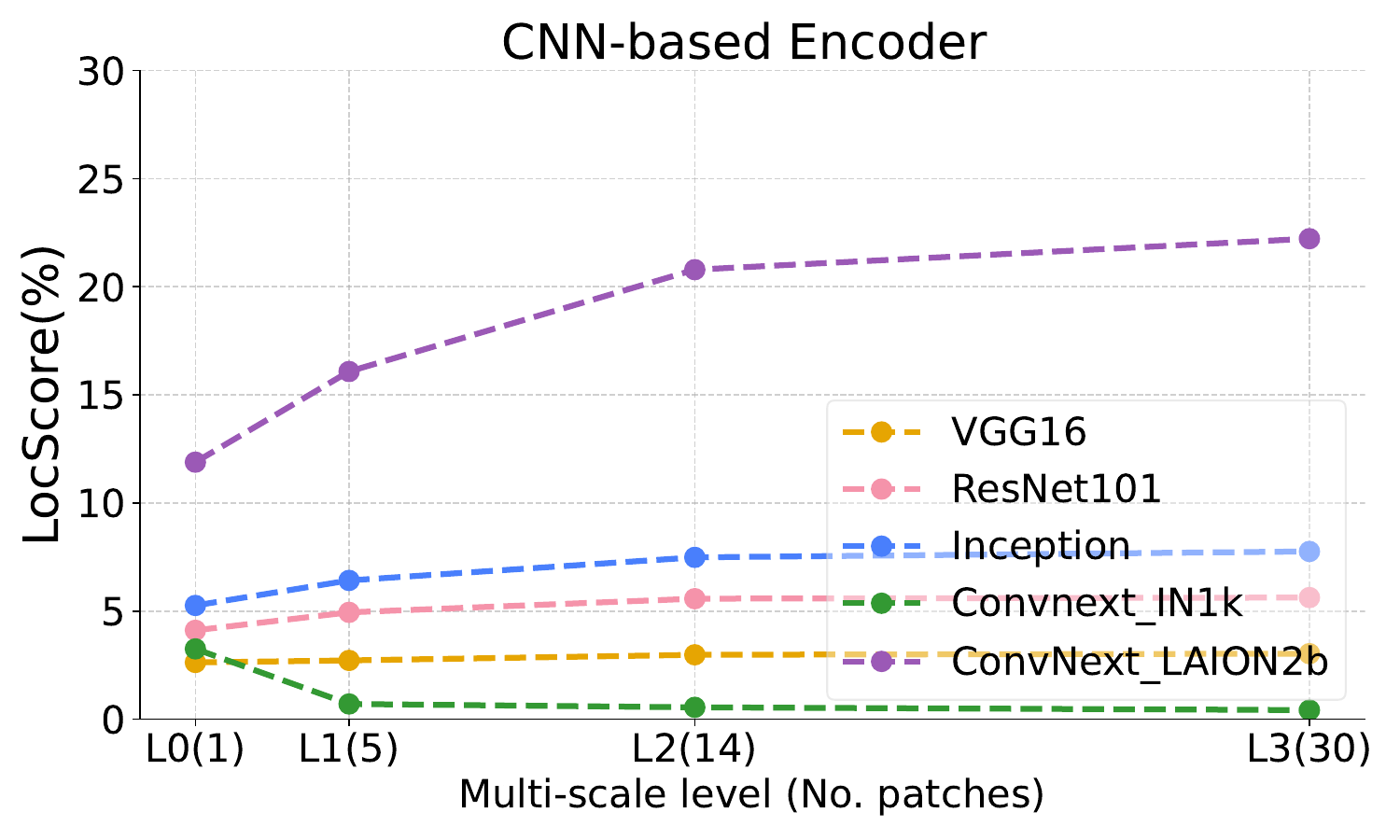}
    \includegraphics[width=0.45\linewidth]{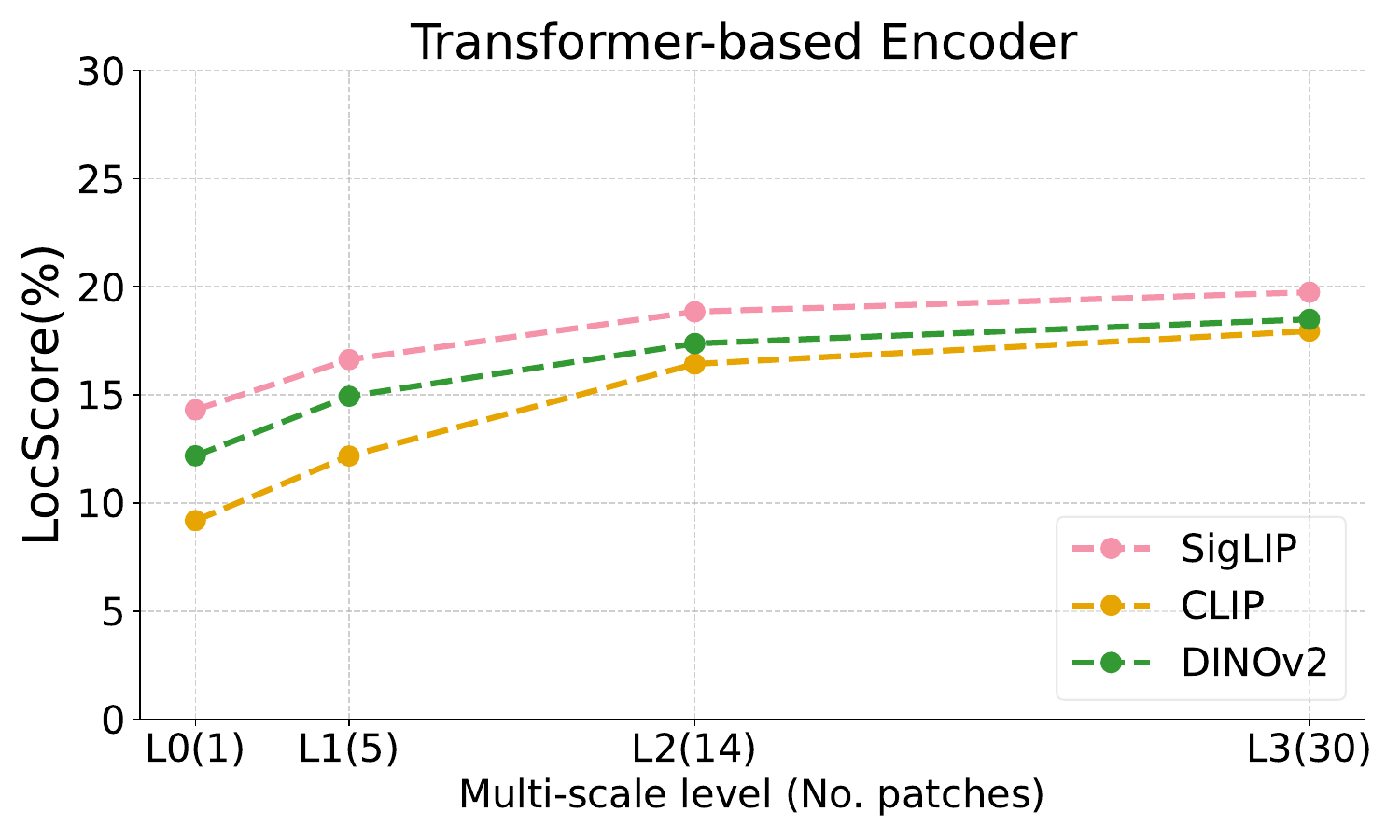}
    \caption{LocScore}
    \label{fig:ilias_lmap_cnn_transformer}
  \end{subfigure}
  \caption{Performance comparison between CNN-based backbone and Transformer-based models based on (a) mAP (\%) and (b) LocScore (\%) metrics on ILIAS. For comparison, we apply the same scale to plots of CNN-based and Transformer-based models. We observe that transformer-based models show higher performance compared to CNN-based models, except for ConvNext\_laion2b.}
  \label{fig:multiscale_backbone}
\end{figure*}

\paragraph{Characteristic of visual encoders}
We compare instance retrieval performance across encoder models with CNN-based and Transformer-based backbones. As shown in \Cref{fig:multiscale_backbone}, based on our analysis using the ILIAS core dataset, as we use more local features, the performance generally increases. ConvNeXt trained on ImageNet is an exceptional case where the performance decreases. Although ConvNeXt pretrained on LAION-2B shows high performance in CNN-based models, transformer-based models generally achieve higher performance compared to CNN-based models.

\begin{table}[t]
\centering
\caption{Impact of pretraining data size on instance retrieval performance of encoders. }
\resizebox{\linewidth}{!}{
\begin{tabular}{cllcc}
\toprule
 \multirow{2}[2]{*}{\textbf{\makecell{Dataset \\ (Size)}}} & \multirow{2}[2]{*}{\textbf{\makecell[l]{Encoder \\ (Feat. Dim.)}}} & \multirow{2}[2]{*}{\textbf{Type}} & \multicolumn{2}{c}{\textbf{mAP}} \\
\cmidrule(lr){4-5}
& & & \textbf{INSTRE} & \textbf{ILIAS} \\
\midrule
\multirow{9}{*}{\makecell{ImageNet-1K\\(1M)}} 
& \multirow{2}{*}{\makecell[l]{VGG16\\(4096)}} 
& global & 26.90 & 5.86 \\
& & local & 53.56 & 8.76 \\
\cmidrule(lr){2-5}
& \multirow{2}{*}{\makecell[l]{ResNet101\\(2048)}} 
& global & 36.29 & 10.40 \\
& & local & 61.38 & 16.19 \\
\cmidrule(lr){2-5}
& \multirow{2}{*}{\makecell[l]{Inception\\(1536)}} 
& global & 27.94 & 14.62 \\
& & local & 45.82 & 23.63 \\
\cmidrule(lr){2-5}
& \multirow{2}{*}{\makecell[l]{ConvNext\\(1024)}} 
& global & 38.97 & 8.61 \\
& & local & 56.25 & 1.90 \\
\cmidrule(lr){1-5}
\multirow{2}{*}{\makecell{LVD-142M\\(142M)}}
& \multirow{2}{*}{\makecell[l]{DINOv2\\(1024)}} 
& global & 57.70 & 40.56 \\
& & local & 72.54 & 57.51 \\
\cmidrule(lr){1-5}
\multirow{2}{*}{\makecell{LAION-400M\\(400M)}}
& \multirow{2}{*}{\makecell[l]{CLIP\\(768)}} 
& global & 73.83 & 31.60 \\
& & local & 87.57 & 53.35 \\
\cmidrule(lr){1-5}
\multirow{2}{*}{\makecell{LAION-2B\\(2B)}}
& \multirow{2}{*}{\makecell[l]{ConvNext\\(768)}} 
& global & 77.95 & 41.25\\
& & local & 92.87 & 64.44\\
\cmidrule(lr){1-5}
\multirow{2}{*}{\makecell{WebLI\\(10B)}}
& \multirow{2}{*}{\makecell[l]{SigLIP\\(1024)}} 
& global & 78.48 & 55.03 \\
& & local & 87.01 & 65.16 \\ 
\bottomrule

\end{tabular}
}
\label{tab:map_comparison}
\end{table}

\begin{figure}[t]
  \centering
  \begin{subfigure}[t]{0.7\linewidth}
    \centering
    \includegraphics [width=\linewidth]{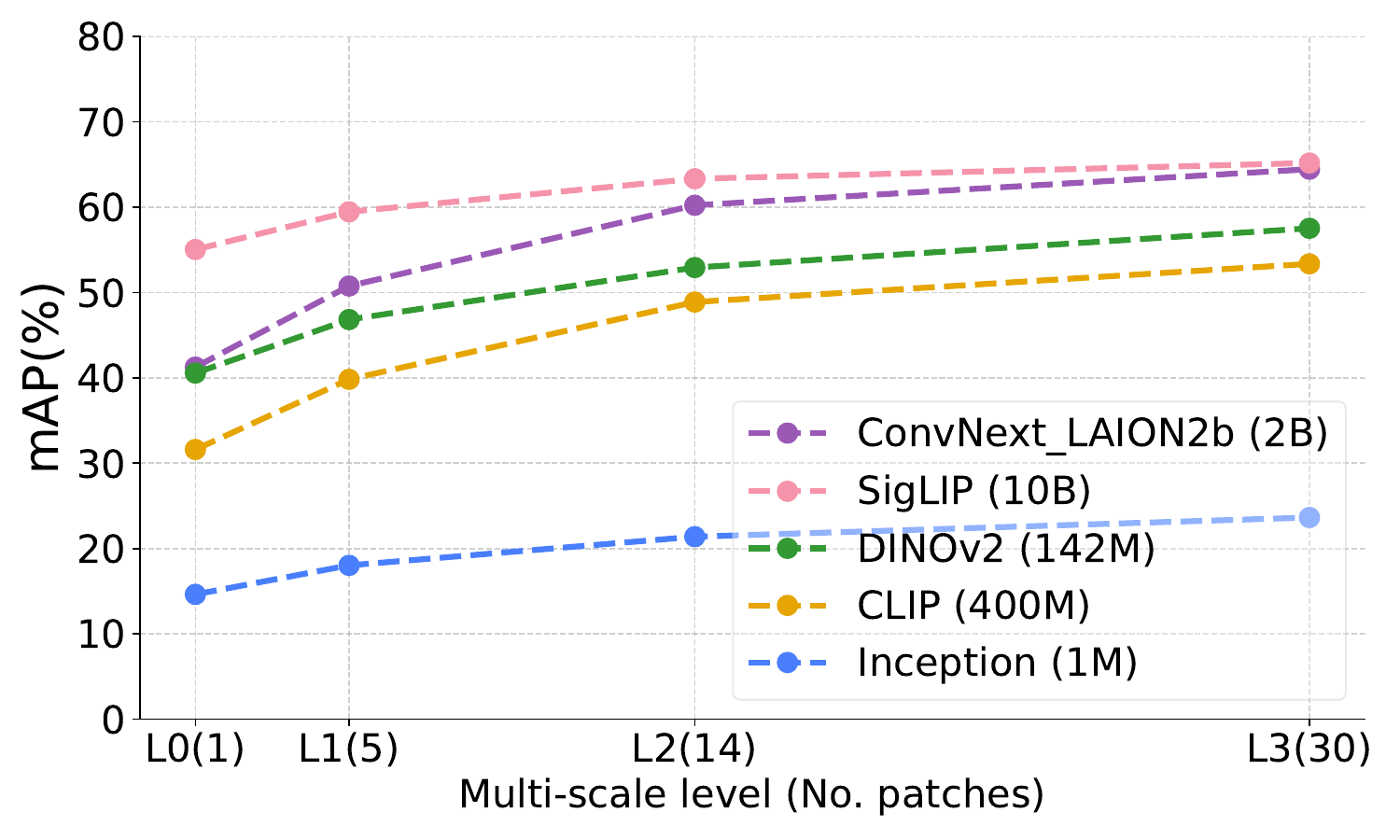}
    \caption{mAP}
    \label{fig:ilias_map_datasize}
  \end{subfigure}
  \hfill
  \begin{subfigure}[t]{0.8\linewidth}
    \centering
    \includegraphics[width=\linewidth]{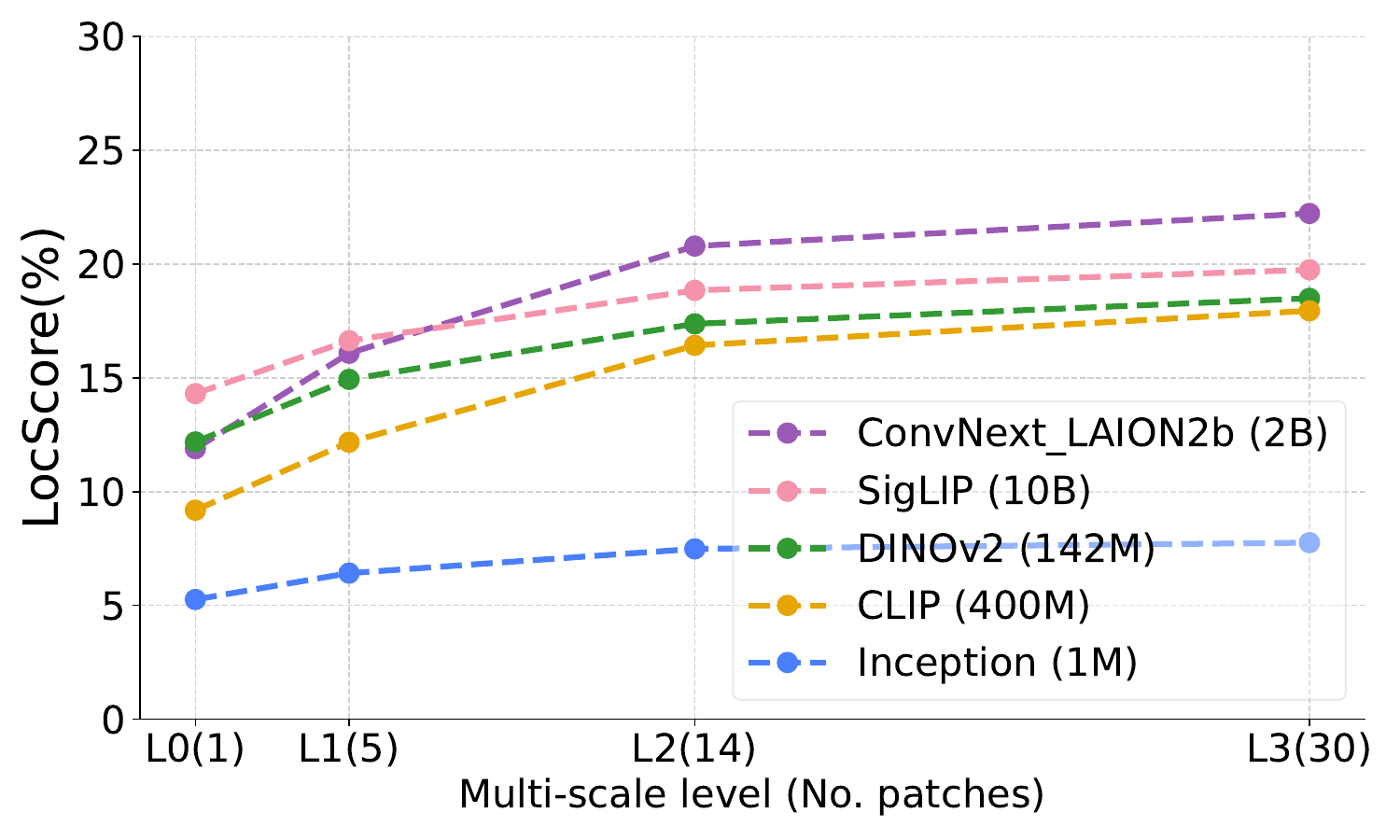}
    \caption{LocScore}
    \label{fig:ilias_lmap_datasize}
  \end{subfigure}

  \caption{Performance comparison on ILIAS under varying pretraining data sizes. We observe that performance is low if the amount of pre-training data is not sufficient.}
  \label{fig:multiscale_datasize}
\end{figure}

\begin{figure*}[!t]
    \centering
    \includegraphics[width=0.85\linewidth]{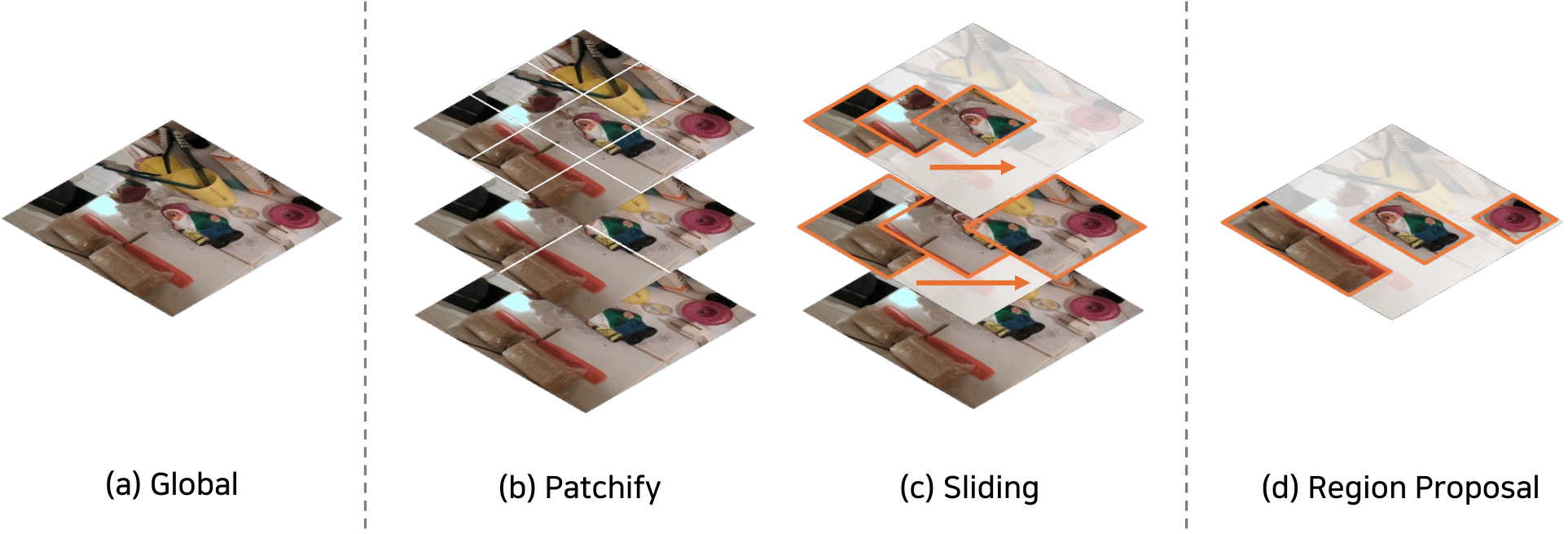}
    \caption{Methods used in analysis. In Global, we use only one embedding. In Patchify, the local features are extracted from the patches of an image. Sliding is similar to Patchify, but the patches have overlap. In Region Proposal, the detector is used to find the informative object in an image and extract the local features by cropping the object. }
    \label{fig:local_feature_method}
\end{figure*}

\paragraph{Impact of pretraining data size of encoder}
In terms of the exception case of ConvNeXt pretrained on LAION-2B, we hypothesize that the size of the dataset used during pretraining plays a significant role.
As shown in \Cref{tab:map_comparison} and \Cref{fig:multiscale_datasize}, model performance consistently improves with the scale of training data. 
Regardless of feature dimensionality, increasing the amount of training data improves generalization performance, which in turn enhances instance retrieval performance.
This result indicates that representation generalization is influenced by both model capacity and data scale, consistent with findings in prior work~\citep{zhai2022scaling, zhou2022domain}, and further confirms that this relationship holds for instance retrieval task as well
that such an effect also extends to instance retrieval tasks.


\section{Different region selection strategies}
\label{sec:region_strategies}
We describe the experimental details for evaluating different region selection strategies. We adopt SigLIP as the visual encoder throughout, and conduct evaluations on both the INSTRE and ILIAS core datasets. The following outlines each strategy illustrated in \Cref{fig:local_feature_method}:
\vspace{-1em}
\paragraph{Global}
This method follows the standard global feature approach, where the entire image is passed through the encoder to produce a single feature vector.
\vspace{-1em}
\paragraph{Patchify}
This is our proposed method. The image is divided into non-overlapping grid patches according to a predefined configuration (e.g., L0, L1, etc.). Each patch is independently passed through the image encoder to extract local descriptors. L0 corresponds to a single $1 \times 1$ patch (i.e., the global setting), L1 includes both $1 \times 1$ and $2 \times 2$ patches, L2 adds $3 \times 3$ patches on top of L1, and L3 further includes $4 \times 4$ patches. This cumulative multi-scale design allows for spatial interpretability, as the most similar patch can be traced back to a specific region in the image, revealing which part contributed most to the retrieval.
\vspace{-1em}
\paragraph{Sliding}
To assess whether denser and more exhaustive spatial coverage can improve performance, we implement a sliding window approach on the Patchify setting with two different strides: 0.5 and 0.25 relative to the patch size. Unlike Patchify, the sliding window method introduces overlapping patches for denser spatial coverage. This allows us to assess how performance changes as spatial precision improves due to denser sampling. Unlike naive patchify with a grid-based approach, this method produces overlapping patches, potentially increasing the likelihood of including a well-localized region. The goal is to evaluate whether finer sampling improves retrieval accuracy through better spatial alignment.
\vspace{-1em}
\paragraph{Region Proposal}
We further evaluate a more semantically aware region extraction method using Grounding DINO~\cite{liu2024groundingdinomarryingdino}, a recent region proposal network capable of producing high-quality object-level bounding boxes. We extract up to 20 bounding boxes per image  with a fixed text prompt "object" and use output bounding boxes as regions for feature extraction. Since these proposals are semantically meaningful and often closely aligned with ground-truth objects, this setting allows us to assess the potential upper bound of retrieval performance when informative and well-localized regions are used. Note that the use of region proposal method for our method is not limited to only Grounding DINO. If any of methods that can produce meaningful and tight region can be used. For example, Segment Anything~\cite{kirillov2023segment}. Our method can benefit from producing object-aligned patches and
informative positives for PQ training, which aligns with our
finding that PQ benefits from informative patches.

\begin{table}[t]
\centering
\caption{Comparison with SOTA methods on ILIAS core. Patchify achieves competitive performance as a single-stage method and further boosts performance when integrated into reranking pipelines.}
\resizebox{\linewidth}{!}{
\begin{tabular}{l l c}
\toprule
\textbf{Setting} & \textbf{Feature Extraction Method} & \textbf{mAP (\%)} \\
\midrule
\multirow{2}{*}{Single-stage} 
  & SigLIP (Linear Adaptation)~\cite{tolias2025ilias} & 63.96 \\
  & Patchify (SigLIP~\cite{zhai2023sigmoid}) & \textbf{65.16} \\
\midrule
\multirow{4}{*}{Two-stage} 
  & DINOv2 w/ registers~\cite{darcet2024visiontransformersneedregisters} & 63.37 \\
  & SigLIP~\cite{tolias2025ilias} & 75.61 \\
  & Patchify (SigLIP~\cite{zhai2023sigmoid}) & 79.54 \\
  & Patchify (SigLIP, Linear Adaptation~\cite{tolias2025ilias}) & \textbf{83.52} \\
\bottomrule
\end{tabular}
}
\label{tab:sota_comparison}
\end{table}

\section{Comparison with SOTA Methods}
\label{sec:sota}
In this section, we describe the implementation details for the experiments conducted in \Cref{sec:sota_comparison}. Furthermore we also report retrieval performance is measured on ILIAS core set~\cite{tolias2025ilias}, which was not reported in the main paper.

\subsection{Implementation details}
\paragraph{AMES}
For AMES~\cite{suma2024ames}, we adopt its asymmetric transformer-based reranking method. During inference, the top-$m$ images retrieved using global similarity are reranked based on local similarities computed via transformer interaction between the query and database local descriptors. We use the binary distilled variant of AMES with a universal model trained across varying local descriptor counts. The ensemble similarity between global and local features is computed as a weighted combination, tuned via grid search.

\paragraph{ILIAS Baseline}
We also benchmark against the reranking strategy proposed in the ILIAS benchmark~\cite{tolias2025ilias}. This method involves reranking a shortlist of images retrieved via global descriptors using dense feature correlation with ground-truth instance boxes. As this approach depends on densely sampled or region proposal-based local features, it tends to have higher memory and computation overhead.

\paragraph{Hyperparameter Search}
Since reranking approaches combine global and local similarities, hyperparameter tuning is critical. We perform a grid search over the following parameters to identify the best configuration:
\begin{itemize}
    \item Number of reranked candidates: \texttt{[0, 100, 200, 400, 800, 1000]}
    \item Global-local balance weight $\lambda$: \texttt{[0.0, 0.1, 0.2, 0.3, 0.4, 0.5, 0.6, 0.7, 0.8, 0.9, 1.0]}
    \item Temperature scaling $\gamma$: \texttt{[0.0, 0.05, 0.1, 0.2, 0.3, 0.4, 0.5, 0.6, 0.7, 0.8, 0.9, 1.0]}
\end{itemize}
We report the performance using the best configuration selected from this grid for each method.

\subsection{Retrieval performance on ILIAS core set}
We compare against two SOTA methods: DINOv2 w/ registers~\cite{darcet2024visiontransformersneedregisters} from AMES~\cite{suma2024ames}, and SigLIP with linear adaptation~\cite{tolias2025ilias} from ILIAS. As summmarized in Table~\ref{tab:sota_comparison}, Patchify achieves stronger performance than the fine-tuned baseline on both single-stage and reranking methods. This is aligned with the results on mini-ILIAS, which was reported in the main paper.

\subsection{LocScore on mini-ILIAS}
As shown in Table \ref{tab:mini_ilias_locscore}, consistent with the earlier results, we observe the same trend: as mAP increases, LocScore increases as well.


\section{Product Quantization}
\label{sec:pq}
To enable scalable retrieval with local features, we adopt Inverted File with Product Quantization (IVFPQ), a widely used technique that combines coarse clustering with quantization to allow efficient approximate nearest neighbor search with significantly reduced memory and latency overhead. To configure IVFPQ, we perform a grid search over the number of subvectors $m \in \{16, 32, 64\}$ and the number of clusters $nlist \in \{2048, 4096\}$, and select $m = 64$, $nlist = 4096$, and $nbits = 8$ based on retrieval performance.

\begin{table}[t!]
\centering
\caption{Impact of multi-scale feature levels on storage size (MB) before and after applying product quantization. CR denotes the compression ratio. We observe that compression is scalable to high memory database.}
\resizebox{\columnwidth}{!}{
\begin{tabular}{lcccc}
\toprule
\multirow{2}[2]{*}{\textbf{\makecell[l]{Level\\(No. Patches)}}}
& \multicolumn{2}{c}{\textbf{INSTRE}} 
& \multicolumn{2}{c}{\textbf{ILIAS}} \\ 
\cmidrule(rl){2-3}\cmidrule(rl){4-5}
& \textbf{Non-quant.} & \textbf{Quant (CR)}
& \textbf{Non-quant.} & \textbf{Quant (CR)} \\
\midrule
L0 (1)   & 103.68 & 19.77 (5.24) & 19.08 & 18.21 (1.05) \\
L1 (5)   & 518.42 & 27.41 (18.91) & 95.41 & 19.62 (4.86) \\
L2 (14)  & 1420 & 44.61 (31.83) & 267.15 & 22.78 (11.73) \\
L3 (30)  & 3040 & 71.71 (42.39) & 572.46 & 28.41 (20.15) \\
\bottomrule
\end{tabular}
}
\label{tab:dbsize}
\end{table}

\Cref{tab:dbsize} summarizes the impact of increasing patch granularity on database size before and after applying IVFPQ. As the number of patches increases from L0 to L3, uncompressed storage grows rapidly, by more than 29× on INSTRE and 30× on ILIAS. However, with IVFPQ, this increase is drastically reduced to just 3.6× and 1.6×, respectively. These results confirm that IVFPQ effectively scales to high-resolution patch representations, enabling practical deployment of patch-based retrieval with minimal memory overhead.

We investigate how the choice of training features affects PQ effectiveness. As summarized in \Cref{tab:pq_training_level_comparison}, the highest performance is achieved when PQ is trained with features extracted from ground-truth bounding boxes (G.T.), which provide strong semantic alignment with the target instances. In contrast, using patches from intermediate configurations (e.g., L1 and L2) results in lower performance, possibly due to noisy or irrelevant background content in those features. Interestingly, PQ trained with global features (L0) outperforms L1 and L2, suggesting that lower-resolution but semantically focused representations may be more beneficial than ambiguous fine-grained features.

\begin{figure}[h]
    \centering
    \includegraphics[width=\linewidth]{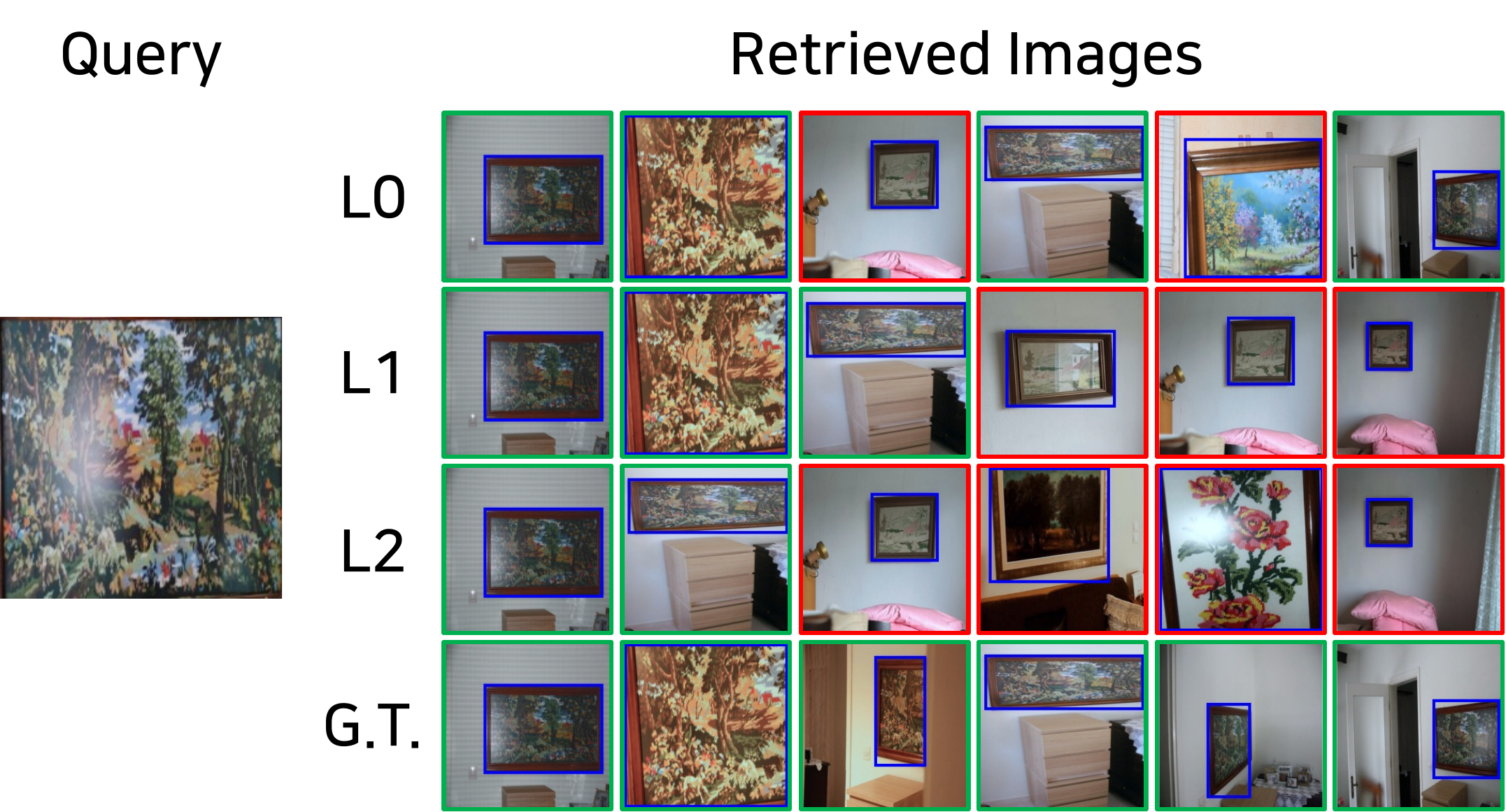}
    \label{fig:pq_qualitative}
  \caption{Qualitative retrieval results using different PQ training features at L3. PQ trained on G.T. finds the correct instance compared to others. Thus, we also consider the informative features when training PQ.}
  \label{fig:combined_pq_results}
\vspace{-1em}
\end{figure}

Qualitative examples in \Cref{fig:combined_pq_results} support this observation: G.T.-trained PQ accurately retrieves the correct instance under challenging conditions such as occlusion, scale change, and viewpoint variation, while PQ trained on L2 fails to distinguish between visually similar but incorrect instances. These findings highlight the importance of using informative training features for PQ optimization and offer valuable guidance for future work on compressed local feature retrieval.

\begin{figure}[t]
  \centering
    \centering
    \includegraphics[width=\linewidth]{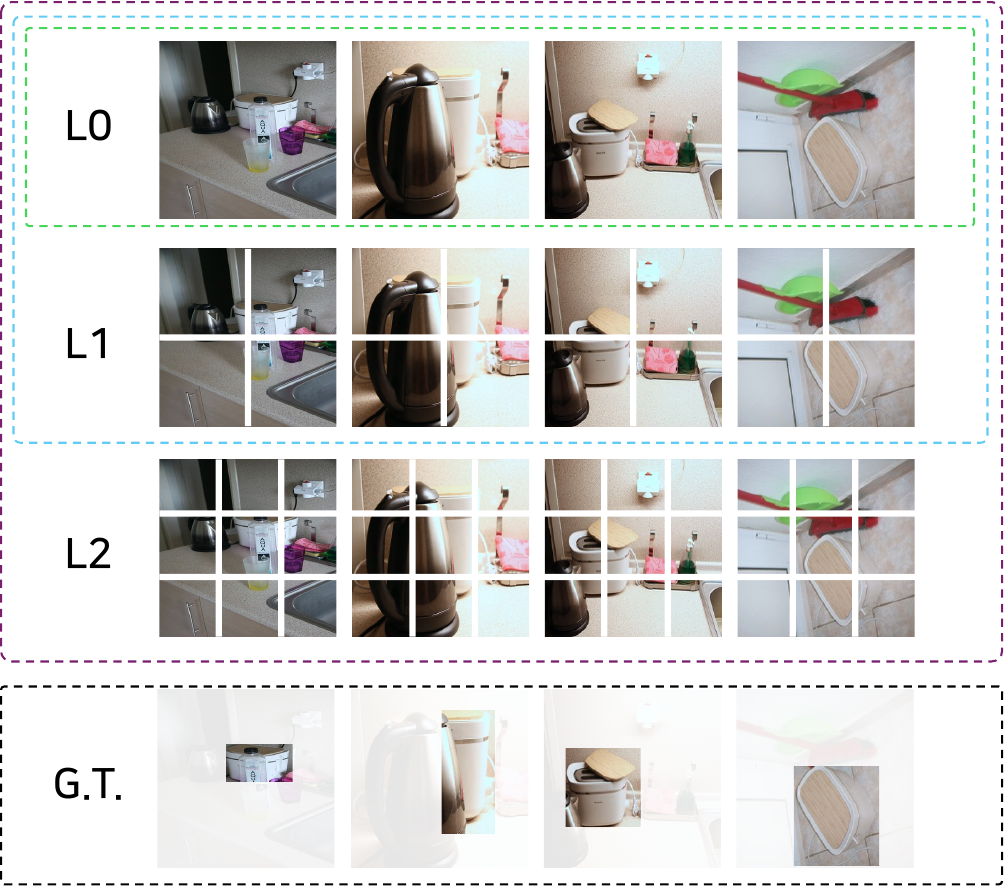}
    \caption{Images used to train PQ indices in each Patchify level and G.T. We can observe that features from G.T. are related to the instance, directly.}
    \label{fig:pq_train_images}
\end{figure}

\section{LocScore}
\label{sec:sup_locscore}

This section provides extended analyses of the localization behavior measured by LocScore that were omitted from the main paper due to space limits. While the main paper introduced the metric and its motivation, here we investigate how different IoU criteria affect retrieval behavior and how localization quality varies across feature levels and model variants.

\begin{figure*}[p]
    \centering

    \begin{subfigure}[b]{0.45\textwidth}
        \includegraphics[width=\textwidth]{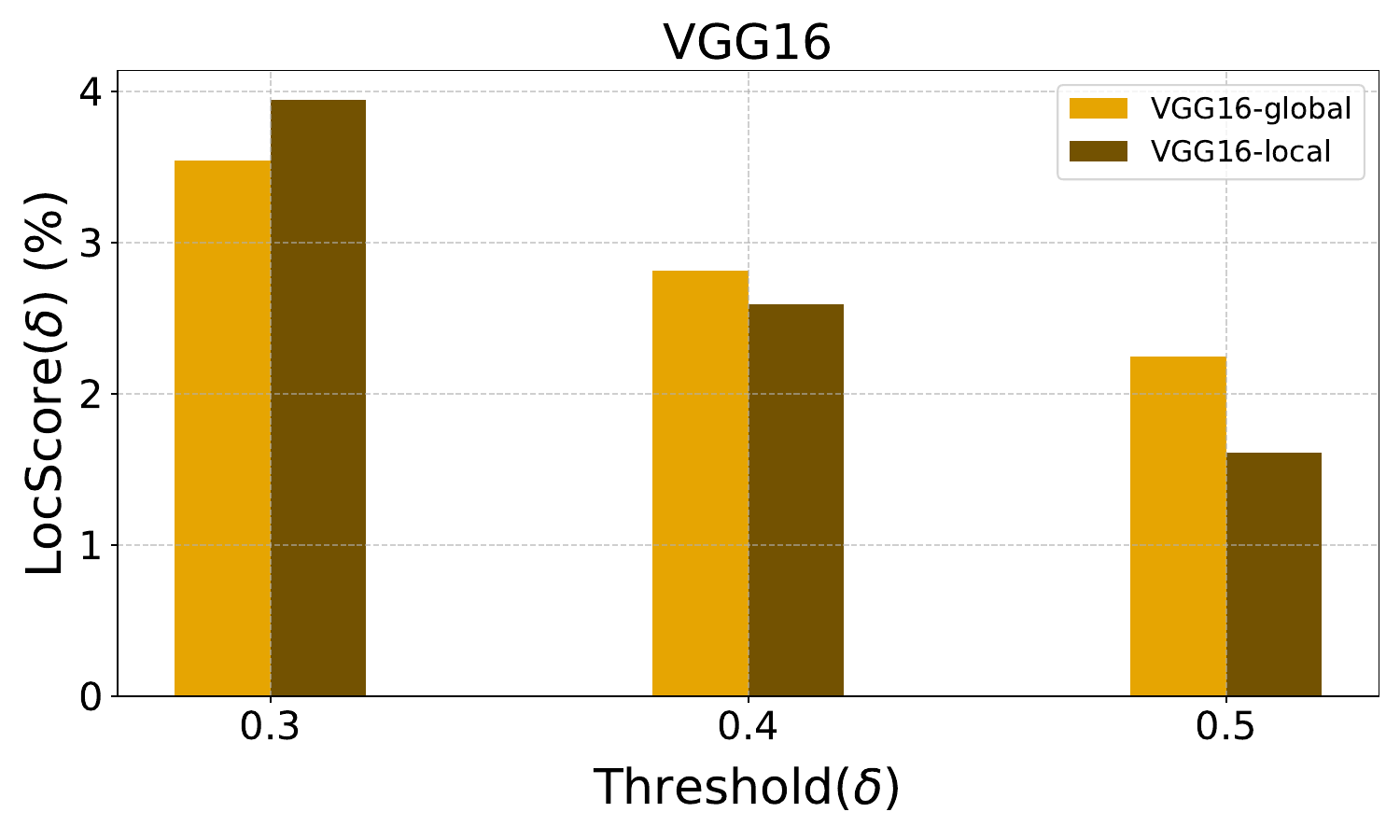}
        \caption{VGG16}
    \end{subfigure}
    \hfill
    \begin{subfigure}[b]{0.45\textwidth}
        \includegraphics[width=\textwidth]{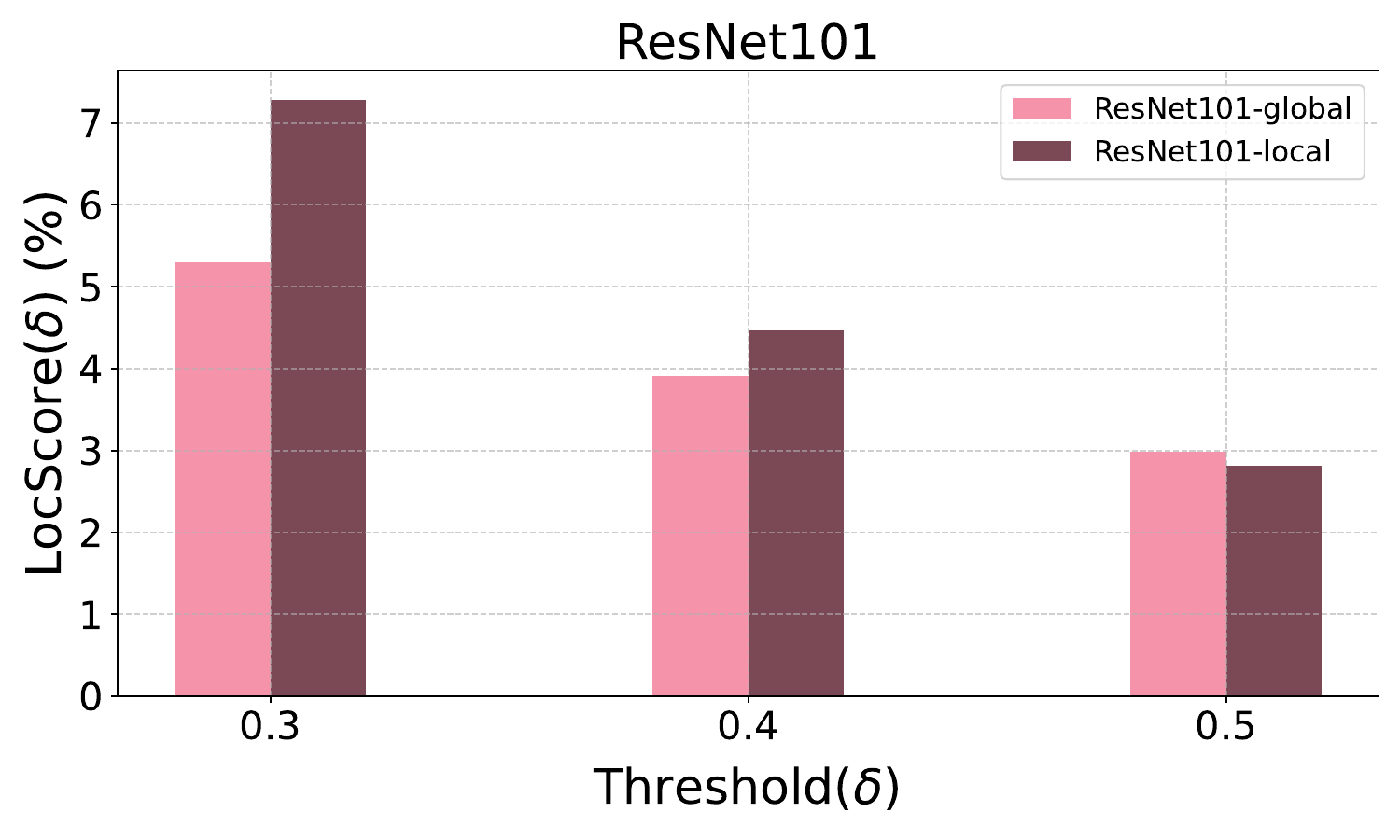}
        \caption{ResNet}
    \end{subfigure}

    \vspace{0.45cm}
    
    \begin{subfigure}[b]{0.45\textwidth}
        \includegraphics[width=\textwidth]{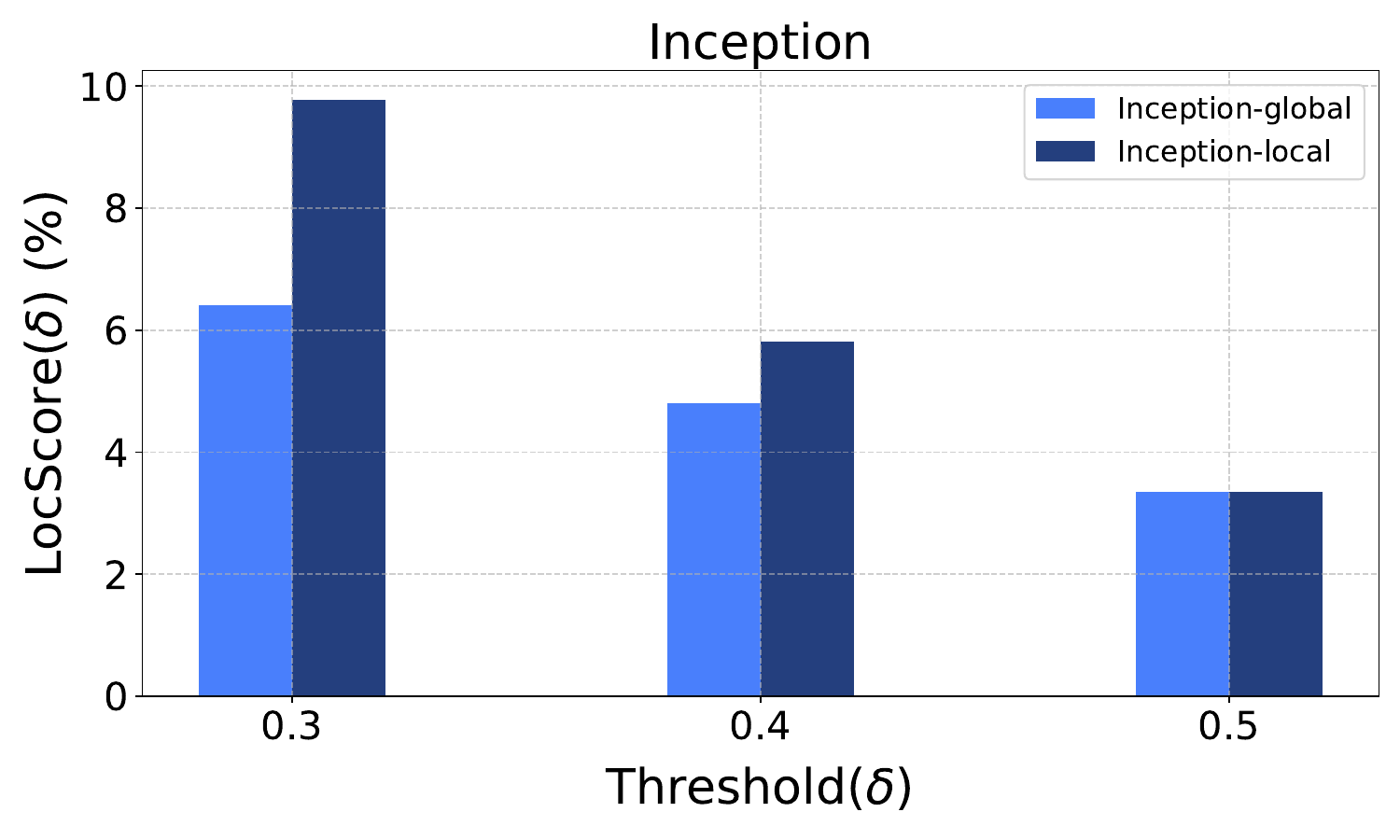}
        \caption{Inception}
    \end{subfigure}
    \hfill
    \begin{subfigure}[b]{0.45\textwidth}
        \includegraphics[width=\textwidth]{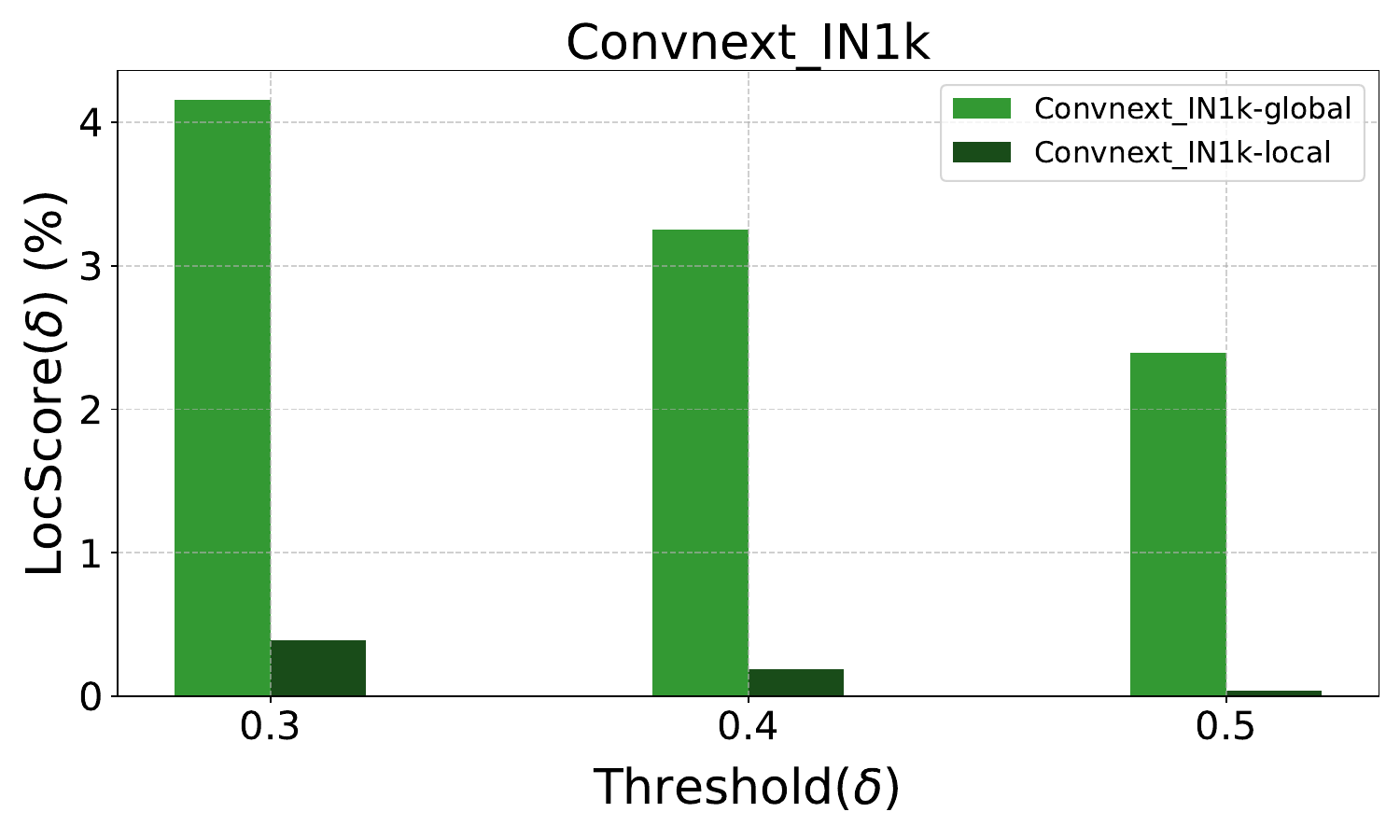}
        \caption{ConvNext-IN1k}
    \end{subfigure}
    \vspace{0.45cm}

    \begin{subfigure}[b]{0.45\textwidth}
        \includegraphics[width=\textwidth]{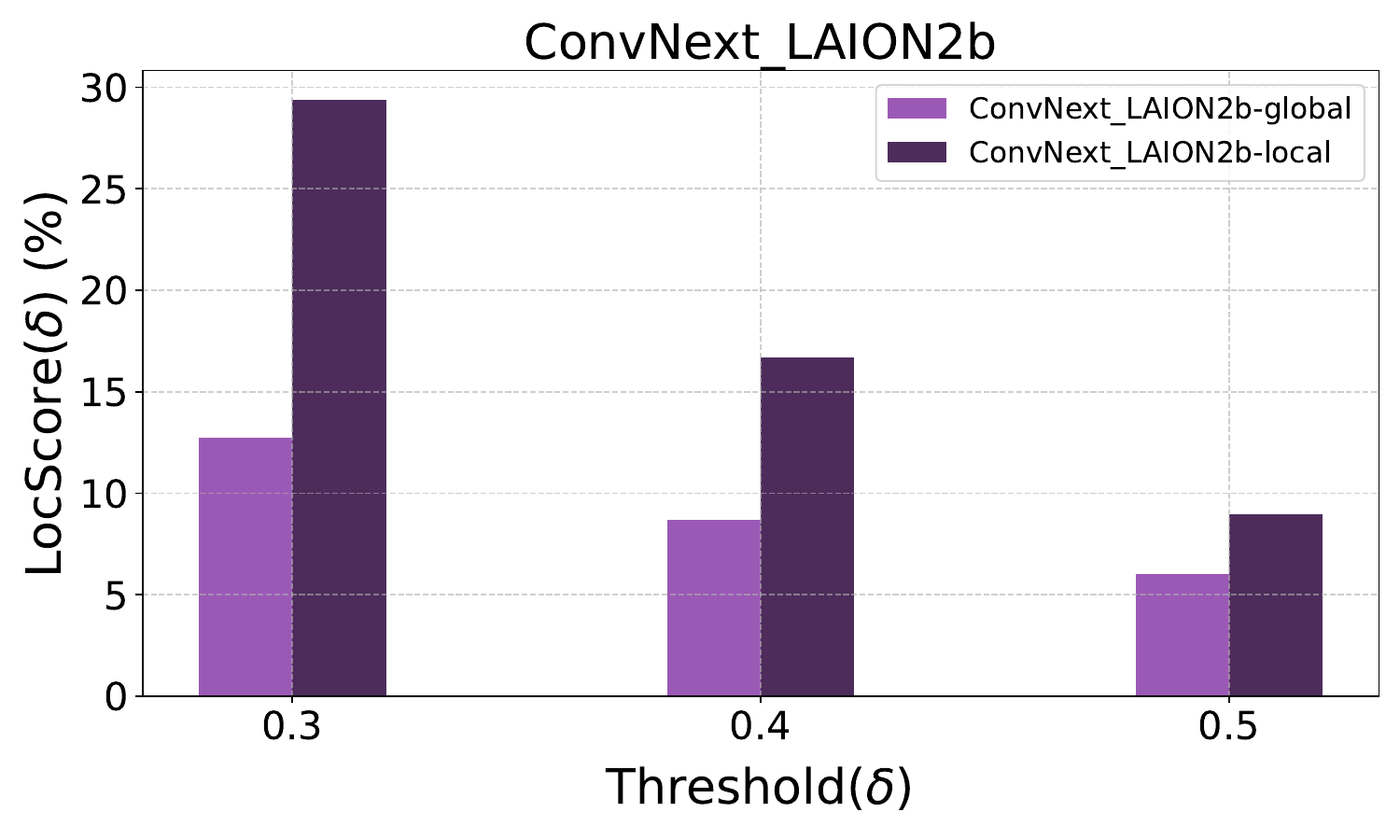}
        \caption{ConvNext-LAION2b}
    \end{subfigure}
    \hfill
    \begin{subfigure}[b]{0.45\textwidth}
        \includegraphics[width=\textwidth]{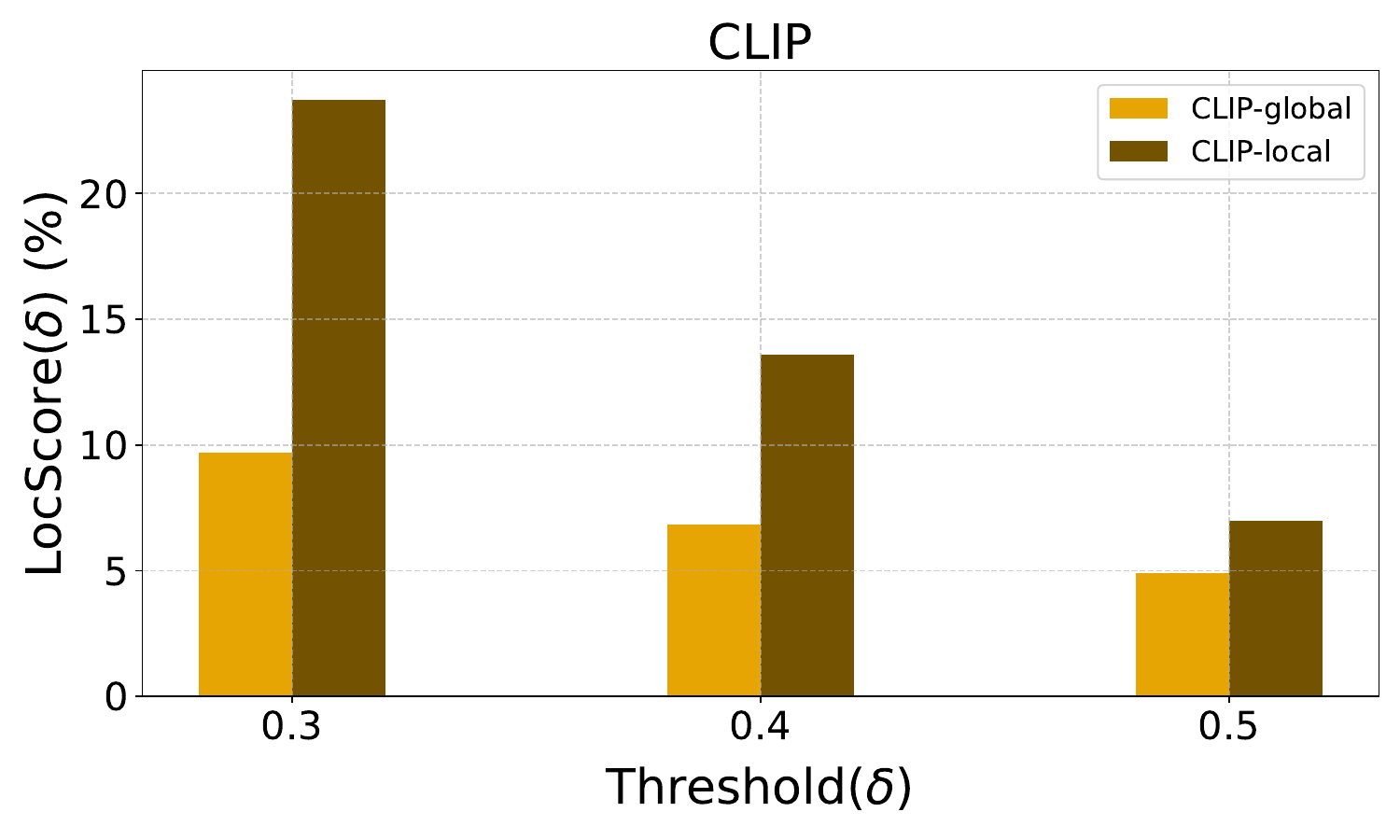}
        \caption{CLIP}
    \end{subfigure}

    \vspace{0.45cm}

    \begin{subfigure}[b]{0.45\textwidth}
        \includegraphics[width=\textwidth]{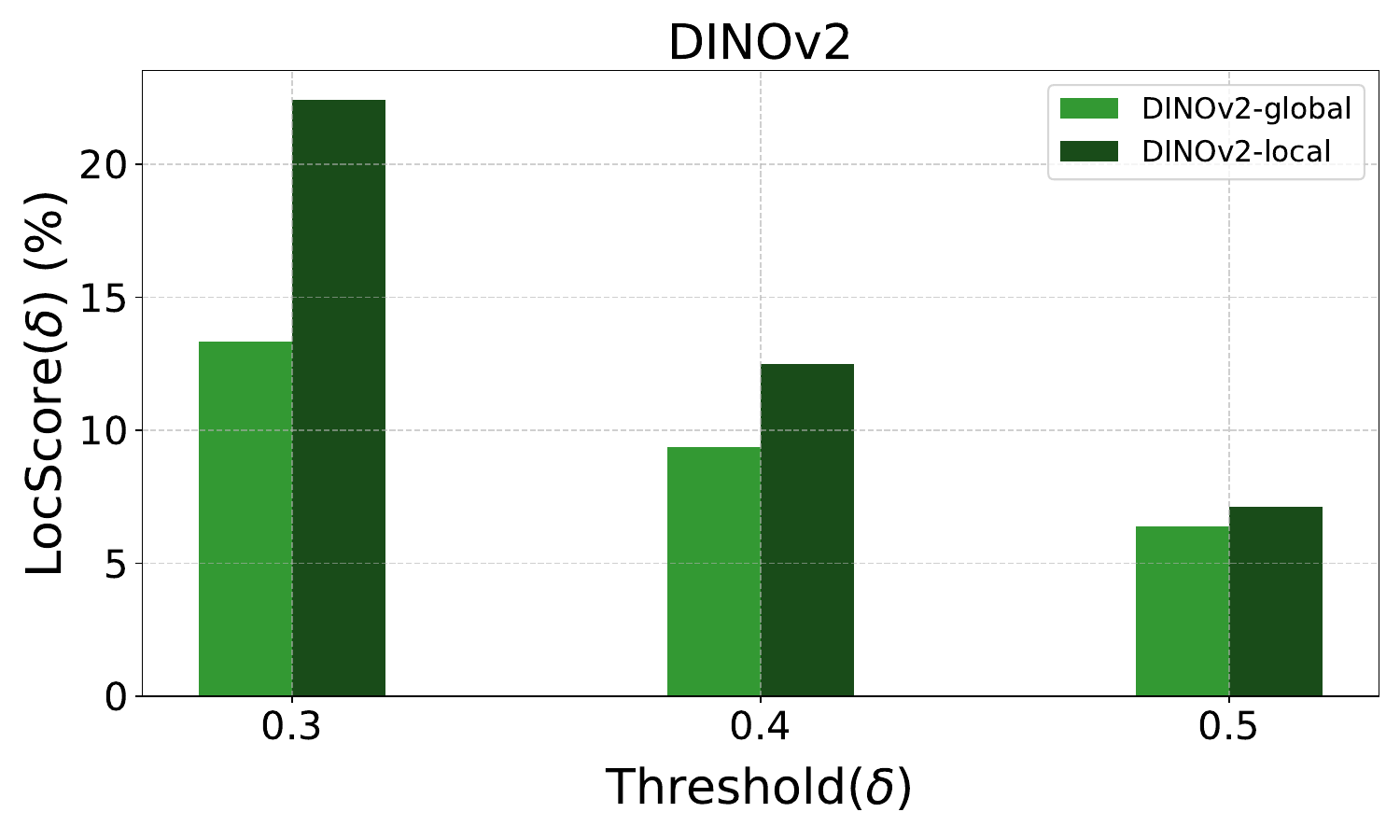}
        \caption{DINOv2}
    \end{subfigure}
    \hfill
    \begin{subfigure}[b]{0.45\textwidth}
        \includegraphics[width=\textwidth]{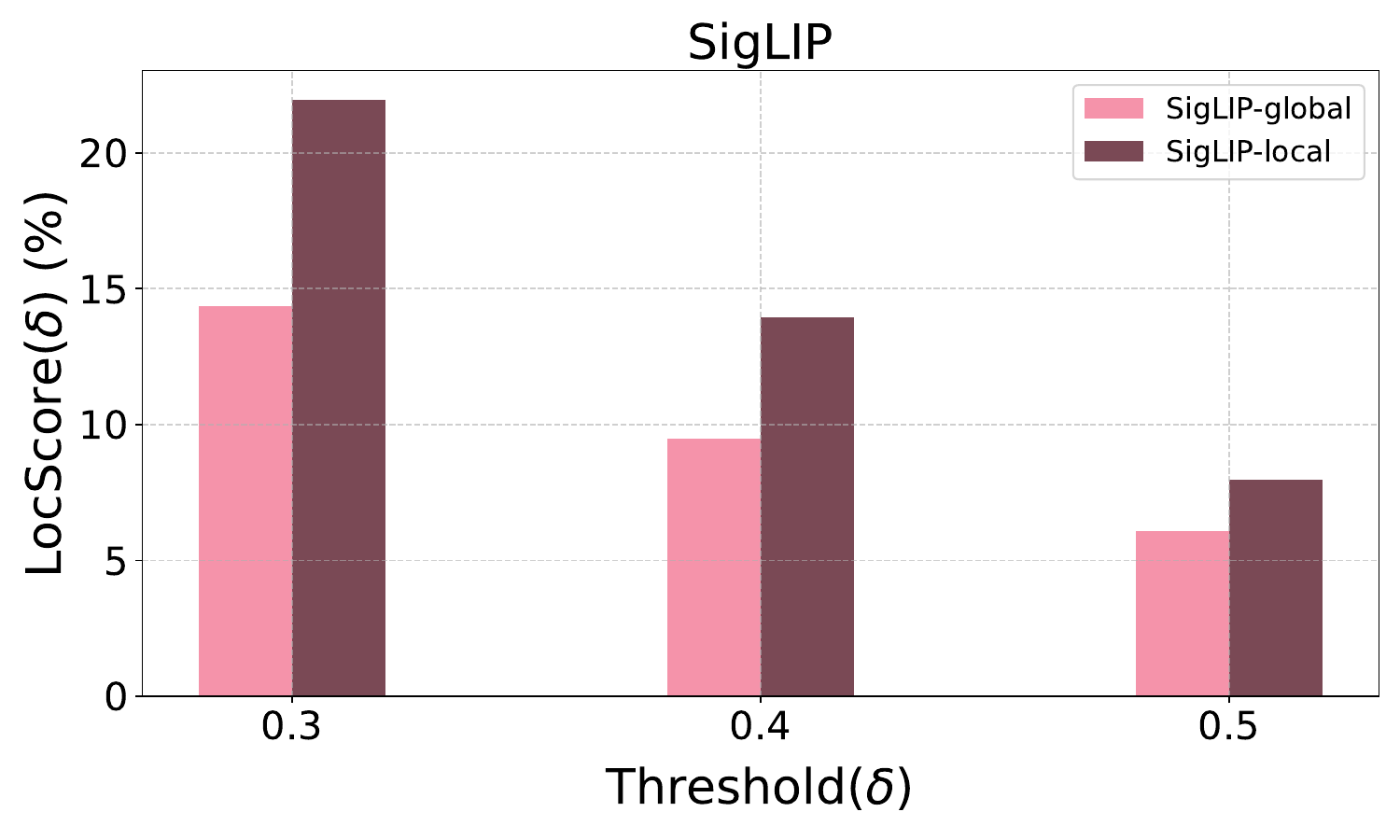}
        \caption{SigLIP}
    \end{subfigure}
    \hfill
    \begin{subfigure}[b]{0.45\textwidth}
        ~ 
    \end{subfigure}

    \caption{LocScore($\delta$) visualizations across threshold $\delta=0.3, 0.4, 0.5$ of CNN-based and Transformer-based models on ILIAS.}
    \label{fig:ilias_grid}
\end{figure*}

\begin{figure*}[p]
    \centering

    \begin{subfigure}[b]{0.45\textwidth}
        \includegraphics[width=\textwidth]{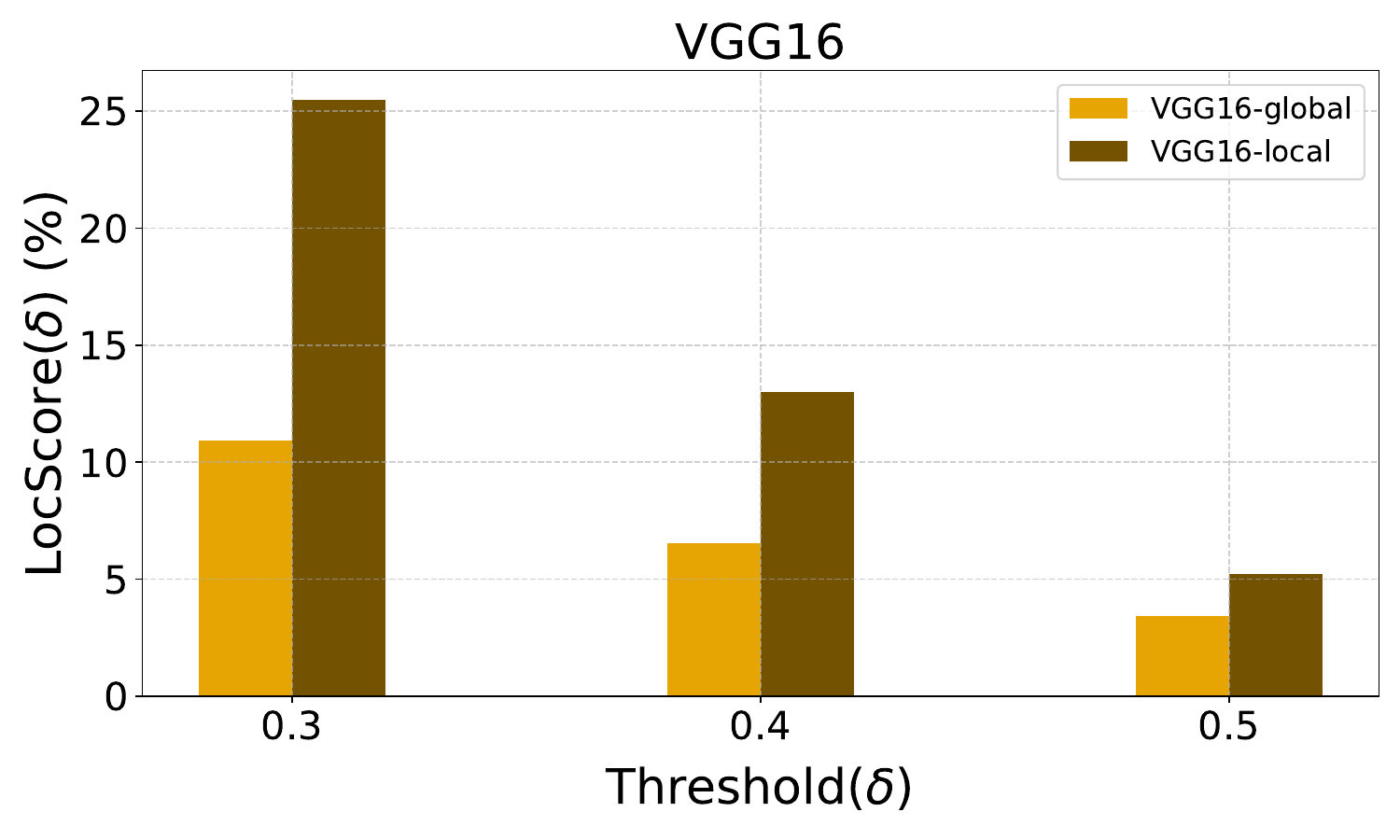}
        \caption{VGG16}
    \end{subfigure}
    \hfill
    \begin{subfigure}[b]{0.45\textwidth}
        \includegraphics[width=\textwidth]{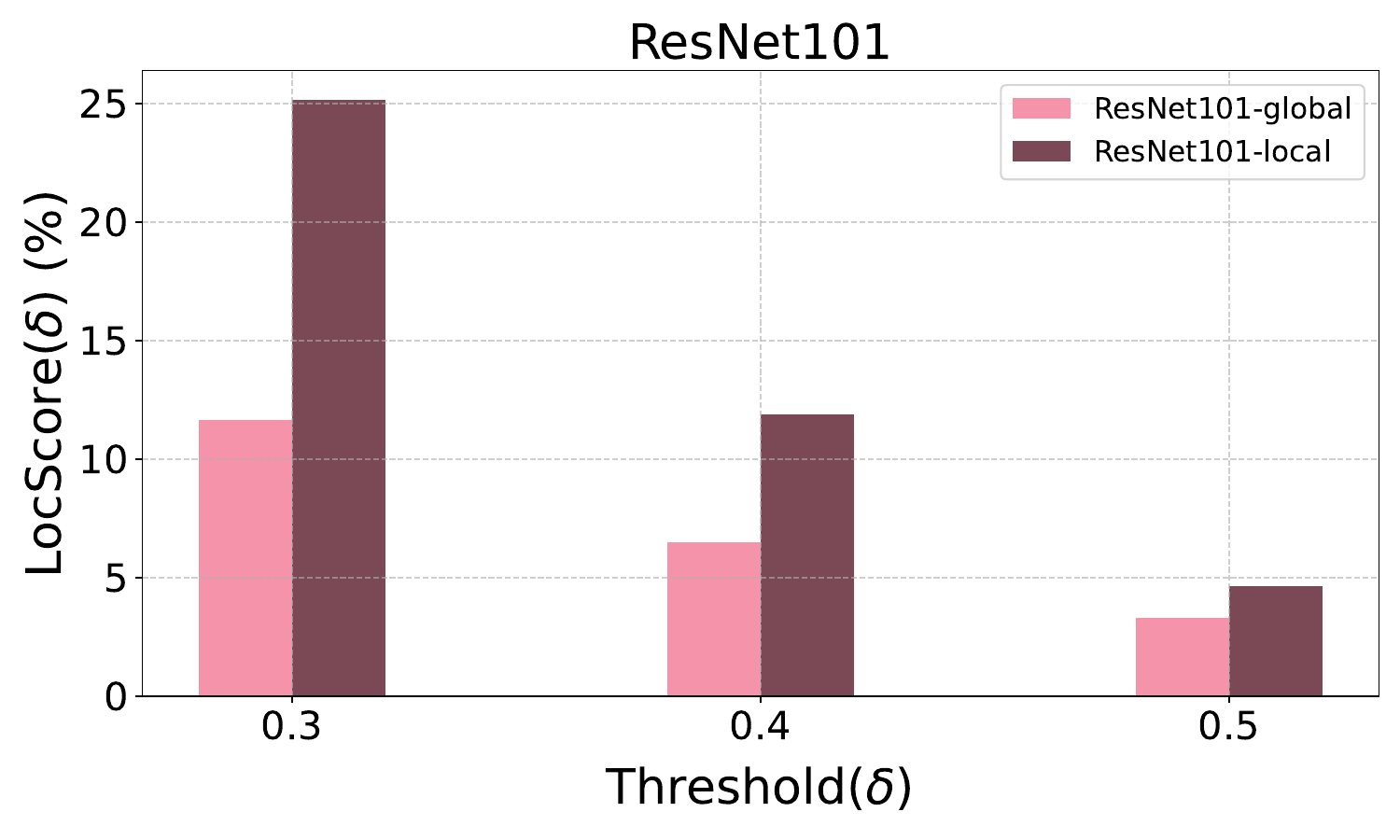}
        \caption{ResNet}
    \end{subfigure}

    \vspace{0.45cm}
    
    \begin{subfigure}[b]{0.45\textwidth}
        \includegraphics[width=\textwidth]{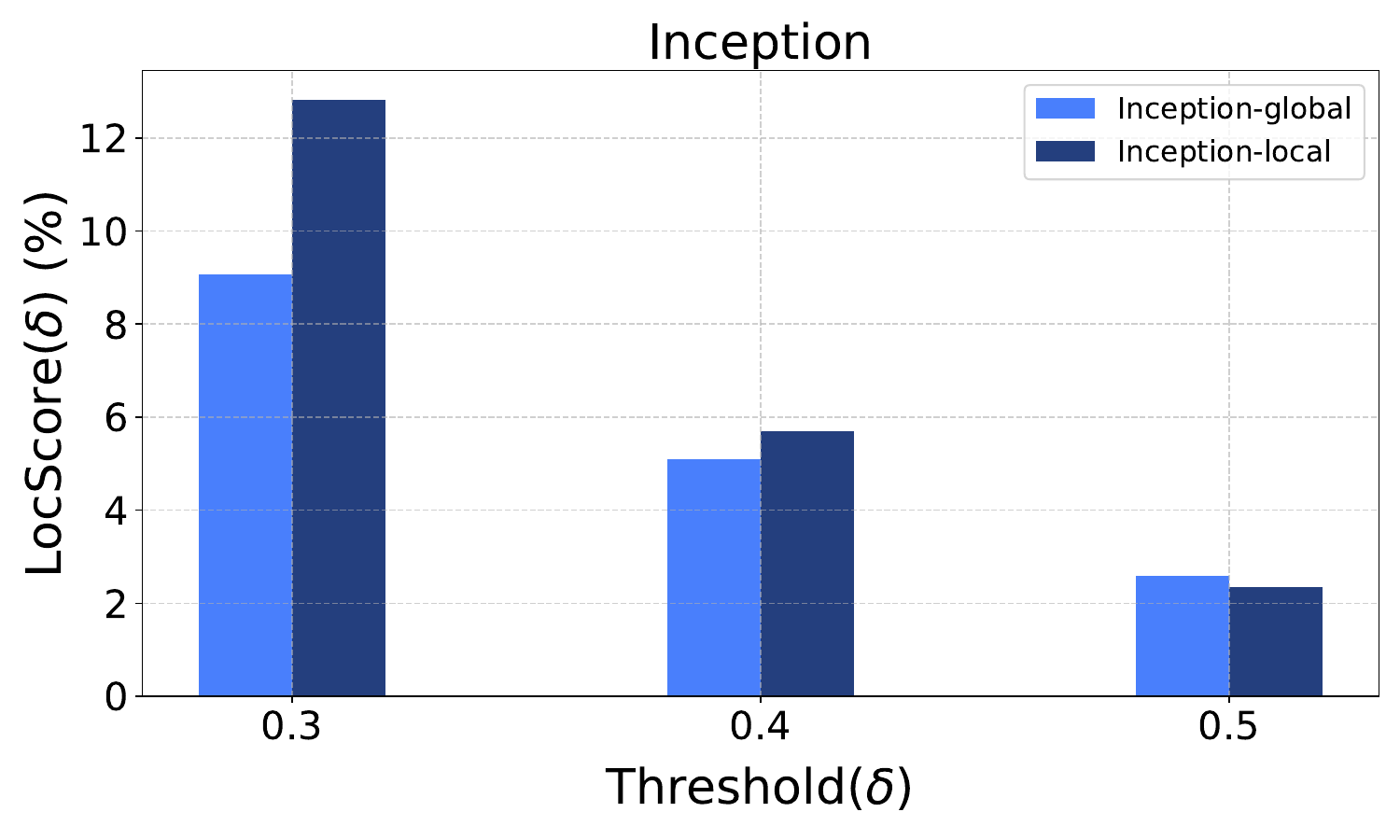}
        \caption{Inception}
    \end{subfigure}
    \hfill
    \begin{subfigure}[b]{0.45\textwidth}
        \includegraphics[width=\textwidth]{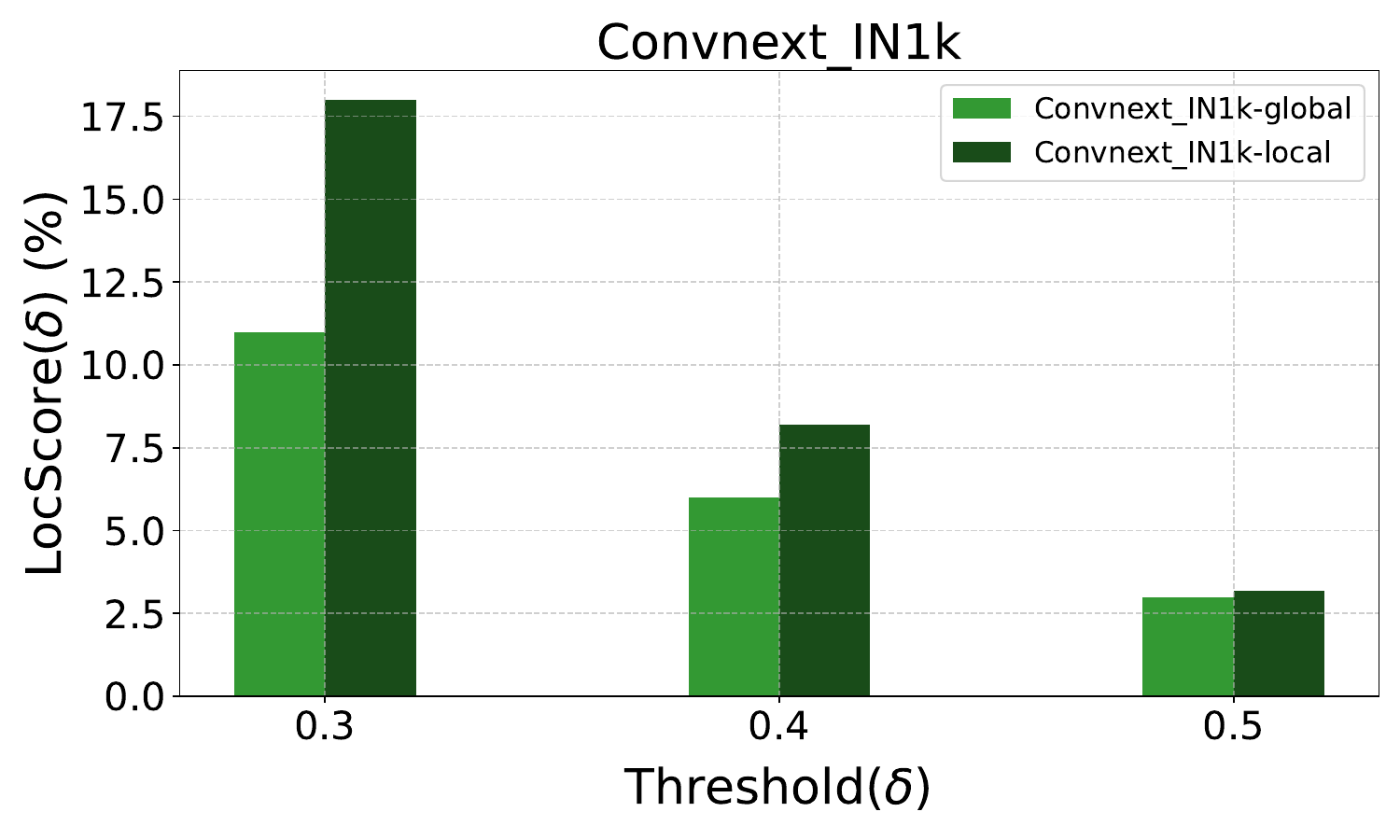}
        \caption{ConvNext-IN1k}
    \end{subfigure}
    \vspace{0.45cm}

    \begin{subfigure}[b]{0.45\textwidth}
        \includegraphics[width=\textwidth]{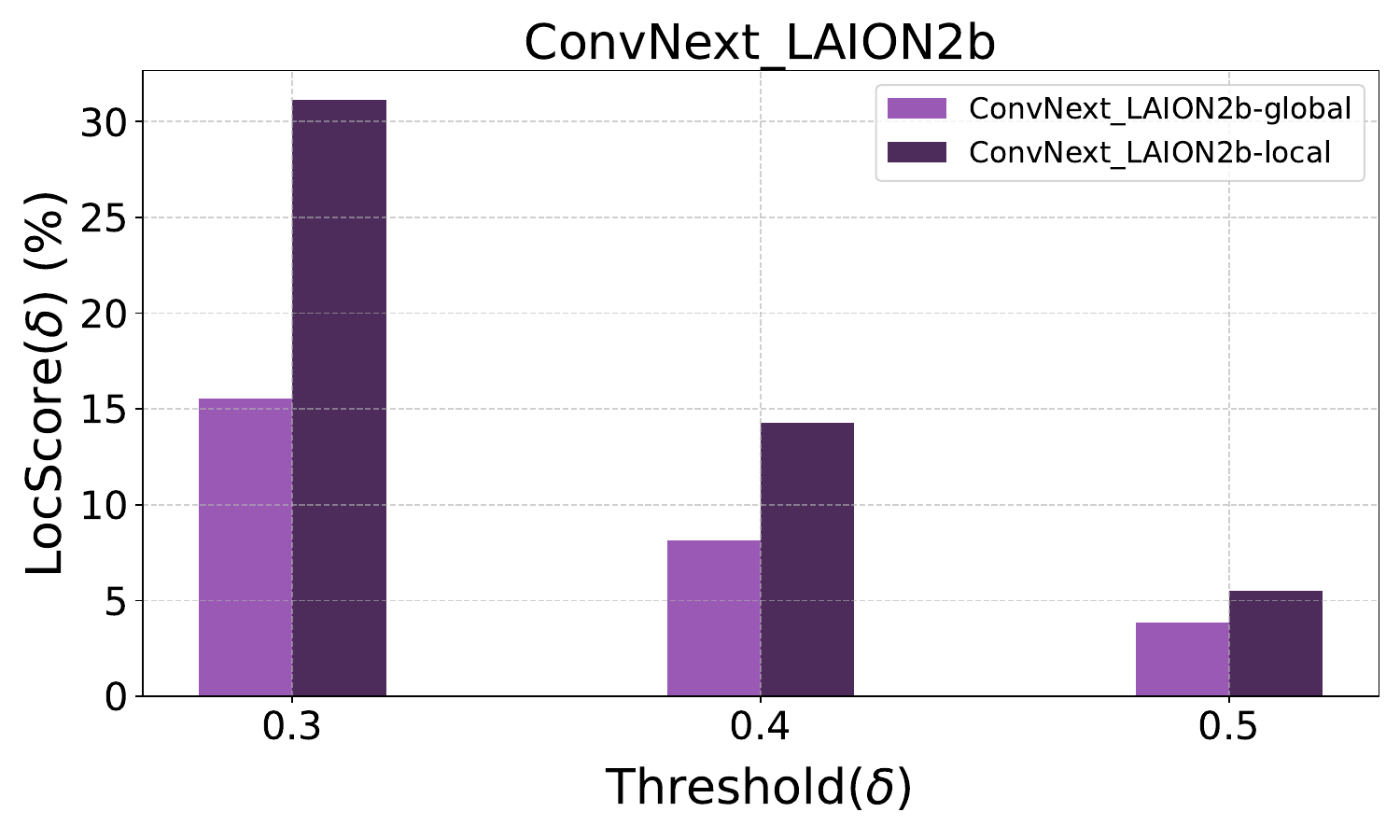}
        \caption{ConvNext-LAION2b}
    \end{subfigure}
    \hfill
    \begin{subfigure}[b]{0.45\textwidth}
        \includegraphics[width=\textwidth]{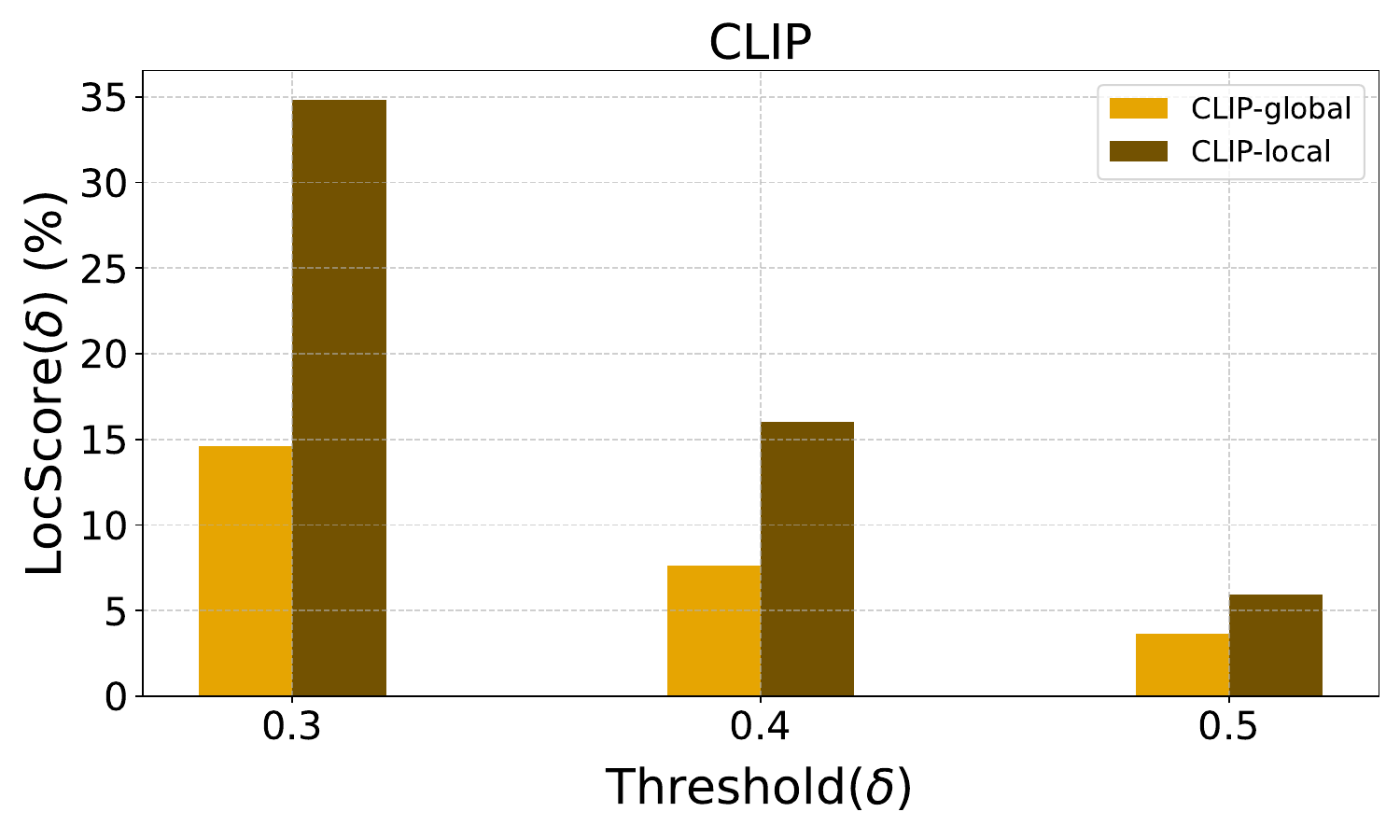}
        \caption{CLIP}
    \end{subfigure}

    \vspace{0.45cm}

    \begin{subfigure}[b]{0.45\textwidth}
        \includegraphics[width=\textwidth]{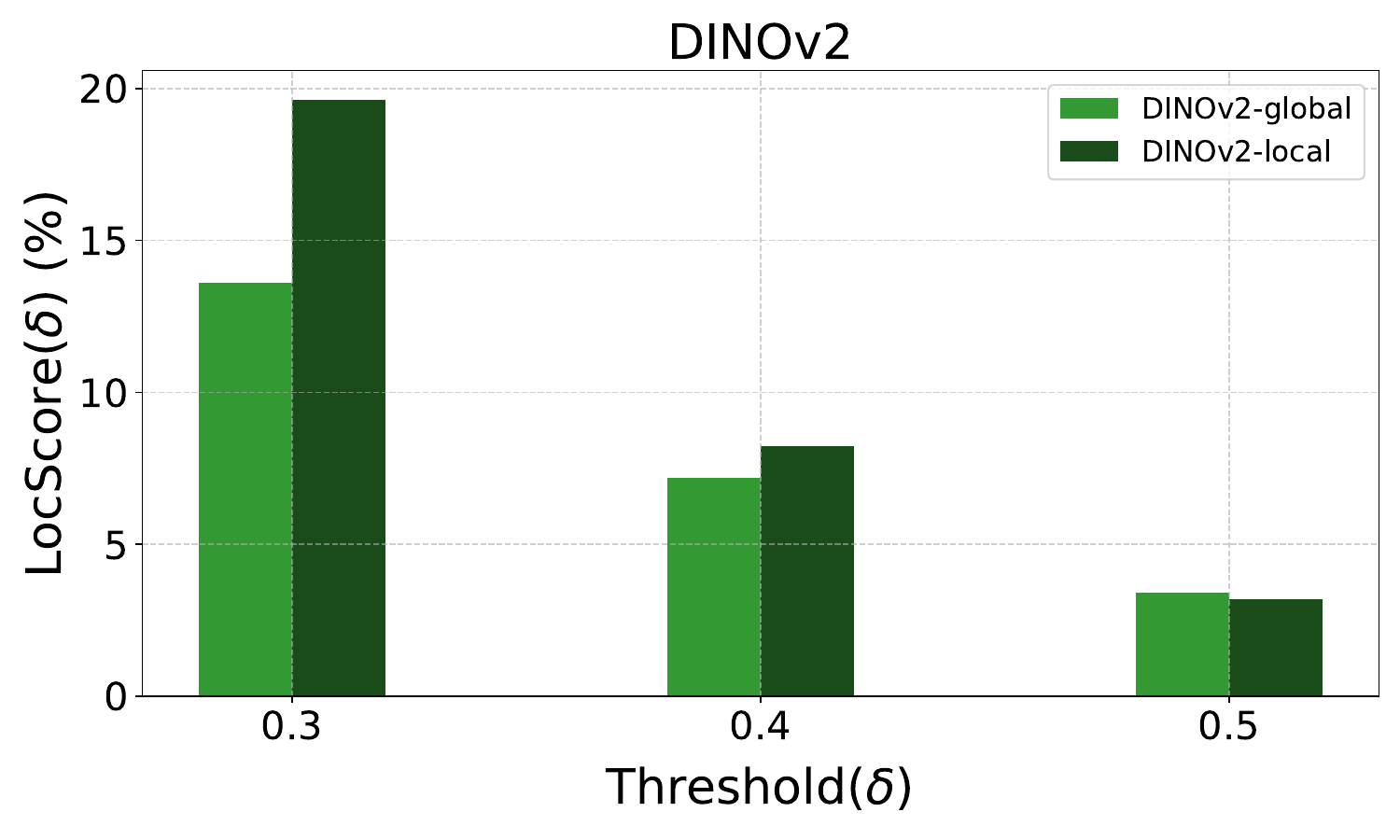}
        \caption{DINOv2}
    \end{subfigure}
    \hfill
    \begin{subfigure}[b]{0.45\textwidth}
        \includegraphics[width=\textwidth]{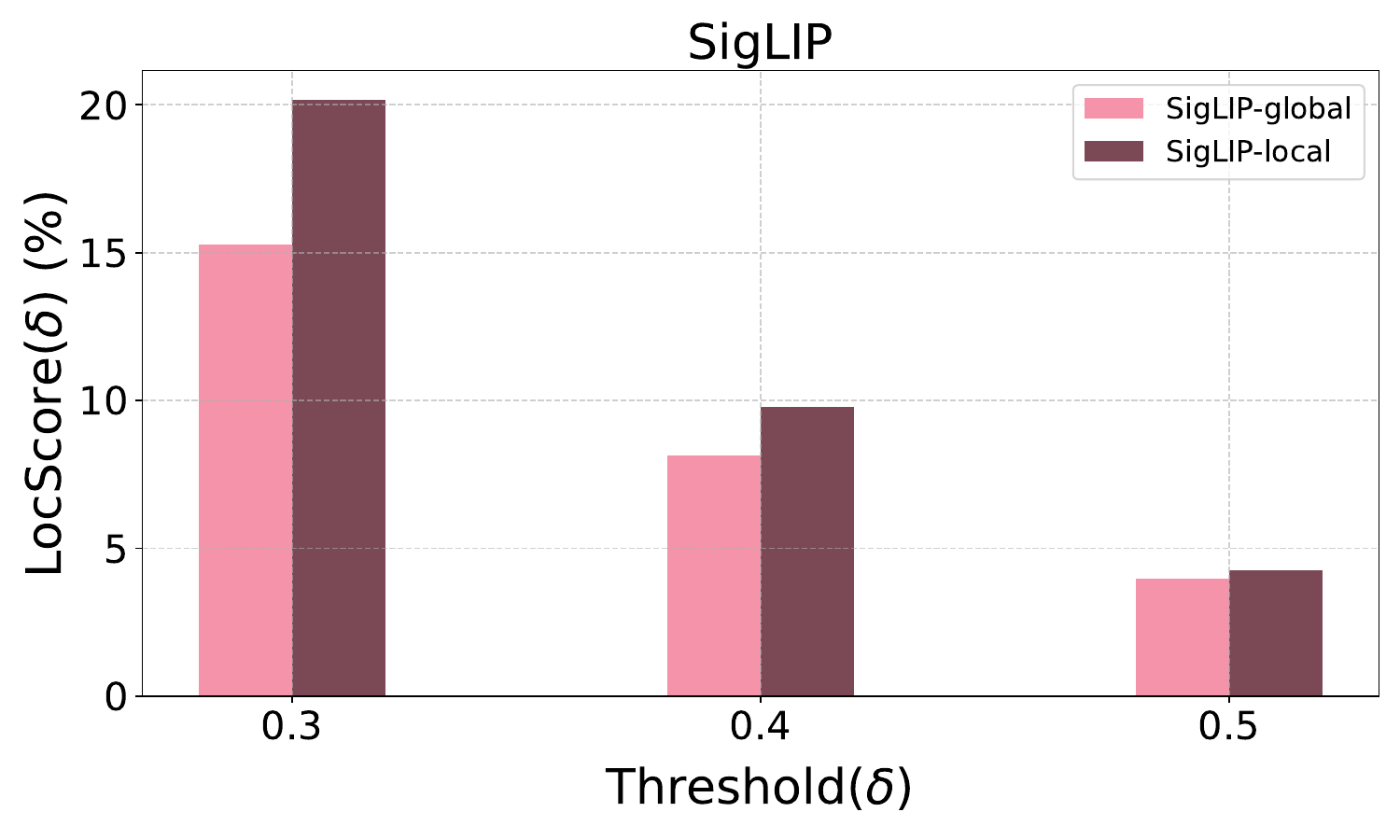}
        \caption{SigLIP}
    \end{subfigure}
    \hfill
    \begin{subfigure}[b]{0.45\textwidth}
        ~ 
    \end{subfigure}

    \caption{LocScore($\delta$) visualizations across threshold $\delta=0.3, 0.4, 0.5$ of CNN-based and Transformer-based models on INSTRE.}
    \label{fig:instre_grid}
\end{figure*}

Figures~\ref{fig:ilias_grid} and \ref{fig:instre_grid} visualize the thresholded LocScore($\delta$) for multiple IoU thresholds on ILIAS-core and INSTRE. These results reveal that different models degrade at different rates as the spatial criterion becomes stricter. For example, methods that rely on coarse global features show a rapid decline beyond $\delta\!=\!0.3$, whereas patch-based retrieval remains more stable up to $\delta\!=\!0.5$. This indicates that localization robustness is tightly linked to the granularity of visual evidence used by the model.

Figures~\ref{supple_encoder_map_locscore1} and \ref{supple_encoder_map_locscore2} report the LocScore averaged across thresholds, providing a summary measure that reflects overall spatial reliability. This aggregated score smooths out threshold-specific fluctuations and highlights consistent performance trends that are not fully visible from single-threshold evaluations. In particular, models with strong mid-level representations maintain competitive scores across a wide range of $\delta$, suggesting that localization accuracy is distributed across multiple feature layers rather than dominated by a single level.

Figure~\ref{fig:comparison_locscore_thr} illustrates how different IoU criteria impact qualitative retrieval outcomes. Unlike the continuous variant, which reflects gradual changes in spatial alignment, thresholded LocScore($\delta$) exposes failure modes where retrieved patches partially overlap the target but fail to satisfy a strict IoU requirement. These cases highlight the sensitivity of certain models to small spatial displacements. The contrast between the two forms, therefore, helps to diagnose whether a model's failure stems from ranking errors (low IoU overall) or boundary precision (IoU slightly below $\delta$).

Together, these supplementary results provide a more complete understanding of the spatial behavior of retrieval systems. They show how localization precision evolves under increasingly strict criteria, how robustness varies across feature hierarchies, and how different formulations of LocScore reveal distinct types of spatial failure.

\begin{figure*}[t]
  \centering

  \begin{subfigure}{0.95\linewidth}
    \centering
    \includegraphics[width=\linewidth]{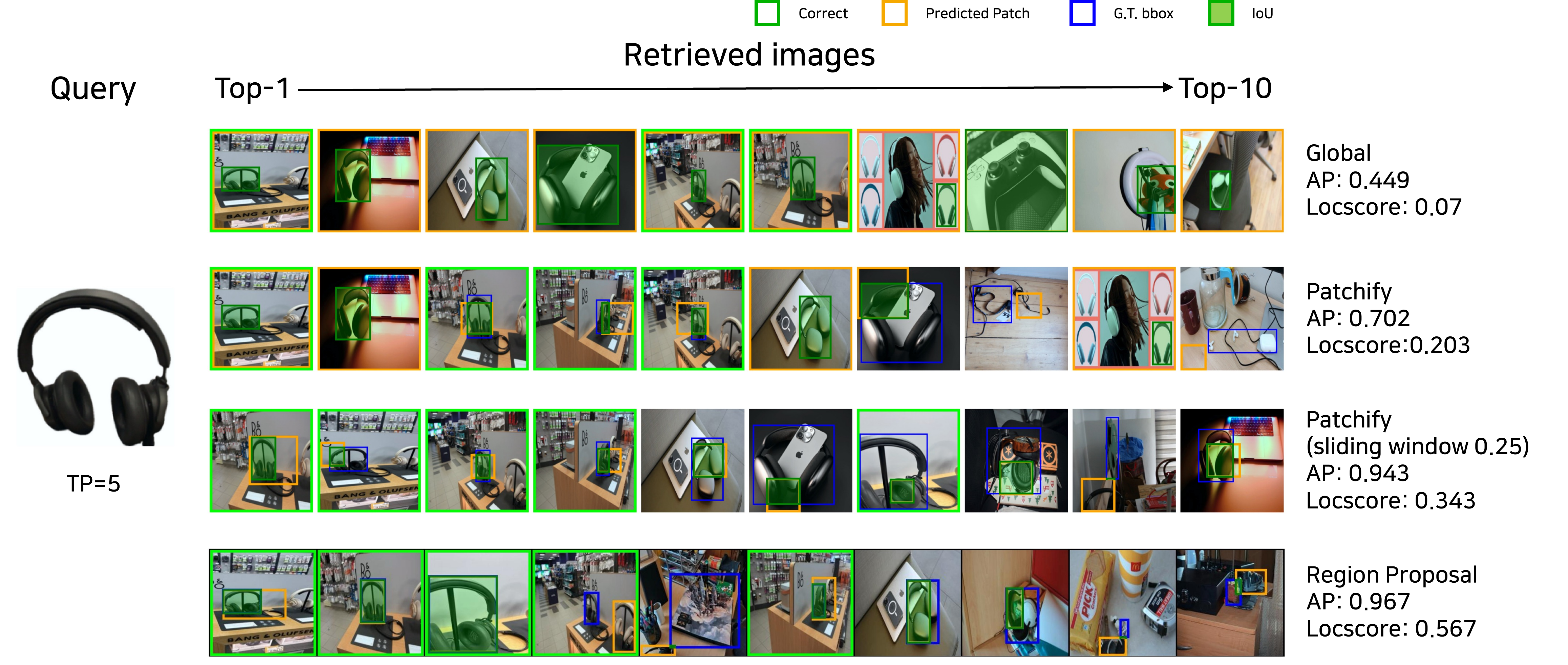}
    \caption{Comparison of LocScore by method}
    \label{fig:comparison_locscore_by_method}
  \end{subfigure}

  \vspace{1.0em}

  \begin{subfigure}{0.95\linewidth}
    \centering
    \includegraphics[width=\linewidth]{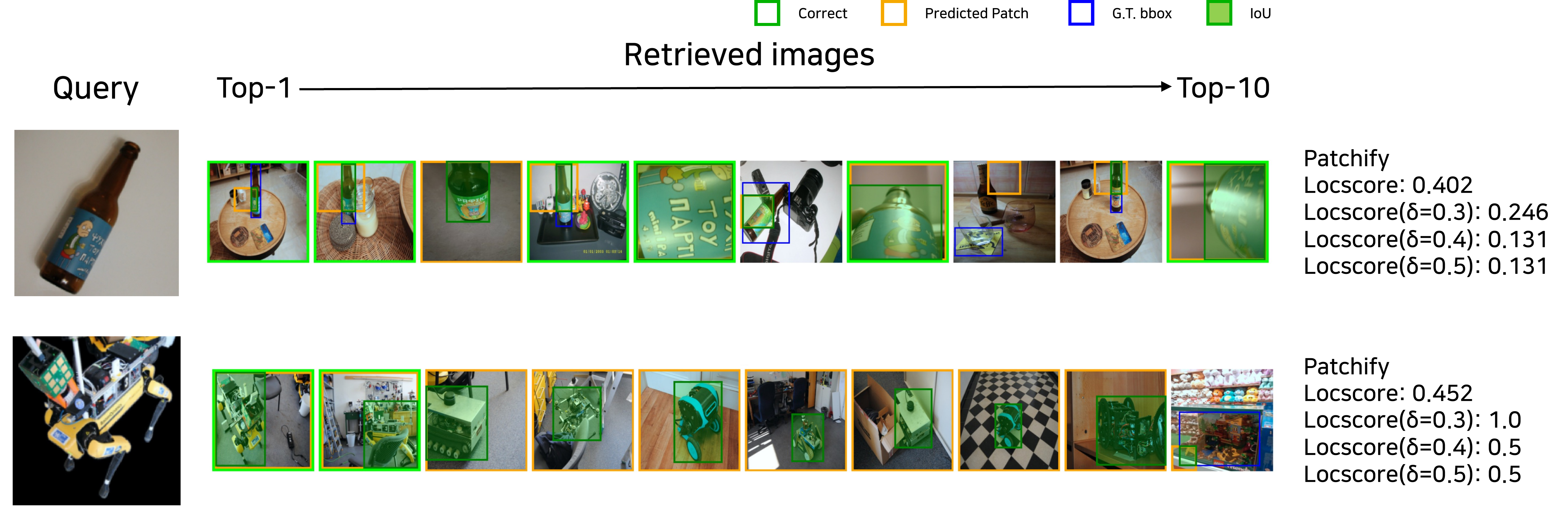}
    \caption{Comparison of LocScore and LocScore($\delta$)}
    \label{fig:comparison_locscore_locscorethr}
  \end{subfigure}

  \vspace{1.0em}

  \begin{subfigure}{0.5\linewidth}
    \centering
    \includegraphics[width=\linewidth]{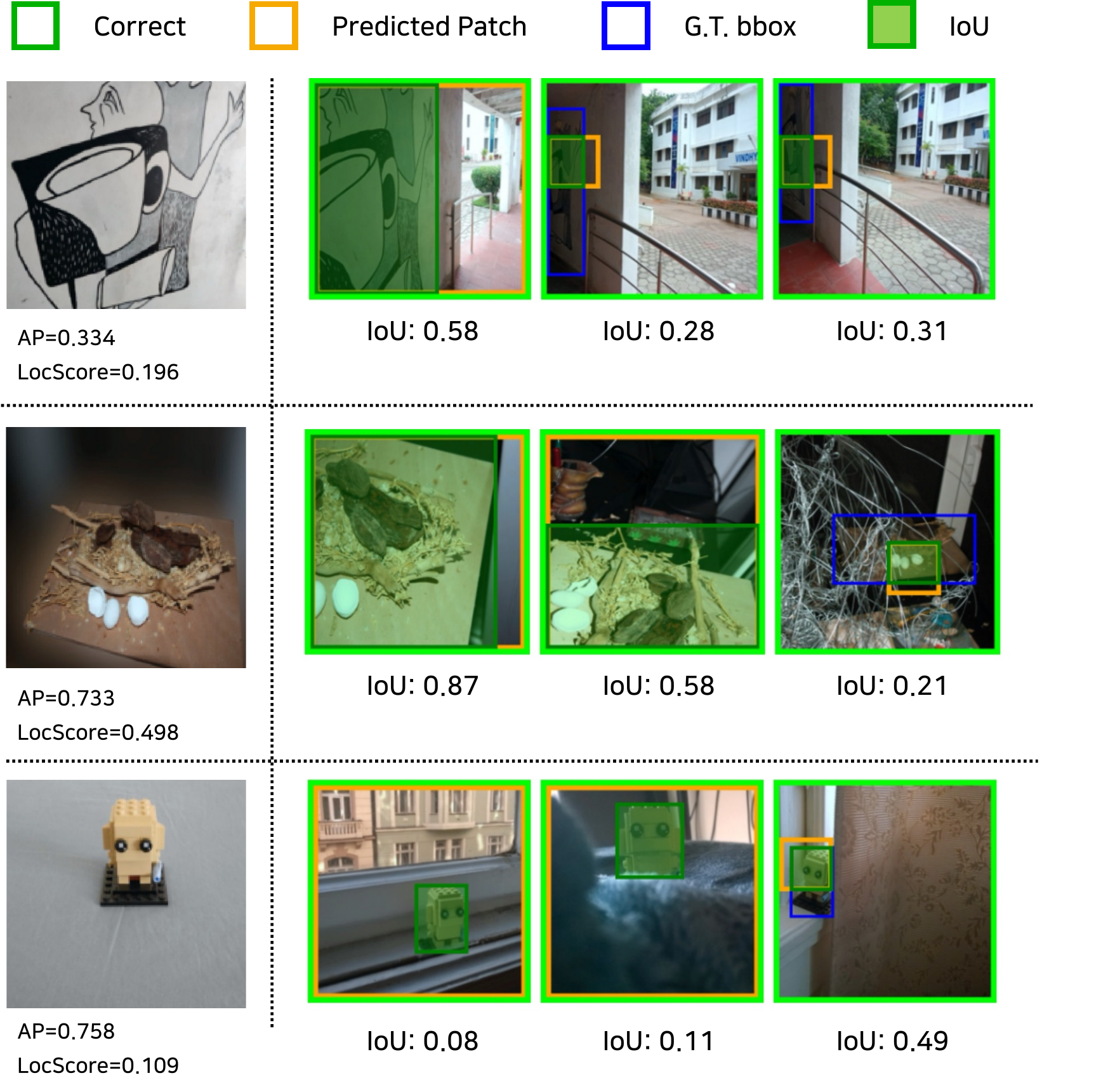}
    \caption{Comparison of AP and LocScore}
    \label{fig:comparison_locscore_ap2}
  \end{subfigure}

  \caption{\textbf{Qualitative results of LocScore variants.}}
  \label{fig:qualitative_locscore_all}
\end{figure*}

\paragraph{Qualitative results of LocScore variants}
Figure~\ref{fig:comparison_locscore_by_method} compares different localization strategies for the same query—Global, Patchify, Patchify with a sliding window (stride 0.25), and the Region Proposal variant. Improved localization accuracy leads to higher IoU, which subsequently improves both AP and LocScore. The Region Proposal method exhibits the best localization quality and obtains the highest scores.

Figure~\ref{fig:comparison_locscore_locscorethr} contrasts the continuous LocScore with its thresholded versions LocScore($\delta$) at different IoU thresholds. Although the continuous LocScore differs only by about 0.05 between the two rows, the thresholded scores change dramatically as $\delta$ increases, illustrating how sensitive LocScore($\delta$) is to the definition of acceptable localization quality.

Figure~\ref{fig:comparison_locscore_ap2} mirrors the example in the main paper, highlighting the relationship between AP and LocScore. Even when AP is relatively high, LocScore can remain low if the predicted regions have poor overlap with the ground-truth boxes, showing that improving LocScore requires not only strong ranking (high AP) but also accurate localization with high IoU.

\paragraph{Limitations.}
LocScore is tightly coupled with AP because it multiplies rank-based contribution by IoU, so the score can be overly driven by the localization quality of a few top-ranked positives while good localization at lower ranks has little influence. The thresholded variant partly alleviates this by enforcing a minimum IoU, but it introduces subjectivity since the notion of “good localization” depends on a human-chosen threshold. To reduce this ambiguity, we additionally report a mean version of LocScore and present both the continuous and thresholded variants side by side. We added this discussion to the supplementary.

\begin{table*}[t]
\centering
\renewcommand{\arraystretch}{1.15}
\begin{adjustbox}{width=0.9\linewidth}
\newcommand{\LocRow}[8]{%
\multirow{3}{*}{\textbf{#1}} & \multirow{3}{*}{\textbf{#2}} & \multirow{3}{*}{#3} & \multirow{3}{*}{#4} & \multirow{3}{*}{#5} & $\delta{=}0.3:\ #6$ \\
 & & & & & $\delta{=}0.4:\ #7$ \\
 & & & & & $\delta{=}0.5:\ #8$ \\}
\begin{tabular}{ll|cccc}
\toprule
\textbf{Model} & \textbf{Level} & \textbf{mAP@1k} & \textbf{LocScore (cont.)} & \textbf{mLocScore} & \textbf{LocScore ($\delta$)} \\
\midrule
\LocRow{SigLIP}{L0 (Global)}{0.2041}{6.59}{5.60}{7.55}{5.38}{3.86}
\LocRow{SigLIP}{L1}{0.2968}{10.08}{8.92}{12.56}{8.37}{5.83}
\LocRow{SigLIP}{L2}{0.5027}{18.11}{16.21}{24.78}{15.43}{8.41}
\LocRow{SigLIP}{L3}{0.2608}{8.81}{7.55}{10.58}{7.32}{4.76}
\addlinespace[4pt]
\LocRow{SigLIP$^\dagger$}{L0 (Global)}{0.3386}{9.59}{7.40}{10.39}{7.11}{4.72}
\LocRow{SigLIP$^\dagger$}{L1}{0.3655}{11.08}{8.99}{12.69}{8.50}{5.78}
\LocRow{SigLIP$^\dagger$}{L2}{0.3924}{12.29}{9.92}{14.22}{9.46}{6.08}
\LocRow{SigLIP$^\dagger$}{L3}{0.4048}{12.82}{10.33}{15.04}{9.77}{6.19}
\addlinespace[4pt]
\LocRow{DINOv3}{L0 (Global)}{0.2180}{8.02}{7.40}{9.55}{7.26}{5.39}
\LocRow{DINOv3}{L1}{0.3147}{11.48}{10.25}{14.22}{9.68}{6.86}
\LocRow{DINOv3}{L2}{0.3954}{15.71}{15.31}{22.33}{14.53}{9.08}
\LocRow{DINOv3}{L3}{0.4298}{17.05}{16.44}{24.72}{15.32}{9.29}
\bottomrule
\end{tabular}
\end{adjustbox}
\caption{Quantitative results on mini-ILIAS with mAP@1k and LocScore variants. A dagger ($^\dagger$) denotes the linear adaptation baseline from ILIAS; $\delta$ is the IoU threshold for the thresholded LocScore.}
\label{tab:mini_ilias_locscore}
\end{table*}

\label{fig:locscore_threshold}

\newpage
\begin{table*}[t]
\centering
\renewcommand{\arraystretch}{1.2}
\begin{adjustbox}{width=0.8\linewidth}
{\Large
\begin{tabular}{ll|cccc|cccc}
\hline
\textbf{}                          & \textbf{}                    & \multicolumn{4}{c}{\textbf{INSTRE}}& \multicolumn{4}{c}{\textbf{ILIAS}} \\
\textbf{model}                              & \textbf{Level}               & \textbf{mAP}           & \textbf{LocScore}      & \textbf{mLocScore} & \textbf{LocScore($\delta$)} & \textbf{mAP}           & \textbf{LocScore}      & \textbf{mLocScore} & \textbf{LocScore($\delta$)} \\ \hline
\multirow{3}{*}{\textbf{VGG16}}             & \multirow{3}{*}{\textbf{L0}} & \multirow{3}{*}{26.690} & \multirow{3}{*}{8.842}  & \multirow{3}{*}{6.960}  & $\delta$=0.3 : 10.928        & \multirow{3}{*}{5.864}  & \multirow{3}{*}{2.625}  & \multirow{3}{*}{2.866}  & $\delta$=0.3 : 3.540         \\
                                            &                              &                         &                         &                         & $\delta$=0.4 : 6.526         &                         &                         &                         & $\delta$=0.4 : 2.814         \\
                                            &                              &                         &                         &                         & $\delta$=0.5 : 3.426         &                         &                         &                         & $\delta$=0.5 : 2.245         \\
\multirow{3}{*}{}                           & \multirow{3}{*}{\textbf{L1}} & \multirow{3}{*}{39.859} & \multirow{3}{*}{13.697} & \multirow{3}{*}{9.453}  & $\delta$=0.3 : 16.400        & \multirow{3}{*}{7.399}  & \multirow{3}{*}{2.718}  & \multirow{3}{*}{2.468}  & $\delta$=0.3 : 3.359         \\
                                            &                              &                         &                         &                         & $\delta$=0.4 : 8.295         &                         &                         &                         & $\delta$=0.4 : 2.396         \\
                                            &                              &                         &                         &                         & $\delta$=0.5 : 3.665         &                         &                         &                         & $\delta$=0.5 : 1.649         \\
\multirow{3}{*}{}                           & \multirow{3}{*}{\textbf{L2}} & \multirow{3}{*}{51.186} & \multirow{3}{*}{19.189} & \multirow{3}{*}{14.331} & $\delta$=0.3 : 24.852        & \multirow{3}{*}{8.485}  & \multirow{3}{*}{2.978}  & \multirow{3}{*}{2.728}  & $\delta$=0.3 : 3.852         \\
                                            &                              &                         &                         &                         & $\delta$=0.4 : 12.865        &                         &                         &                         & $\delta$=0.4 : 2.614         \\
                                            &                              &                         &                         &                         & $\delta$=0.5 : 5.277         &                         &                         &                         & $\delta$=0.5 : 1.719         \\
\multirow{3}{*}{}                           & \multirow{3}{*}{\textbf{L3}} & \multirow{3}{*}{53.562} & \multirow{3}{*}{19.813} & \multirow{3}{*}{14.561} & $\delta$=0.3 : 25.461        & \multirow{3}{*}{8.764}  & \multirow{3}{*}{3.030}  & \multirow{3}{*}{2.716}  & $\delta$=0.3 : 3.941         \\
                                            &                              &                         &                         &                         & $\delta$=0.4 : 12.999        &                         &                         &                         & $\delta$=0.4 : 2.595         \\
                                            &                              &                         &                         &                         & $\delta$=0.5 : 5.223         &                         &                         &                         & $\delta$=0.5 : 1.613         \\ \hline 
\multirow{3}{*}{\textbf{ResNet101}}         & \multirow{3}{*}{\textbf{L0}} & \multirow{3}{*}{36.288} & \multirow{3}{*}{10.763} & \multirow{3}{*}{7.147}  & $\delta$=0.3 : 11.643        & \multirow{3}{*}{10.396} & \multirow{3}{*}{4.111}  & \multirow{3}{*}{4.064}  & $\delta$=0.3 : 5.296         \\
                                            &                              &                         &                         &                         & $\delta$=0.4 : 6.505         &                         &                         &                         & $\delta$=0.4 : 3.908         \\
                                            &                              &                         &                         &                         & $\delta$=0.5 : 3.293         &                         &                         &                         & $\delta$=0.5 : 2.989         \\   
\multirow{3}{*}{}                           & \multirow{3}{*}{\textbf{L1}} & \multirow{3}{*}{48.757} & \multirow{3}{*}{15.892} & \multirow{3}{*}{9.725}  & $\delta$=0.3 : 17.410        & \multirow{3}{*}{13.300} & \multirow{3}{*}{4.941}  & \multirow{3}{*}{4.367}  & $\delta$=0.3 : 6.110         \\
                                            &                              &                         &                         &                         & $\delta$=0.4  : 8.259        &                         &                         &                         & $\delta$=0.4 : 4.158         \\
                                            &                              &                         &                         &                         & $\delta$=0.5 : 3.504         &                         &                         &                         & $\delta$=0.5 : 2.832         \\
\multirow{3}{*}{}                           & \multirow{3}{*}{\textbf{L2}} & \multirow{3}{*}{58.588} & \multirow{3}{*}{21.058} & \multirow{3}{*}{14.150} & $\delta$=0.3 : 25.269        & \multirow{3}{*}{15.410} & \multirow{3}{*}{5.573}  & \multirow{3}{*}{4.969}  & $\delta$=0.3 : 7.309         \\
                                            &                              &                         &                         &                         & $\delta$=0.4 : 12.284        &                         &                         &                         & $\delta$=0.4 : 4.640         \\
                                            &                              &                         &                         &                         & $\delta$=0.5 : 4.896         &                         &                         &                         & $\delta$=0.5 : 2.959         \\
\multirow{3}{*}{}                           & \multirow{3}{*}{\textbf{L3}} & \multirow{3}{*}{61.382} & \multirow{3}{*}{21.558} & \multirow{3}{*}{13.893} & $\delta$=0.3 : 25.138        & \multirow{3}{*}{16.193} & \multirow{3}{*}{5.625}  & \multirow{3}{*}{4.853}  & $\delta$=0.3 : 7.276         \\
                                            &                              &                         &                         &                         & $\delta$=0.4 : 11.894        &                         &                         &                         & $\delta$=0.4 : 4.467         \\
                                            &                              &                         &                         &                         & $\delta$=0.5 : 4.646         &                         &                         &                         & $\delta$=0.5 : 2.815         \\ \hline   
\multirow{3}{*}{\textbf{Inception}}         & \multirow{3}{*}{\textbf{L0}} & \multirow{3}{*}{27.943} & \multirow{3}{*}{8.435}  & \multirow{3}{*}{5.585}  & $\delta$=0.3 : 9.066         & \multirow{3}{*}{14.624} & \multirow{3}{*}{5.260}  & \multirow{3}{*}{4.859}  & $\delta$=0.3 : 6.410         \\
                                            &                              &                         &                         &                         & $\delta$=0.4 : 5.105         &                         &                         &                         & $\delta$=0.4 : 4.812         \\
                                            &                              &                         &                         &                         & $\delta$=0.5 : 2.584         &                         &                         &                         & $\delta$=0.5 : 3.355         \\
\multirow{3}{*}{}                           & \multirow{3}{*}{\textbf{L1}} & \multirow{3}{*}{34.592} & \multirow{3}{*}{10.866} & \multirow{3}{*}{6.325}  & $\delta$=0.3 : 11.091        & \multirow{3}{*}{18.026} & \multirow{3}{*}{6.418}  & \multirow{3}{*}{5.545}  & $\delta$=0.3 : 7.807         \\
                                            &                              &                         &                         &                         & $\delta$=0.4 : 5.464         &                         &                         &                         & $\delta$=0.4 : 5.275         \\
                                            &                              &                         &                         &                         & $\delta$=0.5 : 2.422         &                         &                         &                         & $\delta$=0.5 : 3.552         \\
\multirow{3}{*}{}                           & \multirow{3}{*}{\textbf{L2}} & \multirow{3}{*}{42.077} & \multirow{3}{*}{13.755} & \multirow{3}{*}{7.676}  & $\delta$=0.3 : 13.787        & \multirow{3}{*}{21.385} & \multirow{3}{*}{7.484}  & \multirow{3}{*}{6.557}  & $\delta$=0.3 : 9.831         \\
                                            &                              &                         &                         &                         & $\delta$=0.4 : 6.513         &                         &                         &                         & $\delta$=0.4 : 6.187         \\
                                            &                              &                         &                         &                         & $\delta$=0.5 : 2.728         &                         &                         &                         & $\delta$=0.5 : 3.652         \\
\multirow{3}{*}{}                           & \multirow{3}{*}{\textbf{L3}} & \multirow{3}{*}{45.822} & \multirow{3}{*}{14.112} & \multirow{3}{*}{6.952}  & $\delta$=0.3 : 12.808        & \multirow{3}{*}{23.631} & \multirow{3}{*}{7.757}  & \multirow{3}{*}{6.310}  & $\delta$=0.3 : 9.769         \\
                                            &                              &                         &                         &                         & $\delta$=0.4 : 5.698         &                         &                         &                         & $\delta$=0.4 : 5.810         \\
                                            &                              &                         &                         &                         & $\delta$=0.5 : 2.350         &                         &                         &                         & $\delta$=0.5 : 3.350         \\ \hline   
\multirow{3}{*}{\textbf{ConvNext\_IN1k}}    & \multirow{3}{*}{\textbf{L0}} & \multirow{3}{*}{38.972} & \multirow{3}{*}{11.006} & \multirow{3}{*}{6.649}  & $\delta$=0.3 : 10.981        & \multirow{3}{*}{8.606}  & \multirow{3}{*}{3.253}  & \multirow{3}{*}{3.266}  & $\delta$=0.3 : 4.153         \\
                                            &                              &                         &                         &                         & $\delta$=0.4 : 5.993         &                         &                         &                         & $\delta$=0.4 : 3.253         \\
                                            &                              &                         &                         &                         & $\delta$=0.5 : 2.974         &                         &                         &                         & $\delta$=0.5 : 2.394         \\
\multirow{3}{*}{}                           & \multirow{3}{*}{\textbf{L1}} & \multirow{3}{*}{47.535} & \multirow{3}{*}{14.549} & \multirow{3}{*}{7.727}  & $\delta$=0.3 : 13.879        & \multirow{3}{*}{2.517}  & \multirow{3}{*}{0.705}  & \multirow{3}{*}{0.505}  & $\delta$=0.3 : 0.820         \\
                                            &                              &                         &                         &                         & $\delta$=0.4 : 6.535         &                         &                         &                         & $\delta$=0.4 : 0.451         \\
                                            &                              &                         &                         &                         & $\delta$=0.5 : 2.768         &                         &                         &                         & $\delta$=0.5 : 0.243         \\
\multirow{3}{*}{}                           & \multirow{3}{*}{\textbf{L2}} & \multirow{3}{*}{54.045} & \multirow{3}{*}{17.939} & \multirow{3}{*}{10.173} & $\delta$=0.3 : 18.403        & \multirow{3}{*}{2.232}  & \multirow{3}{*}{0.549}  & \multirow{3}{*}{0.355}  & $\delta$=0.3 : 0.658         \\
                                            &                              &                         &                         &                         & $\delta$=0.4 : 8.663         &                         &                         &                         & $\delta$=0.4 : 0.312         \\
                                            &                              &                         &                         &                         & $\delta$=0.5 : 3.452         &                         &                         &                         & $\delta$=0.5 : 0.096         \\
\multirow{3}{*}{}                           & \multirow{3}{*}{\textbf{L3}} & \multirow{3}{*}{56.254} & \multirow{3}{*}{18.163} & \multirow{3}{*}{9.793}  & $\delta$=0.3 : 17.984        & \multirow{3}{*}{1.902}  & \multirow{3}{*}{0.421}  & \multirow{3}{*}{0.206}  & $\delta$=0.3 : 0.392         \\
                                            &                              &                         &                         &                         & $\delta$=0.4 : 8.203         &                         &                         &                         & $\delta$=0.4 : 0.187         \\
                                            &                              &                         &                         &                         & $\delta$=0.5 : 3.193         &                         &                         &                         & $\delta$=0.5 : 0.041              \\   

\hline 

\end{tabular}}
\end{adjustbox}
\caption{Quantitative results with mAP and LocScore under various backbones.}
\label{supple_encoder_map_locscore1}
\end{table*}

\newpage
\begin{table*}[t]
\centering
\renewcommand{\arraystretch}{1.2}
\begin{adjustbox}{width=0.8\linewidth}
{\Large
\begin{tabular}{ll|cccc|cccc}
\hline
\textbf{}                          & \textbf{}                    & \multicolumn{4}{c}{\textbf{INSTRE}}& \multicolumn{4}{c}{\textbf{ILIAS}} \\
\textbf{model}                              & \textbf{Level}               & \textbf{mAP}           & \textbf{LocScore}      & \textbf{mLocScore} & \textbf{LocScore($\delta$)} & \textbf{mAP}           & \textbf{LocScore}      & \textbf{mLocScore} & \textbf{LocScore($\delta$)} \\ \hline
\multirow{3}{*}{\textbf{ConvNext\_LAION2b}} & \multirow{3}{*}{\textbf{L0}} & \multirow{3}{*}{77.949} & \multirow{3}{*}{19.005} & \multirow{3}{*}{9.171}  & $\delta$=0.3 : 15.531        & \multirow{3}{*}{41.253} & \multirow{3}{*}{11.886} & \multirow{3}{*}{9.156}  & $\delta$=0.3 : 12.752        \\
                                            &                              &                         &                         &                         & $\delta$=0.4 : 8.135         &                         &                         &                         & $\delta$=0.4 : 8.672         \\
                                            &                              &                         &                         &                         & $\delta$=0.5 : 3.847         &                         &                         &                         & $\delta$=0.5 : 6.043         \\
\multirow{3}{*}{}                           & \multirow{3}{*}{\textbf{L1}} & \multirow{3}{*}{86.710} & \multirow{3}{*}{24.872} & \multirow{3}{*}{11.977} & $\delta$=0.3 : 21.397        & \multirow{3}{*}{50.754} & \multirow{3}{*}{16.078} & \multirow{3}{*}{12.302} & $\delta$=0.3 : 19.025        \\
                                            &                              &                         &                         &                         & $\delta$=0.4 : 10.228        &                         &                         &                         & $\delta$=0.4 : 11.002        \\
                                            &                              &                         &                         &                         & $\delta$=0.5 : 4.306         &                         &                         &                         & $\delta$=0.5 : 6.880         \\
\multirow{3}{*}{}                           & \multirow{3}{*}{\textbf{L2}} & \multirow{3}{*}{91.524} & \multirow{3}{*}{30.159} & \multirow{3}{*}{16.453} & $\delta$=0.3 : 30.010        & \multirow{3}{*}{60.215} & \multirow{3}{*}{20.789} & \multirow{3}{*}{17.296} & $\delta$=0.3 : 27.039        \\
                                            &                              &                         &                         &                         & $\delta$=0.4 : 13.943        &                         &                         &                         & $\delta$=0.4 : 16.130        \\
                                            &                              &                         &                         &                         & $\delta$=0.5 : 5.407         &                         &                         &                         & $\delta$=0.5 : 8.719         \\
\multirow{3}{*}{}                           & \multirow{3}{*}{\textbf{L3}} & \multirow{3}{*}{92.870} & \multirow{3}{*}{30.936} & \multirow{3}{*}{16.953} & $\delta$=0.3 : 31.109        & \multirow{3}{*}{64.441} & \multirow{3}{*}{22.221} & \multirow{3}{*}{18.333} & $\delta$=0.3 : 29.363        \\
                                            &                              &                         &                         &                         & $\delta$=0.4 : 14.263        &                         &                         &                         & $\delta$=0.4 : 16.688        \\
                                            &                              &                         &                         &                         & $\delta$=0.5 : 5.488         &                         &                         &                         & $\delta$=0.5 : 8.948         \\
                                          
                                            \hline
\multirow{3}{*}{\textbf{DINOv2}}   & \multirow{3}{*}{\textbf{L0}} & \multirow{3}{*}{57.701} & \multirow{3}{*}{15.074} & \multirow{3}{*}{8.063}  & $\delta$=0.3 : 13.593        & \multirow{3}{*}{40.557} & \multirow{3}{*}{12.181} & \multirow{3}{*}{9.698}  & {$\delta$=0.3 : 13.332}       \\
                                   &                              &                         &                         &                         & $\delta$=0.4 : 7.185         &                         &                         &                         & {$\delta$=0.4 : 9.374}        \\
                                   &                              &                         &                         &                         & $\delta$=0.5 : 3.411         &                         &                         &                         & {$\delta$=0.5 : 6.386}        \\
\multirow{3}{*}{}                  & \multirow{3}{*}{\textbf{L1}} & \multirow{3}{*}{64.621} & \multirow{3}{*}{18.738} & \multirow{3}{*}{8.781}  & $\delta$=0.3 : 16.078        & \multirow{3}{*}{46.825} & \multirow{3}{*}{14.932} & \multirow{3}{*}{11.476} & {$\delta$=0.3 : 16.526}       \\
                                   &                              &                         &                         &                         & $\delta$=0.4 : 7.285         &                         &                         &                         & {$\delta$=0.4 : 10.594}       \\
                                   &                              &                         &                         &                         & $\delta$=0.5 : 2.980         &                         &                         &                         & {$\delta$=0.5 : 7.309}        \\
\multirow{3}{*}{}                  & \multirow{3}{*}{\textbf{L2}} & \multirow{3}{*}{69.978} & \multirow{3}{*}{22.183} & \multirow{3}{*}{11.065} & $\delta$=0.3 : 20.583        & \multirow{3}{*}{52.918} & \multirow{3}{*}{17.371} & \multirow{3}{*}{13.460} & {$\delta$=0.3 : 20.722}       \\
                                   &                              &                         &                         &                         & $\delta$=0.4 : 9.056         &                         &                         &                         & {$\delta$=0.4 : 12.330}       \\
                                   &                              &                         &                         &                         & $\delta$=0.5 : 3.556         &                         &                         &                         & {$\delta$=0.5 : 7.328}        \\
\multirow{3}{*}{}                  & \multirow{3}{*}{\textbf{L3}} & \multirow{3}{*}{72.543} & \multirow{3}{*}{22.216} & \multirow{3}{*}{10.343} & $\delta$=0.3 : 19.611        & \multirow{3}{*}{57.515} & \multirow{3}{*}{18.492} & \multirow{3}{*}{13.999} & {$\delta$=0.3 : 22.401}       \\
                                   &                              &                         &                         &                         & $\delta$=0.4 : 8.233         &                         &                         &                         & {$\delta$=0.4 : 12.485}       \\
                                   &                              &                         &                         &                         & $\delta$=0.5 : 3.184         &                         &                         &                         & {$\delta$=0.5 : 7.110}        \\ \hline   
\multirow{3}{*}{\textbf{OpenCLIP}} & \multirow{3}{*}{\textbf{L0}} & \multirow{3}{*}{73.837} & \multirow{3}{*}{18.107} & \multirow{3}{*}{8.638}  & $\delta$=0.3 : 14.594        & \multirow{3}{*}{31.598} & \multirow{3}{*}{9.183}  & \multirow{3}{*}{7.143}  & {$\delta$=0.3 : 9.696}        \\
                                   &                              &                         &                         &                         & $\delta$=0.4 : 7.659         &                         &                         &                         & {$\delta$=0.4 : 6.827}        \\
                                   &                              &                         &                         &                         & $\delta$=0.5 : 3.660         &                         &                         &                         & {$\delta$=0.5 : 4.904}        \\
\multirow{3}{*}{}                  & \multirow{3}{*}{\textbf{L1}} & \multirow{3}{*}{78.812} & \multirow{3}{*}{23.034} & \multirow{3}{*}{11.454} & $\delta$=0.3 : 21.196        & \multirow{3}{*}{39.819} & \multirow{3}{*}{12.170} & \multirow{3}{*}{9.302}  & {$\delta$=0.3 : 14.501}       \\
                                   &                              &                         &                         &                         & $\delta$=0.4 : 9.428         &                         &                         &                         & {$\delta$=0.4 : 8.278}        \\
                                   &                              &                         &                         &                         & $\delta$=0.5 : 3.737         &                         &                         &                         & {$\delta$=0.5 : 5.129}        \\
\multirow{3}{*}{}                  & \multirow{3}{*}{\textbf{L2}} & \multirow{3}{*}{85.594} & \multirow{3}{*}{29.556} & \multirow{3}{*}{18.428} & $\delta$=0.3 : 33.733        & \multirow{3}{*}{48.878} & \multirow{3}{*}{16.425} & \multirow{3}{*}{14.039} & {$\delta$=0.3 : 21.976}       \\
                                   &                              &                         &                         &                         & $\delta$=0.4 : 15.674        &                         &                         &                         & {$\delta$=0.4 : 13.256}       \\
                                   &                              &                         &                         &                         & $\delta$=0.5 : 5.877         &                         &                         &                         & {$\delta$=0.5 : 6.885}        \\
\multirow{3}{*}{}                  & \multirow{3}{*}{\textbf{L3}} & \multirow{3}{*}{87.565} & \multirow{3}{*}{30.370} & \multirow{3}{*}{18.936} & $\delta$=0.3 : 34.804        & \multirow{3}{*}{53.351} & \multirow{3}{*}{17.945} & \multirow{3}{*}{14.767} & {$\delta$=0.3 : 23.706}       \\
                                   &                              &                         &                         &                         & $\delta$=0.4 : 16.053        &                         &                         &                         & {$\delta$=0.4 : 13.598}       \\
                                   &                              &                         &                         &                         & $\delta$=0.5 : 5.953         &                         &                         &                         & {$\delta$=0.5 : 6.998}        \\ \hline   
\multirow{3}{*}{\textbf{SigLIP}}   & \multirow{3}{*}{\textbf{L0}} & \multirow{3}{*}{78.483} & \multirow{3}{*}{18.918} & \multirow{3}{*}{9.120}  & $\delta$=0.3 : 15.261        & \multirow{3}{*}{55.029} & \multirow{3}{*}{14.305} & \multirow{3}{*}{9.969}  & {$\delta$=0.3 : 14.345}       \\
                                   &                              &                         &                         &                         & $\delta$=0.4 : 8.135         &                         &                         &                         & {$\delta$=0.4 : 9.469}        \\
                                   &                              &                         &                         &                         & $\delta$=0.5 : 3.965         &                         &                         &                         & {$\delta$=0.5 : 6.095}        \\
\multirow{3}{*}{}                  & \multirow{3}{*}{\textbf{L1}} & \multirow{3}{*}{83.056} & \multirow{3}{*}{21.719} & \multirow{3}{*}{10.335} & $\delta$=0.3 : 17.815        & \multirow{3}{*}{59.449} & \multirow{3}{*}{16.631} & \multirow{3}{*}{11.911} & {$\delta$=0.3 : 17.504}       \\
                                   &                              &                         &                         &                         & $\delta$=0.4 : 9.037         &                         &                         &                         & {$\delta$=0.4 : 11.195}       \\
                                   &                              &                         &                         &                         & $\delta$=0.5 : 4.154         &                         &                         &                         & {$\delta$=0.5 : 7.033}        \\
\multirow{3}{*}{}                  & \multirow{3}{*}{\textbf{L2}} & \multirow{3}{*}{85.965} & \multirow{3}{*}{23.888} & \multirow{3}{*}{11.526} & $\delta$=0.3 : 20.255        & \multirow{3}{*}{63.319} & \multirow{3}{*}{18.842} & \multirow{3}{*}{13.888} & {$\delta$=0.3 : 20.688}       \\
                                   &                              &                         &                         &                         & $\delta$=0.4 : 9.963         &                         &                         &                         & {$\delta$=0.4 : 13.368}       \\
                                   &                              &                         &                         &                         & $\delta$=0.5 : 4.360         &                         &                         &                         & {$\delta$=0.5 : 7.607}        \\
\multirow{3}{*}{}                  & \multirow{3}{*}{\textbf{L3}} & \multirow{3}{*}{87.015} & \multirow{3}{*}{24.285} & \multirow{3}{*}{11.397} & $\delta$=0.3 : 20.151        & \multirow{3}{*}{65.165} & \multirow{3}{*}{19.745} & \multirow{3}{*}{14.613} & {$\delta$=0.3 : 21.933}       \\
                                   &                              &                         &                         &                         & $\delta$=0.4 : 9.777         &                         &                         &                         & {$\delta$=0.4 : 13.933}       \\
                                   &                              &                         &                         &                         & $\delta$=0.5 : 4.264         &                         &                         &                         & {$\delta$=0.5 : 7.974} \\ \hline    
\end{tabular}}
\end{adjustbox}
\caption{Quantitative results with mAP and LocScore under various backbones.}
\label{supple_encoder_map_locscore2}
\end{table*}
\label{fig:map_locscore}

\section{Qualitative Results}
\begin{center}
\begin{minipage}{\textwidth}
  \centering
  \includegraphics[width=\textwidth]{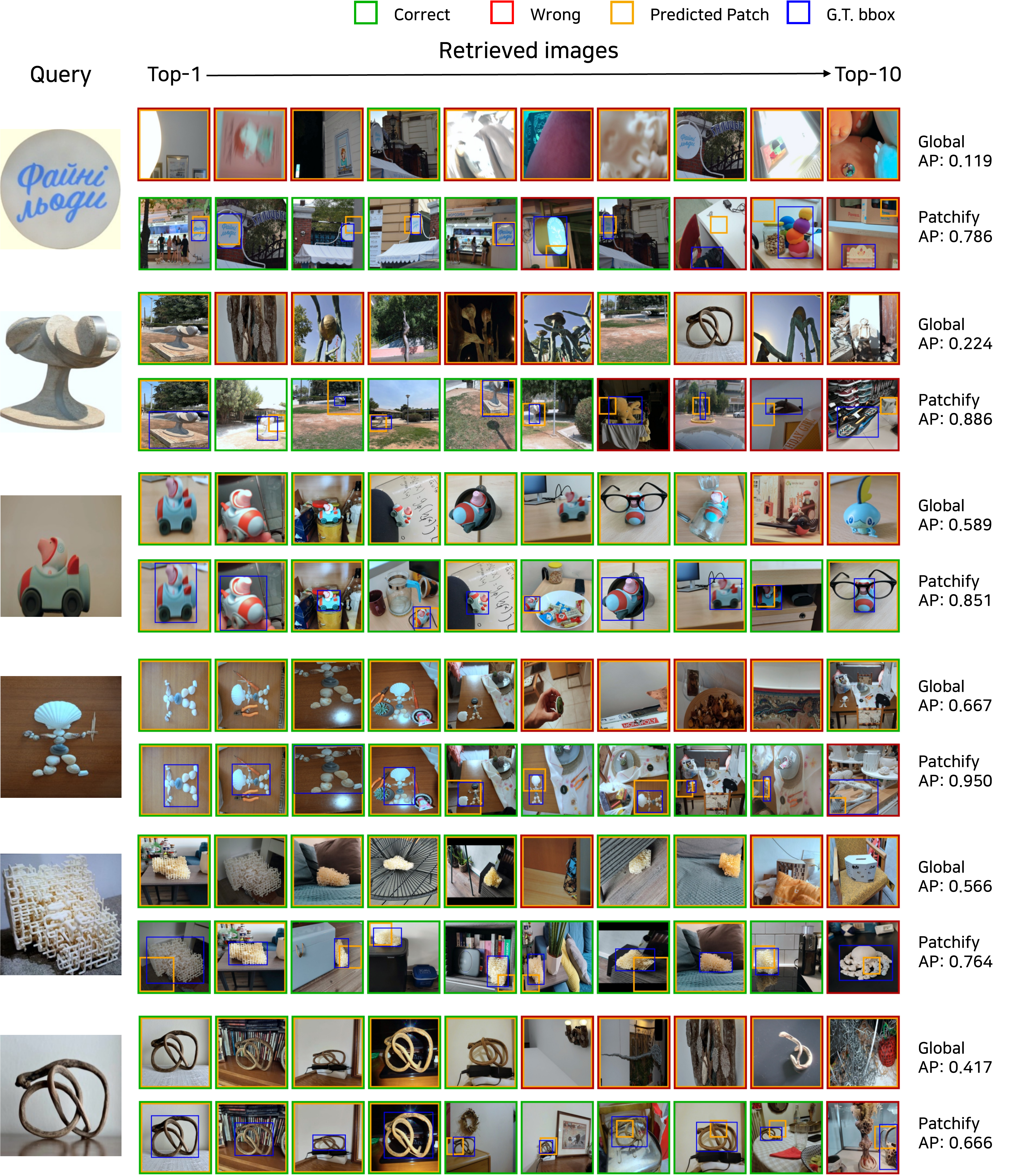}
  \captionof{figure}{Qualitative results of Global and Patchify on ILIAS}
  \label{fig:qualitative_supple_ilias}
\end{minipage}
\end{center}

\begin{figure*}[t]
  \centering
    \includegraphics[width=\linewidth]{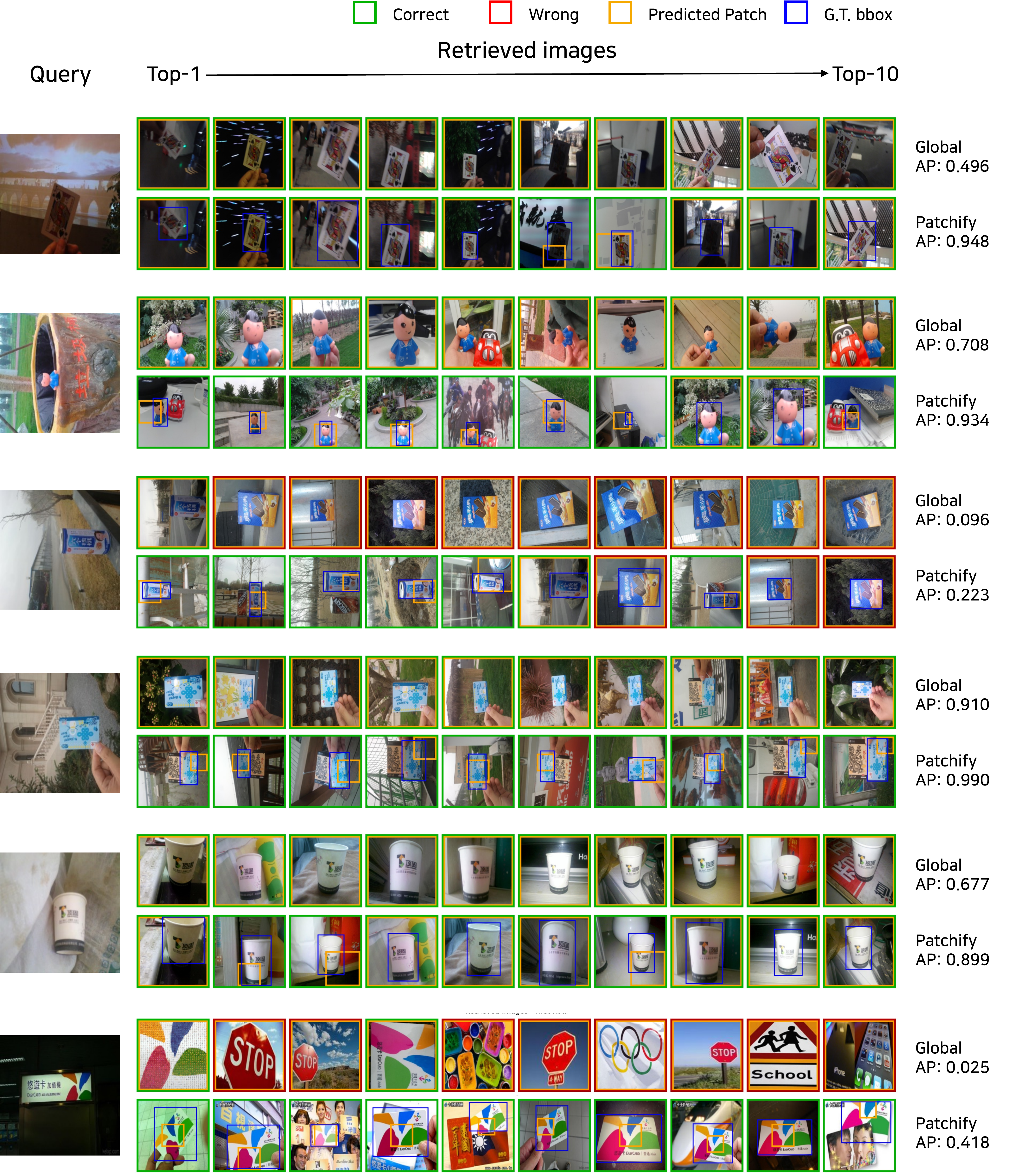}
    \caption{Qualitative results of Global and Patchify on INSTRE}
    \label{fig:qualitative_supple_instre}
  \vspace{1em}
\end{figure*}
\label{sec:qual}

\end{document}